\documentclass[10pt]{article}

\usepackage{amssymb}
\usepackage{amsmath}
\usepackage{bbm}
\usepackage{amsthm}
\usepackage{hyperref}
\usepackage{float}
\usepackage{tikz}
\usetikzlibrary{shapes.misc}
\usepackage{mathrsfs}
\usepackage{enumerate}
\usepackage{graphicx}
\usepackage{booktabs}
\usepackage{subcaption}

\makeatletter
\@tfor\next:=ABCDEFGHIJKLMNOPQRSTUVWXYZ\do{%
  \expandafter\edef\csname c\next\endcsname{\noexpand\mathcal{\next}}
  \expandafter\edef\csname bb\next\endcsname{\noexpand\mathbb{\next}}
  \expandafter\edef\csname bf\next\endcsname{\noexpand\mathbf{\next}}
  \expandafter\edef\csname sc\next\endcsname{\noexpand\mathscr{\next}}
  \expandafter\edef\csname fr\next\endcsname{\noexpand\mathfrak{\next}}
  \expandafter\edef\csname rm\next\endcsname{\noexpand\mathrm{\next}}

}
\makeatother

\NewDocumentCommand{\E}{ e{_} m o }{%
  \bbE\IfValueT{#1}{_{#1}}%
  \IfNoValueTF{#3}{%
    \left[#2\right]%
  }{%
    \left[#2\,\middle|\,#3\right]%
  }%
}

\RenewDocumentCommand{\P}{ e{_} m o }{%
  \bbP\IfValueT{#1}{_{#1}}%
  \IfNoValueTF{#3}{%
    \left(#2\right)%
  }{%
    \left(#2\,\middle|\,#3\right)%
  }%
}

\renewcommand{\d}{\mathrm{d}} 
\newcommand{\e}{\mathrm{e}} 
\newcommand{\R}{\bbR} 
\newcommand{\N}{\bbN} 
\newcommand{\1}{\mathbbm 1} 
\newcommand{\Id}{\mathrm{Id}} 
\newcommand{\id}{\mathrm{id}}

\newcommand{\bigo}{\cO} 
\newcommand{\smallo}{\scalebox{0.7}{\(\cO\)}}

\newcommand{\dsint}{\displaystyle\int}

\newcommand{\widesim}[2][1.5]{
  \mathrel{\overset{#2}{\scalebox{#1}[1]{\(\sim\)}}}
}

\newtheorem{theorem}{Theorem}
\newtheorem{lemma}[theorem]{Lemma}
\newtheorem{proposition}[theorem]{Proposition}
\newtheorem{corollary}[theorem]{Corollary}

\theoremstyle{definition}
\newtheorem{definition}[theorem]{Definition}

\newcommand{\tagdef}[1]{\hypertarget{#1}{(\textbf{#1})}}
\newcommand{\tagref}[1]{\hyperlink{#1}{(\textbf{#1})}}
\newcommand{\figorbox}[2][width=\linewidth]{\IfFileExists{#2}{\includegraphics[#1]{#2}}{\fbox{\parbox{0.9\linewidth}{\centering missing figure: \texttt{\detokenize{#2}}}}}}

\hypersetup{
    colorlinks   = true,
    linkcolor    = blue!70!black,
    urlcolor     = green!60!black,
    citecolor    = red!90!black,
    filecolor    = blue!40!black
    }
    
\begin{document}
\title{Learning on-the-fly quasi-stationarity diagnostics from low-dimensional surrogates}

\author{No\'e Blassel\\
   \small{Mathematics for materials modelling, Institute of Mathematics}\\\small{EPFL, Lausanne, Switzerland}  \\
   \small{\href{mailto:noe.blassel@epfl.ch}{noe.blassel@epfl.ch}}}
\maketitle

\begin{abstract}
    Accelerated dynamics methods, such as parallel replica dynamics, require estimates of the decorrelation time, after which a process trapped in one of its metastable states is distributed according to the quasi-stationary distribution (QSD).
    In a method proposed by Binder, Leli\`evre and Simpson, decorrelation times are defined by coupling a particle approximation of the non-linear conditioned probability flow of the process with MCMC-inspired diagnostics.
    We explore an alternative to MCMC diagnostics, based on coarse-graining of non-linear probability flows and simulation-based inference.
    We first develop a coarse-graining framework for killed Markov jump processes conditioned on survival, proving uniform-in-time, non-asymptotic estimates approximating the non-linear probability flow of
    the true process by the non-linear flow of a surrogate process, defined with a collective variable. The framework does not assume reversibility of the true process, and also gives bounds which scale explicitly in an adiabatic regime.
    Motivated by these results, we then construct a quasi-stationarity diagnostic using simulation-based inference: a recurrent neural network (RNN), trained on particle approximations of non-linear effective flows, estimates posterior probabilities of decorrelation. The learned diagnostic is then transferred to collective-variable traces of high-dimensional particle approximations.
    We investigate the behavior and robustness of this diagnostic on a set of numerical experiments, comparing it to the MCMC diagnostic of Binder et al.\ on both synthetic examples and a molecular system, in and out of its training prior distribution.
    When its prior assumptions hold, the RNN diagnostic outperforms the MCMC diagnostic in ROC analysis, is less conservative, fails less often, and has an interpretable decision threshold.
\end{abstract}


\tableofcontents

\section{Introduction}

Metastability is a common feature in the stochastic dynamics of complex systems (in this work described by a Markov process \(X\)), which has been observed and analyzed for dynamical models in a variety of fields: protein folding~\cite{LSBP13}, defect propagation in materials science~\cite{VMG02}, turbulent flows~\cite{BV12}, climate models~\cite{GL20}, dynamics of ecosystems~\cite{MACFGHLPSZ20}, and collective dynamics of neurons~\cite{RGARMWB19}.
In its own right, it has become a rich topic in the analysis of stochastic processes~\cite{FW12,BDH16,HKN04}.
This phenomenon can be broadly described as follows: due to the existence of nearly-invariant subsets of the configuration space, the dynamics \(X\) remains trapped in these so-called metastable states, leading to a timescale separation between within-state fluctuations and between-states transitions.

As \(X\) may also be stiff (as in molecular dynamics simulations~\cite{AT17}), the characteristic timescales of state-to-state transitions and the time step used to numerically integrate the dynamics are separated by many orders of magnitude.
Metastability therefore poses a fundamental computational challenge to the simulation of long trajectories needed to estimate dynamical properties, which are key in predicting emergent behavior. In molecular dynamics (MD), this challenge is known as the timescale problem.

Starting in 1997, Arthur Voter and his collaborators developed a family of numerical methods, so-called \emph{accelerated molecular dynamics} (AMD) algorithms~\cite{V97,V97b,V98,V98b,SV00}, aimed at the simulation of long trajectories for MD models of materials.
Rigorous justifications for all of these methods rely on a local description of metastability, organized around the notion of quasi-stationary distribution~\cite{MV12}.

\subsection{Local description of metastability}
As \(X\) remains trapped for progressively long times in a metastable subset \(\cX\) of the configuration space, its time-marginal distributions converge (under suitable assumptions) to a local equilibrium within \(\cX\), which we denote~$\nu^X\in\cP(\cX)$.
This local equilibrium, which describes the long-time distribution of~\(X\), conditioned on staying trapped in \(\cX\), is known as the Yaglom limit.
Under general conditions on \((X,\cX)\) (see~\cite{CV23,BCOV25}), this limit can be shown to be unique and independent of the initial distribution of \(X\). Moreover, whenever the Yaglom limit exists, it is a quasi-stationary distribution (QSD). Formally, these two properties can be written as
\begin{equation}
    \label{eq:decorrelation_definition}
    \eta_t^X := \mathrm{Law}\left(X_t\,\middle|\,\tau^X>t\right) \xrightarrow[t\to\infty]{} \nu^X,\qquad \forall\,A\subset \cX,\,\quad\P_{\nu^X}{X_t\in A}[\tau^X>t] = \nu^X(A),
\end{equation}
where \(\tau^X\) denotes the exit time of \(X\) from \(\cX\), \(\eta_0^X\in\cP(\cX)\) is the initial distribution, and~\(A\) is supposed measurable. A fundamental property of the QSD concerns the exit distribution of~\(X\) from~\(\cX\), when~\(X_0\) is distributed according to~\(\nu^X\):
\begin{equation}
    \label{eq:exit_distribution}
    \forall\,A\subset\partial\cX,\qquad \P_{\nu^X}{\tau^X>t,X_{\tau^X}\in\, A}=\e^{-\lambda^X t}\P_{\nu^X}{X_{\tau^X}\in\,A},
\end{equation}
meaning that the exit time~\(\tau^X\) is exponentially distributed (with rate~\(\lambda^X>0\)) and independent of the exit point~\(X_{\tau^X}\).

QSDs were introduced by Yaglom~\cite{Y47} to study the pre-extinction behavior of certain critical branching processes, but the notion has proved useful in population and molecular dynamics, among other fields.
We refer to~\cite{MV12,CMSM13} for an introduction to QSDs and their main properties, as well as~\cite{CV23,BCOV25} for existence results in a large class of Markov processes.

\subsection{Accelerated simulation of metastable dynamics}
AMD methods rely on the crucial property~\eqref{eq:exit_distribution} to efficiently sample long, unbiased trajectories of the system, and on an informal level can be understood as the following simple three-step procedure: i) wait for \(X\) to reach a local-equilibrium within some metastable state~\(\cX\) (i.e. for the limit in~\eqref{eq:decorrelation_definition} to be realized), ii) sample~\((\tau^X,X_{\tau^X})\) efficiently using the property~\eqref{eq:exit_distribution}, iii) continue the simulation from~\(X_{\tau^X}\), incrementing the simulation clock by~\(\tau^X\), until another metastable state is reached, etc.
Unbiasedness of the sequence of visited metastable states is guaranteed by the independence property in~\eqref{eq:exit_distribution}, if samples in step ii) are exact. As a counterpart, details of the dynamics in the quasi-stationary portion of the trajectory, between convergence to the QSD and exit from the state, are sacrificed.

To efficiently sample the exit event, hyperdynamics~\cite{V97} and temperature-accelerated dynamics~\cite{SV00} rely on physical approximations of the exit distribution~\eqref{eq:exit_distribution}, using transition state theory (TST) and harmonic transition state theory (HTST) respectively, see~\cite{HTB90} and references therein. As such, they are only valid in specific physical regimes, which are relevant for solid-state materials applications.
The parallel replica algorithm~\cite{V98}, however, is more general, and relies only on the property~\eqref{eq:exit_distribution} to accelerate step ii) in the following way. A number~\(N\geq 1\) of independent samples from~\(\nu^X\) are drawn, and simulated in parallel until the first replica exits. The property~\eqref{eq:exit_distribution} then implies the equality
\begin{equation}
    \left(N\tau^X_{\min},X_{\tau^X_{\min}}^{(\min)}\right)\overset{\text{in law}}{=} \left(\tau^X,X_{\tau^X}\right),
\end{equation}
where~\(\tau_{\min}^X\) denotes the minimum exit time across all replicas, and~\(X^{(\min)}\) the first replica to exit. Provided the replicas can be simulated with perfect parallel efficiency, this procedure leads to an expected speedup of a factor~\(N\) in simulation time, for a given trajectory length.
The only approximation in the parallel replica algorithm is a time-truncation in the Yaglom limit, which makes it applicable in principle to a large class of metastable processes.
The numerical analysis of this algorithm was initiated in~\cite{LBLLP12,SL13}. For a perspective on implementation aspects, variants and challenges, we refer to~\cite{PUV15}.

\subsection{Challenges in AMD methods}
Despite their apparent simplicity (and generality, in the case of the parallel replica method), AMD methods have not been widely applied beyond materials science.
This can be explained, at least in part, by the fact that AMD users must provide two key parameters, which have natural and suitable definitions for materials, but few beyond this setting.
These are i) definitions of the metastable states~\(\cX\) and ii) definitions of valid decorrelation times~\(t_{\rm corr} = t_{\rm corr}(\cX)>0\) within each state, to be used as truncation parameters in Yaglom's limit~\eqref{eq:decorrelation_definition}.

Regarding problem i), natural definitions of~\(\cX\), when~\(X\) is a small-noise perturbation of a deterministic dynamical system, are available as positively invariant subsets of the deterministic process (e.g. basins of attraction of some fixed point).
Recent mathematical works~\cite{BLS25a,BLS26} have also proposed optimizing the shape of~\(\cX\) to directly increase the efficiency of the parallel replica algorithm. Beyond these, many practical simulation methods still rely on hand-crafted, ad-hoc definitions of~\(\cX\).

Every definition of the decorrelation time to address problem ii) relies, implicitly or explicitly, on a quantitative version of Yaglom's limit, typically in the form
\begin{equation}
    \label{eq:decorrelation_quantitative_definition}
    \forall\,t\geq 0,\qquad\d_{\rm TV}\left(\eta_t^X,\nu^X\right)\leq C(\cX,\eta_0^X)\e^{-\gamma^X t},
\end{equation}
where~\(C(\cX,\eta_0^X)>0\) depends on the definition of the metastable state and the initial distribution~$\eta_0^X$ of~\(X\) within~$\cX$, but is independent of time. The constant~\(\gamma^X>0\) is called a~\emph{spectral gap}, for reasons which will become clear in Section~\ref{sec:theory}, and~\(\d_{\rm TV}\) is the total variation distance (also defined below, see Equation~\eqref{eq:tv_distance}).
We refer to~\cite{CV23} and~\cite[Theorem 1.8]{BCOV25} for sufficient conditions on~\((X,\cX)\) for~\eqref{eq:decorrelation_quantitative_definition} to hold.

Assuming this convergence holds, various solutions to problem ii) have been proposed.
\begin{itemize}
    \item[\(\circ\)]{\emph{Physical heuristics. }{Many implementations of the parallel replica method rely on~\(\cX\)-independent definitions of~\(t_{\rm corr}\), based on domain expertise and validated a posteriori by checking that the exit statistics satisfy~\eqref{eq:exit_distribution}. This includes the original implementations~\cite{V98,V98b} where~\(1\leq t_{\rm corr}\leq 2\,{\rm ps}\). Slightly more principled, but still heuristic, definitions rely, for small-noise diffusions, on linearizations~\cite{PUV15} of the dynamics around some stable equilibrium~\(x_0\in\cX\), whenever~\(\cX\) is the basin of attraction of~\(x_0\) for the zero-noise dynamics. The closed form of this linear process is then used to define~\(t_{\rm corr}(x_0)=1/\gamma^X_{\rm lin}\), where~\(\gamma^X_{\rm lin}\) is the spectral gap of this linearization.}}
    \item[\(\circ\)]{\emph{Quantitative spectral asymptotics. }{When~\(X\) is a reversible diffusion process and~$\cX$ is bounded,~\(\gamma^X=\lambda_2(\cX)-\lambda_1(\cX)\) is the spectral gap of the infinitesimal generator of~\(X\), endowed with homogeneous Dirichlet boundary conditions on~\(\partial\cX\)~\cite[Proposition 6]{LBLLP12}. In the low-temperature regime, it is possible to obtain precise,~\(\cX\)-dependent estimates~\(\gamma^X\sim\gamma^X_{\rm{sc}}(\cX)\) (see~\cite[Corollary 6]{BLS25a}), leading to a principled definition~\(t_{\rm corr}(\cX) := 1/\gamma^X_{\rm sc}(\cX)\). The subscript stands for ``semiclassical'' in allusion to the mathematical tools involved in the proofs of these results.}}
    \item[\(\circ\)]{\emph{Operator learning. }{These physically-inspired definitions are invalid in glassy landscapes encountered in biochemical systems or materials with defects, where metastable states often contain many energy minima separated by shallow barriers~\cite{PUV15}. A recent data-driven alternative~\cite{LNHP26} relies on trajectory data to directly estimate~\(\gamma^X=\lambda_2(\cX)-\lambda_1(\cX)\) in a Koopman operator-learning framework, adapted to the case of Dirichlet boundary conditions. This again leads to the definition~\(t_{\rm corr}(\cX) := 1/\widehat{\gamma}^X\) for reversible processes, where~\(\widehat{\gamma}^X\) is the spectral gap of the Koopman approximation.}}
    \item[\(\circ\)]{\emph{MCMC diagnostics. }This approach, proposed in~\cite{BLS15}, relies on a somewhat different philosophy. It combines a particle approximation to the conditioned time-marginal~\(\eta_t^X\) (in the form of a Fleming--Viot process~\cite{FV79}, see Definition~\ref{def:fleming_viot} below), and so-called convergence diagnostics from the Markov chain Monte Carlo literature (see~\cite{R20} for an overview of this important topic) to detect stationarity in this particle approximation.}
\end{itemize}
The first three approaches are \emph{offline} definitions, where the decorrelation time is a deterministic parameter in the algorithm (albeit a learned parameter in the operator learning approach), and the fourth is an online or~\emph{on-the-fly} definition, in which~\(t_{\rm corr}\) is a random variable whose value is determined dynamically from the simulation itself (in~\cite{BLS15},~\(t_{\rm corr}\) can be understood as a stopping time of the Fleming--Viot particle approximation).

On-the-fly definitions are a promising direction to free the parallel replica method from the user-defined parameter~\(t_{\rm corr}\), and therefore make it applicable in a wider class of metastable systems.
The first and so far only implementation of this online approach within the parallel replica method was proposed in~\cite{BLS15} (giving the so-called~\emph{generalized parallel replica} method), and relies on a diagnostic inspired by the Gelman--Rubin convergence test~\cite{GR92} (see~\eqref{eq:gelman_rubin} below). The method was tested on several biochemical systems in~\cite{HL19}.
Another attractive feature of on-the-fly definitions is their sensitivity to the initial condition~$\eta_0^X$, through the initial-condition-dependent prefactor~\(C(\cX,\eta_0^X)\) in~\eqref{eq:decorrelation_quantitative_definition}. Offline methods, by contrast, which rely on some estimate of the spectral gap~\(\gamma^X\), necessarily neglect this contribution.
It was observed in~\cite{HL19} that setting the decorrelation time uniformly in the initial condition was in many cases suboptimal.

The use of MCMC convergence diagnostics (which should really be understood as nonconvergence diagnostics) as a~quasi-stationary convergence diagnostic (or more simply, \emph{quasi-stationarity diagnostic}) is however heuristic, and just as in the case of physical heuristics, has to be validated a posteriori.
Moreover, it relies on a user-defined scalar decision threshold~\(\alpha\). As this threshold implicitly defines~\(t_{\rm corr}\), the method is not strictly speaking~\(t_{\rm corr}\)-free, but replaces one parameter by another. The question then becomes whether an appropriate value of~\(\alpha\) can be selected for many systems at once.

More critically, the Gelman--Rubin diagnostic proposed in~\cite{BLS15} is prone to being conservative (we provide some concurring numerical evidence in Section~\ref{sec:num}), which poses a challenge for applications to AMD.
Indeed, overestimating~\(t_{\rm corr}\) leads to wasted acceleration in the parallel replica method, by allocating unnecessary computation to simulating quasi-stationary portions of the trajectory.
On the other hand, a systematically lax diagnostic will introduce bias in the distribution of exit events~\eqref{eq:exit_distribution}, and consequently in the long-time kinetics of sampled trajectories.
This efficiency-bias tradeoff is clearly identified in the literature on parallel replica methods~\cite{LBLLP12,PUV15,BLS15,BLS26}.

\subsection{Contributions of this work and outline}
We explore alternatives to MCMC-inspired diagnostics to define on-the-fly decorrelation times. Our method uses simulation-based inference~\cite{CBL20} of decorrelation times, over a prior ensemble of low-dimensional surrogate processes.
In the course of developing our method, we make the following contributions.
\begin{enumerate}
    \item[\(\circ\)]{We develop a coarse-graining framework for the non-linear probability flows~$\eta_t^X$ of killed Markov jump processes conditioned on non-absorption. For the case of finitely many states, we give the first estimates, non-asymptotic in time, on the evolution of~\(\eta_t^X\)~(the conditional law in~\eqref{eq:decorrelation_definition} and~\eqref{eq:decorrelation_quantitative_definition}) in terms of a well-chosen effective flow~\(\eta_t^Z\) (Theorem~\ref{thm:cg_existential_result}).
              The framework, importantly, makes no assumption on the reversibility of~\(X\). This effective description then allows one to prove approximate convergence in the Yaglom limit~\eqref{eq:decorrelation_quantitative_definition} from convergence in the coarse-grained Yaglom limit of \(\eta_t^Z\) (Corollary~\ref{cor:diagnostic_bound}).}
    \item[\(\circ\)]{We identify an adiabatic regime in which the estimates of Theorem~\ref{thm:cg_existential_result} and Corollary~\ref{cor:diagnostic_bound} hold with explicit rates (Theorem~\ref{thm:perturbation_theory} and Corollary~\ref{cor:uniform_coarse_graining}).
                    This regime corresponds to a slow-fast decomposition of the infinitesimal generator of~$X$.}
    \item[\(\circ\)]{We use these theoretical results to motivate using a surrogate process~\(Z\) in a simulation-based Bayesian inference framework for quasi-stationarity diagnostics, in which the surrogate process~\(Z\) is treated as a nuisance parameter (Section~\ref{sec:sbi}) in the sense of~\cite[Section 3.1]{GCSR95}. This approach leads us to define~\(t_{\rm corr}\) as a stopping time for a Fleming--Viot approximation of~\(\eta_t^X\) (as in~\cite{BLS15}), defined using a recurrent neural network (RNN) rather than an MCMC-inspired diagnostic.
    The underlying assumptions are explicitly stated and discussed, for future extensions and mathematical analysis (Assumptions~\tagref{B1}--\tagref{B4}).}
    \item[\(\circ\)]{We prove that the decision threshold~\(\alpha\) in our RNN diagnostic has a probabilistic interpretation (Lemmas~\ref{lemma:randomized_loss_minimizer} and~\ref{lemma:risk_control}), and can therefore be tuned a priori under ideal assumptions, rather than validated a posteriori per-system as MCMC-inspired diagnostics require.}
    \item[\(\circ\)]{We conduct numerical experiments of increasing complexity, including on a molecular system~(Section~\ref{sec:num}). These experiments suggest that the RNN consistently outperforms the Gelman--Rubin diagnostic when its underlying assumptions hold. We also explore cases where these assumptions do not hold to identify plausible failure modes, as well as a simple mitigation strategy.}
\end{enumerate}
The work is organized as follows. In Section~\ref{sec:theory}, we develop our theoretical framework, and present our main theoretical results. In Section~\ref{sec:qs_criteria}, we develop our numerical method. In Section~\ref{sec:num}, we present the results of our experiments. We finally discuss our method and outline some perspectives in Section~\ref{sec:perspectives}, and dedicate Appendices~\ref{sec:proof_perturbation_theory} and~\ref{sec:details} to the proofs of some technical results and implementation details, respectively.

\section{Coarse-graining of killed Markov jump processes}
\label{sec:theory}
We develop a framework to obtain coarse-grained approximations of killed Markov processes \(X\), conditioned on non-absorption, at the level of probability flows.
Given a so-called collective variable \(\xi:\cX\to\cZ\) of the full state, our coarse-graining procedure produces another Markov process, \(Z\) the~\emph{surrogate process}, which is in some sense simpler than~\(X\), and whose conditioned time-marginals approximate those of the non-Markovian process \(\xi(X)\).
The problem is closely related to the study of so-called \emph{effective dynamics}, which are Markovian approximations to the generically non-Markovian processes \(\xi(X)\).
The case without killing has been studied extensively in the literature, see for instance~\cite{LL10,ZHS16,LLO17,DLPSS18} and references therein. Our problem is different, due to non-linearities induced by the conditioning at the level of time-marginals. By contrast, probability flows evolve linearly in the case without killing.

Due to this complexity of the non-linear case, we focus as a first step on the case of~\emph{finite state space} jump processes. We suspect our approach can be extended to more general dynamics, but this requires non-trivial technical adaptations for the case of diffusions.
An important feature of the framework is that the reversibility of~\(X\) is not assumed (the infinitesimal generator~\(\cL^X\) may be non-self-adjoint for every inner product on~\(\ell^\infty(\cX)\), i.e. detailed balance is not required). We therefore hope to extend in future work the framework to non-gradient diffusions, which are relevant to describe the quasi-stationary behavior of driven systems.

The ultimate goal of this framework is to obtain non-asymptotic estimates on the distance between the conditioned time-marginals of \(Z\) and those of \(X\), for the total variation metric. As explained in Section~\ref{sec:qs_criteria}, this objective is well-motivated for algorithmic reasons.
We will proceed in two steps, first bounding the difference between the conditional marginals of \(Z\) and those of \(\xi(X)\), then between those of \(\xi(X)\) and \(X\).
Under a formal assumption encoding the intuition that \(\xi\) is a \emph{slow variable} of the system (but only mildly so), we will derive non-asymptotic in time error bounds. These bounds in particular show that the coarse-graining bias is controlled uniformly in time by an approximation parameter encoding a separation of timescales between slow and fast variables.

\subsection{Killed Markov jump processes}
We consider \(X\), \(Z\) two jump processes on finite state spaces \(\overline{\cX}=\cX\cup\partial^{\cX}\) and \(\overline{\cZ}=\cZ\cup\partial^{\cZ}\), where \(\partial^{\cX},\partial^{\cZ}\) are so-called~\textit{killing boundaries}.

\paragraph{Conventions used throughout this work.}
Given a set \(S\), we denote by \(\cP(S)\) the set of probability measures on \(S\). We work in \(\ell^\infty(\cX),\ell^\infty(\cZ)\), with the standard norms. All vector and operator norms (be they on \(\cX\) or \(\cZ\)) will be denoted with \(\|\cdot\|\), while the dual (\(\ell^1\)) norm on covectors will be denoted by \(\|\cdot\|_*\).

\paragraph{The killed semigroup.}
In the following, \((Y,\cY)\in\left\{(X,\cX),(Z,\cZ)\right\}\). The killing time of the process \(Y\) is defined as the stopping time
\begin{equation}
    \label{eq:abstract_exit_times}
    \tau^Y = \inf\,\left\{t>0:\,Y_t\in\partial^{\cY}\right\}.
\end{equation}
We identify~\(\partial^\cY\) with a unique absorbing state, that is satisfying
\begin{equation}
    \P{Y_t \in \partial^{\cY}}[t\geq\tau^Y]=1,
\end{equation}
and assume that every non-absorbing state is connected to every other non-absorbing state, and to the boundary:
\begin{equation}
    \label{eq:irreducibility}
    \forall\,y,y'\in \cY,\qquad y \leftrightarrow y',\qquad y \rightarrow \partial^{\cY},
\end{equation}
where~\(y\rightarrow y'\) means~\(\P_y{Y_t = y'}>0\) for some and therefore all~\(t>0\), and~\(y\leftrightarrow y'\) means~\(y\rightarrow y'\) and~\(y'\rightarrow y\). This condition, which we call~\emph{conditional irreducibility}, ensures in particular that \(\tau^Y\) is almost-surely finite.

Absorption (or killing) at the boundary gives rise to a sub-Markovian evolution, encoded in the following linear semigroup,
\begin{equation}
    \label{eq:killed_semigroup}
    \forall\, f\in \ell^\infty(\cY),\qquad Q_t^Y f(y) = \E{f(Y_t)\1_{\tau^Y>t}}[Y_0=y],
\end{equation}
the so-called \textit{killed semigroup}. This semigroup can be seen equivalently as acting on the full state space \(\overline{\cY}\) with Dirichlet boundary conditions on \(\partial^{\cY}\).

That is, denoting \(\overline{\cL}^Y\) the infinitesimal generator of the unconditioned process in \(\overline{\cY}\), the semigroup \(Q^Y\) is generated by its restriction \(\cL^Y\) to the space
\begin{equation}
    \left\{f\in \ell^\infty(\overline{\cY}): \left. f\right|_{\partial^{\cY}} = 0 \right\} \cong \ell^\infty(\cY),
\end{equation}
which is invariant under the \(Q^Y\)-flow.

Using the notation
\begin{equation}
    \eta(f) = \int_{\cY} f(y)\,\eta(\d y)
\end{equation}
for \(\eta\in\cP(\cY)\), the law of \(Y_t\) conditioned on survival is given for any~$t\geq 0$ by
\begin{equation}
    \label{eq:non_linear_flow}
    \eta_t^Y(f) := \E{f(Y_t)}[\tau^Y>t] = \frac{\eta_0^Y\left(Q_t^Y f\right)}{\eta_0^Y\left(Q_t^Y \1_{\cY}\right)},
\end{equation}
where \(Y_0\sim\eta_0^Y\in\cP(\cY)\). The map \(t\mapsto \eta_t^Y\) defines a non-linear, probability measure-valued flow in \(\cP(\cY)\).

The key tool is the following Feynman--Kac formula, which in our simplified setting is just a reformulation of the backward Kolmogorov equation for the killed process.
For any observable \(f \in \ell^\infty(\cY)\), the trajectory \({u(t) := Q_t^Y f}\) solves the Cauchy problem
\begin{equation}
    \label{eq:feynman_kac}
    \left\{\begin{aligned}
        \partial_t u(t,y) & = \cL^Y u(t,y), & (t,y)\in (0,\infty)\times\cY, \\
        u(0,y)            & = f(y),         & y\in\cY.
    \end{aligned}\right.
\end{equation}

The following classical result (see~\cite[Section 3.1]{CMSM13}), which is a consequence of the Perron--Frobenius theorem, will be used extensively in the following.
\begin{proposition}[Spectral properties]
    \label{prop:spectral_properties}
    For each \(Y\in\{X,Z\}\), the following holds.
    \begin{itemize}
        \item[i)]{
                  There exists a unique principal eigentriplet
                  \begin{equation}
                      \left(\lambda^Y,\nu^Y,h^Y\right) \in (0,\infty)\times \cP(\cY)\times \ell^\infty(\cY)
                  \end{equation}
                  for the semigroup \(Q^Y\), i.e. for any \(\eta\in\cP(\cY)\), \(f\in \ell^\infty(\cY)\) and \(t>0\),
                  \begin{equation}
                      \label{eq:eigenrelation}
                      \nu^Y(Q_t^Y f) = \e^{-\lambda^Y t}\nu^Y(f),\qquad \eta(Q_t^Y h^Y) = \e^{-\lambda^Y t}\eta(h^Y),
                  \end{equation}
                  with exponential decay of the semigroup: there exists~\(C_\lambda^Y>0\) such that
                  \begin{equation}
                      \label{eq:principal_decay}
                      \|Q_t^Y\| \leq C_\lambda^Y \e^{-\lambda^Y t}.
                  \end{equation}
              }
        \item[ii)]{The eigenvector \(h^Y>0\) satisfies the normalization property
                  \begin{equation}
                      \label{eq:h_normalization}
                      \nu^Y(h^Y) = 1.
                  \end{equation}}
        \item[iii)]{
                  Denote by
                  \begin{equation}
                      \label{eq:spectral_projector}
                      P^Y f = f - \frac{\nu^Y(f)}{\nu^Y(h^Y)}h^Y
                  \end{equation}
                  the~\(\nu^Y\)-centering projector along the direction~\(h^Y\).

                  There exists a spectral gap, that is, constants \(\gamma^Y,C_\gamma^Y>0\) verifying the estimate
                  \begin{equation}
                      \label{eq:spectral_gap_estimate}
                      \|Q_t^Y P^Y\|\leq C_\gamma^Y\e^{-(\lambda^Y+\gamma^Y)t}.
                  \end{equation}
              }
    \end{itemize}
\end{proposition}
Differentiating in time the eigenrelations~\eqref{eq:eigenrelation} in the first item of Proposition~\ref{prop:spectral_properties}, we also obtain, for any \(f\in \ell^\infty(\cY)\), the generator-level eigenrelation
\begin{equation}
    \label{eq:dual_relations}
    \cL^Y h^Y = -\lambda^Y h^Y,\qquad \nu^Y(\cL^Y f) = -\lambda^Y\nu^Y(f).
\end{equation}
The probability measure \(\nu^Y\) is called the~\emph{quasi-stationary distribution} of the process \(Y\) in the metastable state \(\cY\).

\subsection{Coarse-graining approximation and bias}

Let \(\xi:\cX\to\cZ\) be a surjective \emph{coarse-graining} map.
Whenever \(f\in \ell^\infty(\cZ)\), we denote the pullback through \(\xi\) via
\begin{equation}
    \label{eq:lift}
    \xi^\star f := f\circ\xi \in \ell^\infty(\cX),
\end{equation}
and its dual, the pushforward acting on measures \(\eta\in\cP(\cX)\) by
\begin{equation}
    \label{eq:pushforward}
    \xi_\star\eta = \eta(\xi^\star\cdot).
\end{equation}
Since we have~$\xi(\cX)=\cZ$, we may assume by a suitable extension of~$\xi$ to~$\overline\cX$ that~$\xi(\partial^\cX)=\partial^{\cZ}$, so that~$\tau^X$ is a stopping time for~$\xi\circ X$.

For some fixed observable \(f\in \ell^\infty(\cZ)\), our first objective is to control the coarse-graining bias
\begin{equation}
    \label{eq:cg_bias}
    \xi_\star\eta_t^X(f) - \eta_t^Z(f)
\end{equation}
as a function of the time variable \(t\).
The flow~\(\eta_t^Z\) is the non-linear flow~\eqref{eq:non_linear_flow} associated with the Markovian dynamics~\(Z\). We say it is a non-linear Markovian flow. The flow~\(\xi_\star \eta_t^X\) is the pushforward of the non-linear Markovian flow~\(\eta_t^X\), but is not itself Markovian.
We say it is a non-linear non-Markovian flow.

Given a measurable space~\((E,\cE)\) and any two probability measures~\(\mu,\nu\in\cP(E,\cE)\), we recall their distance in total variation is given by
\begin{equation}
    \label{eq:tv_distance}
    \d_{\rm TV}(\mu,\nu)=2\,\underset{A\in\cE}{\sup}\,\left|\nu(A)-\mu(A)\right|.
\end{equation}
If~\(\mu\) and~\(\nu\) are dominated by a common \(\sigma\)-finite measure~\(\lambda\in\cM(\cE)\), the total variation distance is simply the~\(L^1(\lambda)\)-distance of their densities with respect to~\(\lambda\). Moreover, we recall the dual formula
\begin{equation}
    \label{eq:tv_dual_formula}
    \d_{\mathrm{TV}}(\mu,\nu) = \underset{\|f\|_{B^{\infty}(E)}\leq 1}{\sup}\, \mu(f)-\nu(f),
\end{equation}
where~\(B^\infty(E)\) denotes the set of real-valued bounded~\((E,\cE)\)-measurable functions.

We estimate~\eqref{eq:cg_bias} using backward numerical analysis, which studies approximate solutions of exact problems by comparing them to exact solutions of approximate problems. The central idea is the definition of well-chosen probability flows~\eqref{eq:individual_error_process} encoding the coarse-graining error in a convenient way.
These definitions are inspired by a recent construction in the literature on particle approximations of non-linear semigroups, the backward error process from~\cite{JR24}.

We first introduce a family of probability measures on the level sets or \emph{fibers} of \(\xi\)
\begin{equation}
    \label{eq:fiber_measures}
    \cF := \left(\pi_z\right)_{z\in\cZ},\qquad \pi_z\in\cP\left(\xi^{-1}\{z\}\right),
\end{equation}
and define associated operators
\begin{equation}
    \label{eq:fiber_operators}
    \Pi^Z f(z) = \pi_z(f),\qquad \Pi^X f(x) = \pi_{\xi(x)}(f) = \xi^\star\Pi^Z f(x).
\end{equation}
The operator \(\Pi^Z\) sends the observable \(f\) on \(\cX\) to a fiber-averaged observable on \(\cZ\), which is then pulled back by \(\xi^\star\) to the space of fiber-wise constant observables on \(\cX\), giving~$\Pi^X f$.
The following relations are easy to check:
\begin{equation}
    \label{eq:projector_identities}
    \Pi^Z \xi^\star = \id,\qquad \Pi^X\xi^\star = \xi^\star,
\end{equation}
and in particular, \(\Pi^X\Pi^X=\xi^\star(\Pi^Z\xi^\star)\Pi^Z=\Pi^X\) is a projector.

The central objects in our framework are the following time-dependent \emph{backward error measures}.
For each \(Y\in\{X,Z\}\), we set
\begin{equation}
    \label{eq:individual_error_process}
    \psi_{t,T}^Y(g) = \frac{\eta_t^Y\left(\Pi^Y Q_{T-t}^X g \right)}{\eta_t^Y\left(\Pi^Y Q_{T-t}^X \1_{\cX} \right)} = \frac{\eta_0^Y\left(Q_t^Y\Pi^Y Q_{T-t}^X g \right)}{\eta_0^Y\left(Q_t^Y\Pi^Y Q_{T-t}^X \1_{\cX} \right)}.
\end{equation}
These define probability measure-valued flows in \(\cP(\cX)\): the exact \(Q^X\)-evolution is applied to \(g\) on \([t,T]\), preceded by a ``coarse-grained evolution'' \(Q_t^Y\Pi^Y\) on \([0,t]\). This linear evolution is normalized to ensure this approximate flow remains in \(\cP(\cX)\).

The usefulness of the error measures~\eqref{eq:individual_error_process} is clarified by the following reformulation of the bias~\eqref{eq:cg_bias}.
\begin{proposition}
    \label{prop:bias_integral}
    Fix \(T>0\), and assume that the initial conditions for \(X\) and \(Z\) are compatible, in the sense that \({\eta_0^Z = \xi_\star\eta_0^X}\).

    Then,
    \begin{equation}
        \label{eq:bias_integral}
        \begin{aligned}
            \xi_\star\eta_T^X(f) - \eta_T^Z(f) & = \int_0^T \partial_t\left[\psi^X_{t,T}-\psi^Z_{t,T}\right](\xi^\star f)\,\d t                                                                                        \\
                                               & = \int_0^T \, \frac{\eta_t^X\left(A^X Q_{T-t}^X\left(\xi^\star f-\psi_{t,T}^X(\xi^\star f)\1_{\cX}\right)\right)}{\eta_t^X\left(\Pi^X Q_{T-t}^X\1_{\cX}\right)}\,\d t \\
                                               & \quad -\int_0^T\frac{\eta_t^Z\left(A^Z Q_{T-t}^X\left(\xi^\star f-\psi_{t,T}^Z(\xi^\star f)\1_{\cX}\right)\right)}{\eta_t^Z\left(\Pi^Z Q_{T-t}^X\1_{\cX}\right)}\d t,
        \end{aligned}
    \end{equation}
    where the operators \(A^X\) and \(A^Z\) are defined by
    \begin{equation}
        \label{eq:defect_operators}
        A^X = \cL^X\Pi^X-\Pi^X\cL^X,\qquad A^Z = \cL^Z\Pi^Z - \Pi^Z\cL^X.
    \end{equation}
\end{proposition}
\begin{proof}
    First, setting \(t=T\) in~\eqref{eq:individual_error_process} gives
    \begin{equation}
        \psi_{T,T}^Y(\xi^\star f) = \frac{\eta_T^Y(\Pi^Y \xi^\star f)}{\eta_T^Y(\Pi^Y\1_{\cX})}.
    \end{equation}
    For \(Y=X\), this gives, using \(\Pi^X \xi^\star f = \xi^\star f\) and \(\1_{\cX}=\xi^\star\1_{\cZ}\)
    \begin{equation}
        \label{eq:final_point_error_X}
        \psi_{T,T}^X(\xi^\star f) = \frac{\eta_T^X(\xi^\star f)}{\eta_T^X(\1_{\cX})} = \xi_\star\eta_T^X(f),
    \end{equation}
    since \(\eta_T^X\in\cP(\cX)\). For \(Y=Z\), using \(\Pi^Z\xi^\star=\id\), we obtain
    \begin{equation}
        \label{eq:final_point_error_Z}
        \psi_{T,T}^Z(\xi^\star f) = \frac{\eta_T^Z(f)}{\eta_T^Z(\1_{\cZ})} = \eta_T^Z(f).
    \end{equation}

    On the other hand, setting \(t=0\) in~\eqref{eq:individual_error_process} gives
    \begin{equation}
        \label{eq:initial_point_error_match}
        \begin{aligned}
            \psi_{0,T}^X(\xi^\star f) & = \frac{\eta_0^X\left(\Pi^X Q_T^X \xi^\star f\right)}{\eta_0^X\left(\Pi^X Q_T^X \xi^\star \1_{\cZ}\right)}                  \\
                                      & = \frac{\eta_0^X\left(\xi^\star\Pi^Z Q_T^X \xi^\star f\right)}{\eta_0^X\left(\xi^\star\Pi^Z Q_T^X \xi^\star\1_{\cZ}\right)} \\
                                      & = \frac{\xi_\star\eta_0^X\left(\Pi^Z Q_T^X \xi^\star f\right)}{\xi_\star\eta_0^X\left(\Pi^Z Q_T^X \xi^\star\1_{\cZ}\right)} \\
                                      & = \psi_{0,T}^Z(\xi^\star f),
        \end{aligned}
    \end{equation}
    using the compatibility condition~\(\eta^Z_0 = \xi_\star\eta^X_0\) in the last line.

    The first equality in~\eqref{eq:bias_integral} then follows from~\eqref{eq:final_point_error_X},~\eqref{eq:final_point_error_Z},~\eqref{eq:initial_point_error_match} and the fundamental theorem of calculus.

    To obtain the second equality, we use the Feynman--Kac formula~\eqref{eq:feynman_kac} twice (once with \(Y\) and once with \(X\)), to compute, for all \(t\in (0,T)\) and \(g\in \ell^\infty(\cX)\):
    \begin{equation}
        \label{eq:general_computation}
        \partial_t \eta_0^Y\left(Q_t^Y\Pi^Y Q_{T-t}^X g\right) = \eta_0^Y\left(Q_t^Y\left[\cL^Y\Pi^Y-\Pi^Y\cL^X\right]Q_{T-t}^X g\right).
    \end{equation}

    By the product rule, this further gives, after a direct computation,
    \begin{equation}
        \label{eq:error_measure_time_derivative}
        \partial_t \psi_{t,T}^Y(g) = \frac{\eta_t^Y\left(\left[\cL^Y\Pi^Y-\Pi^Y\cL^X\right]Q_{T-t}^X\left(g-\psi_{t,T}^Y(g)\1_{\cX}\right)\right)}{\eta_t^Y\left(\Pi^Y Q_{T-t}^X\1_{\cX}\right)}.
    \end{equation}
    Applying this identity to \(Y=X,Z\) and \(g=\xi^\star f\), integrating over \(t\in(0,T)\), we obtain the second equality in~\eqref{eq:bias_integral}.
\end{proof}

We will use this integral formulation to obtain bounds on the coarse-graining bias~\eqref{eq:cg_bias} in Section~\ref{sec:proof_cg}.

The commutator~\(A^X = [\cL^X,\Pi^X]\) is a closure defect, measuring how the evolution of a fiber-wise constant observable deviates from being fiber-wise constant under the dynamics of \(X\).
Consequently, the contribution of the term involving \(A^X\) is related to the intrinsic non-Markovianity of \(\xi(X)\), and indeed is independent of the definition of the surrogate process \(Z\).

The operator~\(A^Z = \cL^Z\Pi^Z-\Pi^Z\cL^X\) involves the quality of~\(Z\). Viewing \(\Pi^Z\) as a Markov kernel \(\cZ\to\cP(\cX)\) via the identification
\begin{equation}
    \Pi^Z(z,\d x) = \pi_z(\d x),
\end{equation}
and denoting the dual action of~\(\cL^Y\) on probability measures with a right-multiplication, the exact case \(A^Z=0\) corresponds to the algebraic identity
\begin{equation}
    \cL^Z\Pi^Z = \Pi^Z\cL^X,
\end{equation}
which is known in the probabilistic literature as an \emph{intertwining relation}. Intertwining~\cite{RP81,DF90} is a powerful tool in the analysis of complex Markov processes, allowing one to prove various properties via transfer from equivalent but simpler models.
We understand \(A^Z\) as measuring an intertwining defect.

We finally stress that, up to this point, the definition of the fiber measures~\(\cF\) (and therefore of~\(\Pi^X\) and~\(\Pi^Z\)) has not been fixed. In fact, since the bias~\eqref{eq:cg_bias} we wish to control is independent of these objects, \(\cF\) should be seen as a parameter in the proofs.
A specific choice, with several interesting properties, will be fixed in the remainder of this work, but one could possibly optimize our arguments with respect to~\(\cF\) in specific situations.

\subsection{A canonical choice of~\(Z\)}
We now introduce a specific modeling choice, specifying both the choice of the fiber measures \(\cF\) and of the surrogate process \(Z\).
This choice has several interesting consequences, listed in Proposition~\ref{prop:perfect_model} below.

\begin{definition}[Canonical model]
    \label{def:perfect_model}

    \begin{itemize}
        \item[i)]{
                  The family \(\cF\) of probability measures is defined as a disintegration of \(\nu^X\) with respect to \(\xi\), that is
                  \begin{equation}
                      \label{eq:nu_disintegration}
                      \int_{\cX}f(x)\nu^X(\d x) = \int_{\cZ}\int_{\xi^{-1}(z)}f(x)\pi_z(\d x)\xi_\star\nu^X(\d z)
                  \end{equation}
                  for any \(f\in L^1(\nu^X)\).
              }
        \item[ii)]{
                  The surrogate process~\(Z\) is the killed jump process generated by the coarse-grained generator
                  \begin{equation}
                      \label{eq:cg_generator}
                      \cL^Z = \Pi^Z \cL^X\xi^\star.
                  \end{equation}
              }

    \end{itemize}
\end{definition}
One can show that the operator~\(\cL^Z\) is the generator of a conditionally irreducible jump process, which can be understood as a fiber-to-fiber jump process, where rates are given by fiber-averages of the~$\cL^X$-rates under the QSD~\(\nu^X\).
This property relies only on the positivity of~$\nu^X$, and would apply to any choice of~$\cF$ defined by disintegration of a strictly positive measure. We refer to the proof in Appendix~\ref{sec:proof_perturbation_theory} where a very similar property is proven.
The definition~\eqref{eq:cg_generator} is closely related to classical definitions of effective dynamics (see~\cite{LL10,ZHS16}) where Markovian approximations are defined in terms of conditional averages with respect to the global invariant measure.
Here, we perform an analogous construction with respect to the local equilibrium measure~$\nu^X$.

Many simplifications come from this choice of~$Z$, as the following proposition shows.
\begin{proposition}[Consequences of Definition~\ref{def:perfect_model}]
    \label{prop:perfect_model}
    Suppose that \(\cF\) is defined as in item i) from Definition~\ref{def:perfect_model}. Then the following properties are satisfied.
    \begin{itemize}
        \item[i)]{
                  The QSD is invariant under fiber-averaging:
                  \begin{equation}
                      \label{eq:qsd_projection_invariance_X}
                      \forall\,f \in L^1(\nu^X),\qquad\nu^X(\Pi^X f) = \nu^X(f).
                  \end{equation}
              }
        \item[ii)]{
                  The closure defect is QSD-centered:
                  \begin{equation}
                      \label{eq:centered_defects_X}
                      P^X A^X = A^X.
                  \end{equation}
              }
    \end{itemize}

    Suppose furthermore that \(\cL^Z\) is defined as in item ii) from Definition~\ref{def:perfect_model}.
    \begin{itemize}
        \item[iii)]{
                  The \(Z\)-counterparts to~\eqref{eq:qsd_projection_invariance_X} and~\eqref{eq:centered_defects_X} hold:
                  \begin{equation}
                      \label{eq:Z_counterparts}
                      \forall\, f\in L^1(\nu^X),\quad\nu^Z(\Pi^Z f) = \nu^X(f),\qquad P^Z A^Z = A^Z.
                  \end{equation}
              }
        \item[iv)]{
                  The QSD for \(Z\) is explicitly given by
                  \begin{equation}
                      \label{eq:explicit_qsd}
                      \nu^Z = \xi_\star\nu^X,\qquad \lambda^Z=\lambda^X.
                  \end{equation}}
        \item[v)]{The intertwining defect is controlled by the closure defect:
                  \begin{equation}
                      \label{eq:intertwining_defect_control}
                      A^Z = \Pi^Z A^X.
                  \end{equation}

              }
    \end{itemize}

\end{proposition}
\begin{proof}
    We first show~\eqref{eq:qsd_projection_invariance_X}, computing, since \(\Pi^X f\) is constant on fibers:
    \begin{equation}
        \begin{aligned}
            \int_{\cX} \Pi^X f(x)\nu^X(\d x) & = \int_{\cZ}\int_{\xi^{-1}(z)}\Pi^X f(x)\,\pi_z(\d x)\,\xi_\star\nu^X(\d z) \\
                                             & =\int_{\cZ}\int_{\xi^{-1}(z)}\pi_z(f)\,\pi_z(\d x)\,\xi_\star\nu^X(\d z)    \\
                                             & =\int_{\cZ}\pi_z(f)\,\xi_\star\nu^X(\d z)                                   \\
                                             & = \int_{\cZ}\int_{\xi^{-1}(z)}f(x)\,\pi_z(\d x)\,\xi_\star\nu^X(\d z)       \\
                                             & =\nu^X(f).
        \end{aligned}
    \end{equation}

    We now show~\eqref{eq:centered_defects_X}. For any \(Y\in\{X,Z\}\),
    \begin{equation}
        \label{eq:perfect_model_useful_identity}
        \nu^Y A^Y = \nu^Y\left[\cL^Y \Pi^Y -\Pi^Y \cL^X\right] = -\lambda^Y \nu^Y \Pi^Y -\nu^Y\Pi^Y\cL^X = -\lambda^Y \nu^X -\nu^X\cL^X,
    \end{equation}
    from which~\eqref{eq:centered_defects_X} follows immediately, as well as the second identity in~\eqref{eq:Z_counterparts} once~\eqref{eq:explicit_qsd} is established.

    We now assume that~\(Z\) is defined as in~\eqref{eq:cg_generator}, and proceed to show~\eqref{eq:explicit_qsd}. Let \(f\in \ell^\infty(\cZ)\). We compute
    \begin{equation}
        \begin{aligned}
            \xi_\star\nu^X (\cL^Z f) & = \nu^X(\xi^\star \Pi^Z \cL^X\xi^\star f) \\
                                     & = \nu^X(\Pi^X\cL^X\xi^\star f)            \\
                                     & = \nu^X(\cL^X \xi^\star f)                \\
                                     & = -\lambda^X\nu^X(\xi^\star f)            \\
                                     & = -\lambda^X \xi_\star \nu^X (f)
        \end{aligned}
    \end{equation}
    using the definition of \(\Pi^X\) in the second line,~\eqref{eq:qsd_projection_invariance_X} in the  third line, and the eigenequation~\eqref{eq:dual_relations} in the fourth line.
    This proves that \(\xi_\star\nu^X\) is a positive eigenmeasure for \(\cL^Z\), and is therefore equal to~\(\nu^Z\) by uniqueness in Proposition~\ref{prop:spectral_properties}, with common negative eigenvalue~\(-\lambda^X=-\lambda^Z\).

    The first identity in~\eqref{eq:Z_counterparts} then follows easily, since
    \begin{equation}
        \nu^Z(\Pi^Z f) = \xi_\star \nu^X(\Pi^Z f) = \nu^X(\xi^\star \Pi^Z f) = \nu^X(\Pi^X f) = \nu^X(f)
    \end{equation}
    by~\eqref{eq:qsd_projection_invariance_X} and~\eqref{eq:explicit_qsd}.

    Finally,~\eqref{eq:intertwining_defect_control} follows from a simple computation using the identities~\eqref{eq:fiber_operators} and~\eqref{eq:projector_identities}:
    \begin{equation}
        \begin{aligned}
            A^Z & = \cL^Z\Pi^Z - \Pi^Z\cL^X                    \\
                & =\Pi^Z\cL^X\xi^\star\Pi^Z - \Pi^Z\cL^X       \\
                & = \Pi^Z\cL^X\Pi^X - \Pi^Z\xi^\star\Pi^Z\cL^X \\
                & = \Pi^Z\cL^X\Pi^X -\Pi^Z\Pi^X\cL^X           \\
                & =\Pi^Z A^X.
        \end{aligned}
    \end{equation}
\end{proof}

\subsection{Statement of main results}

Our first main result is a control on the bias~\eqref{eq:cg_bias}, giving an approximation of the non-linear non-Markovian flow~\(\xi_\star \eta_t^X\) by the non-linear Markovian flow~\(\eta_t^Z\) associated with the canonical model~\(Z\) of Definition~\ref{def:perfect_model}.

\begin{theorem}
    \label{thm:cg_existential_result}
    Let \(\cF\) and \(Z\) be as in Definition~\ref{def:perfect_model}, and assume that a separation of timescales holds: there exist~\(C_\zeta,\zeta>0\) such that
    \begin{equation}
        \label{eq:separation_of_timescales}
        \forall\,t\geq 0,\qquad\left\|\e^{t(I-\Pi^X)\cL^X(I-\Pi^X)}(I-\Pi^X)\right\| \leq C_\zeta \e^{-\zeta t},
    \end{equation}
    with moreover \(\zeta>2(\lambda^X+\gamma^X)\), and that the compatibility condition \(\eta_0^Z = \xi_\star\eta_0^X\) holds.
     Recall  the definition of the closure defect~\(A^X=\cL^X\Pi^X - \Pi^X\cL^X\). It holds that
    \begin{itemize}
        \item[i)]{there exist explicit positive constants \(C\) and \(C'\)  depending on \(\eta_0^X\) but independent of \(T\) such that
                  \begin{equation}
                      \label{eq:tv_bound}
                      \forall\,T >0,\qquad\d_{\mathrm{TV}}\left(\xi_\star \eta_T^X, \eta_T^Z\right) \leq \frac{\|A^X\|}{\zeta}\left(C\e^{-\gamma^X T}+C'T\e^{-(\gamma^Z\land\gamma^X)T}\right),
                  \end{equation}
              }
        \item[ii)]{
                  we have the quasimode estimates
                  \begin{equation}
                      \label{eq:X_quasimode_estimate}
                      \|h^X - \Pi^X h^X\| \leq C_{\lambda}^X C_\zeta\frac{\|A^X\|}{\zeta - \lambda^X}
                  \end{equation}
                  and
                  \begin{equation}
                      \label{eq:Z_quasimode_estimate}
                      \|\Pi^Z h^X-h^Z\| \leq \frac{C_{\lambda}^X C_\zeta C_\gamma^Z}{\gamma^Z}\frac{\|A^X\|^2}{\zeta - \lambda^X},
                  \end{equation}
              }
        \item[iii)]{
                  the following two-sided estimate holds for all \(T>0\),
                  \begin{equation}
                      \label{eq:two_sided_error_estimate}
                      \left|\d_{\mathrm{TV}}(\xi_\star\eta_T^X,\xi_\star\nu^X)-\d_{\mathrm{TV}}(\eta_T^X,\nu^X)\right|\leq\frac{C_\lambda^X C_\gamma^XC_\zeta}{\eta_0^X(h^X)}\left(\e^{-(\zeta-\lambda^X) T} + 2\frac{\|A^X\|}{\zeta}\e^{-\gamma^X T}\right).
                  \end{equation}
              }
    \end{itemize}
\end{theorem}

Theorem~\ref{thm:cg_existential_result} has the following straightforward consequence.
\begin{corollary}
    \label{cor:diagnostic_bound}
    Under the Assumptions of Theorem~\ref{thm:cg_existential_result}, it holds, for all~\(T>0\),
    \begin{equation}
        \label{eq:diagnostic_bound}
        \begin{aligned}
            \left|\d_{\mathrm{TV}}(\eta_T^Z,\nu^Z)-\d_{\mathrm{TV}}(\eta_T^X,\nu^X)\right| & \leq\frac{C_\lambda^X C_\gamma^XC_\zeta}{\eta_0^X(h^X)}\left(\e^{-(\zeta-\lambda^X) T} + 2\frac{\|A^X\|}{\zeta}\e^{-\gamma^X T}\right) \\
                                                                                           & \qquad+\frac{\|A^X\|}\zeta\left(C\e^{-\gamma^X T}+C'T\e^{-(\gamma^Z\land\gamma^X)T}\right),
        \end{aligned}
    \end{equation}
    with the same constants~\(C,C'\) as in~\eqref{eq:tv_bound}.
\end{corollary}
\begin{proof}
    Write
    \begin{equation}
        \label{eq:triangle_inequality}
        \begin{aligned}
            \left|\d_{\mathrm{TV}}(\eta_t^Z,\nu^Z)-\d_{\mathrm{TV}}(\eta_t^X,\nu^X)\right| & \leq \left|\d_{\mathrm{TV}}(\eta_t^Z,\nu^Z)-\d_{\mathrm{TV}}(\xi_\star\eta_t^X,\xi_\star\nu^X)\right|    \\
                                                                                           & \qquad+\left|\d_{\mathrm{TV}}(\xi_\star\eta_t^X,\xi_\star\nu^X)-\d_{\mathrm{TV}}(\eta_t^X,\nu^X)\right|,
        \end{aligned}
    \end{equation}
    use item iv) in Proposition~\ref{prop:perfect_model} and a reverse triangle inequality to bound
    \begin{equation}
        \left|\d_{\mathrm{TV}}(\eta_t^Z,\nu^Z)-\d_{\mathrm{TV}}(\xi_\star\eta_t^X,\xi_\star\nu^X)\right|\leq \d_{\mathrm{TV}}(\eta_t^Z,\xi_\star\eta_t^X).
    \end{equation}
    The proof then follows immediately from the estimates~\eqref{eq:tv_bound} and~\eqref{eq:two_sided_error_estimate}.
\end{proof}

Before proving Theorem~\ref{thm:cg_existential_result}, let us make a few remarks on the nature of these results.

\begin{itemize}
    \item[\(\circ\)]{The estimate~\eqref{eq:diagnostic_bound} is relevant to quasi-stationarity detection, as it ensures that, in the regime where \(\|A^X\|/\zeta\) is much smaller than the prefactors, the convergence of \(\eta_t^Z\) towards \(\nu^Z\) provides an accurate diagnostic for the convergence of~$\eta_t^X$ towards the QSD.}
    \item[\(\circ\)]{The formulas we obtain for the prefactors \(C\) and \(C'\) in~\eqref{eq:tv_bound} and~\eqref{eq:diagnostic_bound} are rather complicated, which is why we provide simplified constants, given for convenience at the end of the proof of Theorem~\ref{thm:cg_existential_result} (see Equation~\eqref{eq:prefactors_expanded} below). Their dependence on \(\eta_0^X\) is completely explicit in terms of \(h^X\) and \(h^Z\), however, we have no a priori control on their relation to \(\zeta\). In the adiabatic regime considered in Section~\ref{sec:perturbation_theory}, we prove explicit scaling estimates for these constants and the parameter~\(\|A^X\|/\zeta\). In Theorem~\ref{thm:perturbation_theory}, we show, in the limit of some small parameter~\(\varepsilon\to 0\), that~\(\|A^X\|/\zeta = \bigo(\varepsilon)\), while all the other constants remain bounded uniformly in~\(\varepsilon\). In this regime, we can therefore obtain convergence of the probability flow~\(\eta^Z\) towards~\(\xi_\star\eta^X\) uniformly in time, at rate~\(\bigo(\varepsilon)\) in total-variation (see Corollary~\ref{cor:uniform_coarse_graining}). A similar quantitative-in-\(\varepsilon\) result holds for the diagnostic bound of Corollary~\ref{cor:diagnostic_bound}.}
    \item[\(\circ\)]{Using a straightforward argument based on a spectral decomposition and a triangle inequality, we can, instead of~\eqref{eq:tv_bound}, obtain an estimate of the form
              \begin{equation}
                  \label{eq:trivial_tv_bound}
                  \d_{\mathrm{TV}}\left(\xi_\star \eta_T^X, \eta_T^Z\right) \leq C''\e^{-\gamma^X T} + C'''\e^{-\gamma^Z T}.
              \end{equation}
              This is the sharper bound in the regime~\(T\gg \zeta/\|A^X\|\). The difference is that the bound~\eqref{eq:trivial_tv_bound} is unaware of any separation of timescale for~\(X\) or of the magnitude of the closure defect~\(A^X\). We therefore understand~\eqref{eq:tv_bound} as sacrificing relative accuracy in the long-time limit for accuracy in the transient phase of decorrelation, in which quasi-stationarity diagnostics are most needed. The upper bound~\eqref{eq:tv_bound} still converges to~\(0\) as~\(T\to+\infty\), and therefore preserves absolute accuracy in the long-time limit.
              Moreover, the trivial bound~\eqref{eq:trivial_tv_bound} gives no information in the adiabatic limit of Section~\ref{sec:perturbation_theory}, where orthogonal degrees of freedom quickly equilibrate at a fixed value of~$\xi$.}
    \item[\(\circ\)]{The assumption~\(\zeta>2(\lambda^X+\gamma^X)\) is not crucial, and the proof shows that it could be relaxed to~\(\zeta>\lambda^X+\gamma^X\), which says roughly that the orthogonal fiber-wise dynamics relax strictly faster than the non-linear semigroup. We understand this condition as a mild adiabatic condition. We keep the more restrictive condition that the orthogonal dynamics relax \emph{twice as fast}, in order to simplify the final estimates.}
    \item[\(\circ\)]{Checking whether a given pair \(X,\xi\) of dynamics and collective variable satisfies the assumptions of Theorem~\ref{thm:cg_existential_result} is typically infeasible for systems with many microstates (i.e. when~\(|\cX|\gg 1\)), as these involve quantitative spectral properties of the generator. For the specific case of complex molecular systems, assuming that \(\|A^X\|/\zeta\) is small will likely remain a working hypothesis in many cases for which a natural definition of \(\xi\) is available, either through domain expertise or computational methods.
    Practical choices of the collective variable~$\xi$ can nevertheless be inspired by the adiabaticity condition, if sufficient knowledge of the system is at hand.
    }
\end{itemize}

\subsection{Proof of Theorem~\ref{thm:cg_existential_result}}
\label{sec:proof_cg}

We now prove Theorem~\ref{thm:cg_existential_result}.
\begin{proof}[Proof of Theorem~\ref{thm:cg_existential_result}]
    For brevity, we will use the notations \(\Pi_0:=\Pi^X\), \(\Pi_1 := I-\Pi_0\), and \( A_{\alpha \beta}:=\Pi_\alpha A \Pi_\beta\) for any operator \(A\) on \(\ell^\infty(\cX)\).

    We often use the convolution bound
    \begin{equation}
        \label{eq:convolution_inequality}
        \forall a,b,T>0,\,a\neq b\qquad \int_0^T \e^{-a t}\e^{-b(T-t)}\,\d t = \frac{\e^{-b T}-\e^{-a T}}{a-b} \leq T \e^{-(a\land b)T},
    \end{equation}
    where the upper bound follows from the mean-value theorem and the convexity of \(x\mapsto \e^{-x T}\). The same inequality obviously holds for~$a=b$.

    We begin by showing the main estimate~\eqref{eq:tv_bound}. This proceeds in two steps. We first establish rough estimates on the error process, using only the principal eigenprojections, and an interesting ``self-correcting'' property of the error measures~\(\psi_{t,T}^Y(\xi^\star f)\), which implies faster decay than generic spectral bounds.
    The condition~\eqref{eq:separation_of_timescales} is then used in a second step in combination with these rough estimates to establish~\eqref{eq:tv_bound}, by splitting the semigroup into its action on slow and fast variables. The main tool~\eqref{eq:duhamel_estimate} used in this step is based on a Duhamel--Dyson splitting identity~\eqref{eq:duhamel_formula}.

    \subsubsection{Preliminary estimates and self-correcting property}

    The first step is to obtain a rough bound on the time derivatives
    \begin{equation}
        \label{eq:time_derivatives}
        \partial_t \psi_{t,T}^Y(\xi^\star f) = \frac{\eta_0^Y Q_t^Y A^Y Q_{T-t}^X r_{t,T}^Y(f)}{\eta_0^Y Q_t^Y \Pi^Y Q_{T-t}^X \1_{\cX}},
    \end{equation}
    where we denote the residual
    \begin{equation}
        \label{eq:residual}
        r_{t,T}^Y(f) := \xi^\star f  -\psi_{t,T}^Y(\xi^\star f)\1_{\cX}.
    \end{equation}

    We will use the following lower estimate
    \begin{equation}
        \label{eq:survival_bound}
        \begin{aligned}
            \eta_0^Y Q_t^Y \Pi^Y Q_{T-t}^X \1_{\cX} & \geq \frac1{\|h^X\|}\eta_0^Y Q_t^Y \Pi^Y Q_{T-t}^X h^X               \\
                                                    & \geq \frac{C_h^Y}{\|h^X\|}\e^{-\lambda^X(T-t)}\eta_0^Y(Q_t^Y h^Y)    \\
                                                    & = \frac{C_h^Y}{\|h^X\|}\e^{-\lambda^X(T-t)-\lambda^Y t}\eta_0^Y(h^Y) \\
                                                    & := C_1^Y \e^{-\lambda^X(T-t)-\lambda^Y t}
        \end{aligned}
    \end{equation}
    which uses the positivity of the operators \(Q_t^X\), \(Q_t^Y\) and \(\Pi^Y\), as well as the pointwise comparison \(\Pi^Y h^X \geq C_h^Y h^Y\), defining
    \begin{equation}
        \label{eq:pointwise_constant}
        C_h^Y := \underset{\cY}{\min} \frac{\Pi^Y h^X}{h^Y},
    \end{equation}
    which is well-defined since \(h^Y\) is positive on \(\cY\).

    Because of item iv) in Proposition~\ref{prop:perfect_model}, we actually have \(\lambda^X=\lambda^Y\), which simplifies the estimate~\eqref{eq:survival_bound}, but this does not turn out to be a crucial property.

    We then decompose spectrally
    \begin{equation}
        \label{eq:spectral_decomposition}
        \begin{aligned}
            \eta_0^Y Q_t^Y A^Y Q_{T-t}^X r_{t,T}^Y(f) & = \eta_0^Y Q_t^Y P^Y A^Y Q_{T-t}^X r_{t,T}^Y(f)                                                                 \\
                                                      & = \eta_0^Y Q_t^Y P^Y A^Y Q_{T-t}^X P^X r_{t,T}^Y(f) + \nu^X(r_{t,T}^Y(f)) \eta_0^Y Q_t^Y P^Y A^Y Q_{T-t}^X h^X,
        \end{aligned}
    \end{equation}
    using the identity \(A^Y = P^Y A^Y\), which is a consequence of the choice of canonical model, as shown in Proposition~\ref{prop:perfect_model} (items ii and iii).

    We estimate the terms in the decomposition as
    \begin{equation}
        \label{eq:numerator_sc_estimates}
        \begin{aligned}
            |\eta_0^Y Q_t^Y P^Y A^Y Q_{T-t}^X P^X r_{t,T}^Y(f)| & \leq C_\gamma^X C_\gamma^Y\|\eta_0^Y\|_{*} \e^{-(\lambda^Y+\gamma^Y)t}\|A^Y\|\e^{-(\lambda^X+\gamma^X)(T-t)}\|r_{t,T}^Y(f)\| \\
                                                                & \leq 2 C_\gamma^X C_\gamma^Y \e^{-(\lambda^Y+\gamma^Y)t}\|A^Y\|\e^{-(\lambda^X+\gamma^X)(T-t)}\|f\|                          \\
                                                                & := C_2^Y \e^{-\lambda^X (T-t)}\e^{-\lambda^Y t}\e^{-\gamma^Y t}\e^{-\gamma^X(T-t)}\|f\|                                      \\
            |\eta_0^Y Q_t^Y P^Y A^Y Q_{T-t}^X h^X|              & \leq C_\lambda^X C_\gamma^Y\e^{-(\lambda^Y+\gamma^Y)t}\|A^Y\|\e^{-\lambda^X(T-t)}\|h^X\|                                     \\
                                                                & := C_3^Y \e^{-\lambda^X (T-t)}\e^{-\lambda^Y t}\e^{-\gamma^Y t}
        \end{aligned}
    \end{equation}
    using \(\|\eta_0^Y\|_*=1\) in the second and third line, and \(\|r_{t,T}^Y(f)\|\leq 2\|f\|\) in the second line.

    In fact, we can recover faster decay of the principal eigenspace contribution by estimating the coefficient \(\nu^X(r_{t,T}^Y(f))\) appearing in~\eqref{eq:spectral_decomposition}. We begin with a computation
    \begin{equation}
        \begin{aligned}
            \nu^X(r_{t,T}^Y(f)) & = \nu^X(\xi^\star f - \psi_{t,T}^Y(\xi^\star f)\1_{\cX})                                                                                       \\
                                & = \nu^X(\xi^\star f)-\psi_{t,T}^Y(\xi^\star f)                                                                                                 \\
                                & = -\psi_{t,T}^Y(\xi^\star f - \nu^X(\xi^\star f)\1_{\cX})                                                                                      \\
                                & = -\psi_{t,T}^Y(P^X\left[\xi^\star f - \nu^X(\xi^\star f)\1_{\cX}\right])                                                                      \\
                                & = -\frac{\eta_0^Y Q_t^Y \Pi^Y Q_{T-t}^X P^X\left[\xi^\star f - \nu^X(\xi^\star f)\1_{\cX}\right]}{\eta_0^Y Q_t^Y \Pi^Y Q_{T-t}^X \1_{\cX}}     \\
                                & = -\frac{\eta_0^Y Q_t^Y P^Y\Pi^Y Q_{T-t}^X P^X\left[\xi^\star f - \nu^X(\xi^\star f)\1_{\cX}\right]}{\eta_0^Y Q_t^Y \Pi^Y Q_{T-t}^X \1_{\cX}},
        \end{aligned}
    \end{equation}
    using the fact that \(\psi_{t,T}^Y\) is a probability measure, and that \(\nu^X(\xi^\star f - \nu^X(\xi^\star f)\1_{\cX})=0\) to insert \(P^X\) in the fourth line, together with the computation
    \begin{equation}
        \nu^Y \Pi^Y Q_{T-t}^X P^X = \nu^Y \Pi^Y P^X Q_{T-t}^X = \nu^X P^X Q_{T-t}^X = 0
    \end{equation}
    to insert \(P^Y\) in the last line. We used items i and iii from Proposition~\ref{prop:perfect_model} in the second-to-last equality.

    Taking operator norms gives the estimate
    \begin{equation}
        \begin{aligned}
             & |\eta_0^Y Q_t^Y P^Y\Pi^Y Q_{T-t}^X P^X\left[\xi^\star f - \nu^X(\xi^\star f)\1_{\cX}\right]|                                                              \\
             & \leq C_\gamma^XC_\gamma^Y\|\eta_0^Y\|_{*}\|\Pi^Y\|\e^{-(\lambda^Y +\gamma^Y)t}\e^{-(\lambda^X+\gamma^X)(T-t)}\|\xi^\star f - \nu^X(\xi^\star f)\1_{\cX}\|
        \end{aligned}
    \end{equation}
    which combined with the lower bound~\eqref{eq:survival_bound} and the fact that \(\|\eta_0^Y\|_*=1\) gives
    \begin{equation}
        \label{eq:nu_component_estimate}
        \begin{aligned}
            |\nu^X(r_{t,T}^Y(f))| & \leq  \frac{C_\gamma^XC_\gamma^Y\|\eta_0^Y\|_{*}\|\Pi^Y\|}{C_1^Y}\e^{-\gamma^Y t}\e^{-\gamma^X(T-t)} \|\xi^\star f - \nu^X(\xi^\star f)\1_{\cX}\| \\
                                  & \leq 2\frac{C_\gamma^XC_\gamma^Y\|\Pi^Y\|}{C_1^Y}\e^{-\gamma^Y t}\e^{-\gamma^X(T-t)} \|f\|                                                        \\
                                  & := C_4^Y\e^{-\gamma^Y t}\e^{-\gamma^X(T-t)} \|f\|
        \end{aligned}
    \end{equation}
    where we used \(\|\xi^\star f - \nu^X(\xi^\star f)\1_{\cX}\|\leq 2\|f\|\), since \(\nu^X\) is a probability measure and \(\|\xi^\star f\|=\|f\|\).

    Combining~\eqref{eq:nu_component_estimate} and~\eqref{eq:numerator_sc_estimates}, using a triangle inequality, and dividing by the lower bound~\eqref{eq:survival_bound}, we arrive at the pointwise estimate in time on~\eqref{eq:time_derivatives}:
    \begin{equation}
        \label{eq:time_derivatives_first_estimates}
        \begin{aligned}
             & |\partial_t \psi_{t,T}^Y(\xi^\star f)| \leq \left(\frac{C_2^Y}{C_1^Y} \e^{-\gamma^Y t} + \frac{C_3^Y C_4^Y}{C_1^Y}\e^{-2\gamma^Y t}\right)\e^{-\gamma^X(T-t)}\|f\|.
        \end{aligned}
    \end{equation}

    It is notable that the contribution from the \(h^X\)-component of \(r_{t,T}^Y(f)\) actually decays at a faster rate than that of the \(P^X\)-invariant component, which is contrary to what occurs for generic time-independent observables. This reflects a self-correcting property of the error measure \(\psi_{t,T}^Y\).

    We also record for later use the following estimate, which is obtained along similar lines
    \begin{equation}
        \label{eq:survival_probability_derivative_bound}
        \begin{aligned}
            \left|\partial_t\log\left(\eta_0^Y Q_t^Y \Pi^Y Q_{T-t}^X \1_{\cX}\right)\right| & = \frac{|\eta_0^Y Q_t^Y P^Y A^Y Q_{T-t}^X \1_{\cX}|}{|\eta_0^Y Q_t^Y \Pi^Y Q_{T-t}^X \1_{\cX}|} \\
                                                                                            & \leq \frac1{C_1^Y} C_\lambda^X C_\gamma^Y \|A^Y\|\|\1_{\cX}\| \e^{-\gamma^Y t}                  \\
                                                                                            & =\frac1{C_1^Y} C_\lambda^X C_\gamma^Y \|A^Y\| \e^{-\gamma^Y t}                                  \\
                                                                                            & := C_5^Y \e^{-\gamma^Y t}
        \end{aligned}
    \end{equation}
    without resorting to a spectral decomposition (since the constant-in-time observable \(\1_{\cX}\) has a non-zero \(h^X\)-component).
    \subsubsection{Slow-fast splitting and proof of~\eqref{eq:tv_bound}}
    We now use the timescale separation condition~\eqref{eq:separation_of_timescales} together with the previous bounds to estimate the bias as in~\eqref{eq:bias_integral}, by considering the decomposition
    \begin{equation}
        \label{eq:fiber_decomposition}
        \ell^\infty(\cX) = \Pi_0 \ell^\infty(\cX) \oplus \Pi_1 \ell^\infty(\cX)
    \end{equation}
    onto the subspaces~\(\Pi_0 \ell^\infty(\cX)\) and~\(\Pi_1 \ell^\infty(\cX)\) of slow and fast variables, respectively.

    The basic idea is to use the splitting
    \begin{equation}
        \partial_t\left(\Pi_1 Q_t^X v\right) = \cL^X_{11} Q_t^X v + \cL^X_{10} Q_t^X v = \cL^X_{11} \Pi_1 Q_t^X v + \cL^X_{10} Q_t^X v,
    \end{equation}
    which gives the Duhamel--Dyson formula
    \begin{equation}
        \label{eq:duhamel_formula}
        \forall\, v\in \ell^\infty(\cX),\qquad  \Pi_1 Q_t^X v = \e^{t\cL^X_{11}}\Pi_1 v + \int_0^t \e^{(t-s)\cL^X_{11}}\cL^X_{10} Q_s^X v\,\d s.
    \end{equation}
    Such formulas are fundamental in the theory of coarse-grained dynamical systems, in particular for the Mori--Zwanzig formalism.

    When it happens that~\(\Pi_1 v =0\), we can estimate, with the spectral decomposition \(v = P^X v + \nu^X(v)h^X\),
    \begin{equation}
        \label{eq:duhamel_estimate}
        \begin{aligned}
            \|\Pi_1 Q_t^X v\| & \leq C_{\gamma}^X C_{\zeta}\int_0^t \e^{-\zeta (t-s)}\|\cL^X_{10}\|\e^{-(\lambda^X+\gamma^X) s}\|P^X v\|\, \d s                                                                                                                      \\
                              & \quad + C_{\lambda}^X C_{\zeta}\int_0^t \e^{-\zeta (t-s)}\|\cL^X_{10}\|\e^{-\lambda^X s}|\nu^X(v)|\|h^X\|\, \d s                                                                                                                     \\
                              & =C_\gamma^X C_\zeta \|\cL_{10}^X\|\frac{\e^{-(\lambda^X+\gamma^X)t}-\e^{-\zeta t}}{\zeta -\lambda^X-\gamma^X}\|P^X v\| + C_\lambda^X C_\zeta \|\cL_{10}^X\|\|h^X\|\frac{\e^{-\lambda^X t}-\e^{-\zeta t}}{\zeta -\lambda^X}|\nu^X(v)| \\
                              & \leq 2C_\gamma^XC_\zeta \|\cL_{10}^X\|\frac{\e^{-(\lambda^X+\gamma^X)t}}{\zeta}\|P^X v\| + 2C_\lambda^X C_\zeta\|\cL^X_{10}\|\|h^X\|\frac{\e^{-\lambda^X t}}{\zeta}|\nu^X(v)|                                                        \\
                              & := C_6^X\frac{\e^{-(\lambda^X+\gamma^X)t}}{\zeta}\|P^X v\| + C_6^{'X}\frac{\e^{-\lambda^X t}}{\zeta}|\nu^X(v)|,
        \end{aligned}
    \end{equation}
    where we use \(\zeta>2(\lambda^X+\gamma^X)\) (note here that a similar estimate would hold assuming only \(\zeta > (1+\alpha)(\lambda^X+\gamma^X)\) for some \(\alpha>0\), so the assumption~\(\zeta>2(\lambda^X+\gamma^X)\) could be relaxed to~\(\zeta>(\lambda^X+\gamma^X)\)).

    To obtain pointwise bounds on \(\partial_t \psi_{t,T}\), we start by noticing an algebraic consequence of Definition~\ref{def:perfect_model},
    \begin{equation}
        \label{eq:AZ_fast}
        \begin{aligned}
            A^Z \Pi_0 & = (\cL^Z \Pi^Z - \Pi^Z \cL^X)\Pi_0                                                    \\
                      & = \cL^Z \left(\Pi^Z \xi^\star\right) \Pi^Z - \left(\Pi^Z \cL^X \xi^\star\right) \Pi^Z \\
                      & = \cL^Z \Pi^Z - \cL^Z \Pi^Z = 0,
        \end{aligned}
    \end{equation}
    which allows us to write
    \begin{equation}
        \partial_{t}\psi_{t,T}^Z(\xi^\star f) = \frac{\eta_0^Z Q_t^Z P^Z A^Z \Pi_1 Q_{T-t}^X r_{t,T}^Z(f)}{\eta_0^Z Q_t^Z \Pi^Z Q_{T-t}^X \1_{\cX}}.
    \end{equation}

    We estimate, using~\eqref{eq:duhamel_estimate} and~\eqref{eq:nu_component_estimate}, and noting that~\(\Pi_1 r_{t,T}^Y(f) = 0\) (since \(\1_{\cX}\) and \(\xi^\star f\) are both constant on fibers),
    \begin{equation}
        \label{eq:residual_duhamel_estimate}
        \begin{aligned}
             & \|\Pi_1 Q_{T-t}^X r_{t,T}^Z(f)\|  \leq C_6^X\frac{\e^{-(\lambda^X+\gamma^X)(T-t)}}{\zeta}\|P^X r_{t,T}^Z(f)\| + C_6^{'X}\frac{\e^{-\lambda^X (T-t)}}{\zeta}|\nu^X(r_{t,T}^Z(f))| \\
             & \leq 2C_6^X\frac{\e^{-(\lambda^X+\gamma^X)(T-t)}}{\zeta}\|P^X\| \|f\| + C_6^{'X}C_4^Z\frac{\e^{-\lambda^X (T-t)}}{\zeta}\e^{-\gamma^Z t}\e^{-\gamma^X(T-t)} \|f\|                \\
             & := \left(C_7^Z\frac{\e^{-(\lambda^X+\gamma^X)(T-t)}}{\zeta} + C_7^{'Z}\frac{\e^{-(\lambda^X+\gamma^X)(T-t)-\gamma^Z t}}{\zeta}\right)\|f\|.
        \end{aligned}
    \end{equation}

    This gives
    \begin{equation}
        \label{eq:slow_fast_Z_component}
        \begin{aligned}
             & |\eta_0^Z Q_t^Z P^Z A^Z \Pi_1 Q_{T-t}^X r_{t,T}^Z(f)| \leq \|Q_t^Z P^Z\|\|A^Z\|\|\Pi_1 Q_{T-t}^X r_{t,T}^Z(f)\|                                                                                 \\
             & \qquad\leq C_\gamma^Z \e^{-(\lambda^Z+\gamma^Z)t}\|A^Z\|\left(C_7^Z\frac{\e^{-(\lambda^X+\gamma^X)(T-t)}}{\zeta} + C_7^{'Z}\frac{\e^{-(\lambda^X+\gamma^X)(T-t)-\gamma^Z t}}{\zeta}\right)\|f\| \\
             & := \e^{-\lambda^Z t - \lambda^X(T-t)}\left(C_8^Z\frac{\e^{-\gamma^X(T-t)}}{\zeta} + C_8^{'Z}\frac{\e^{-\gamma^X (T-t)-\gamma^Z t}}{\zeta}\right)\e^{-\gamma^Z t}\|f\|.
        \end{aligned}
    \end{equation}

    Integrating on \([0,T]\), we find, after dividing the estimate~\eqref{eq:slow_fast_Z_component} by~\eqref{eq:survival_bound},
    \begin{equation}
        \label{eq:Z_contribution_final_bound}
        \begin{aligned}
             & \left|\int_0^T \partial_{t}\psi_{t,T}^Z(\xi^\star f)\,\d t\right|  \leq \int_0^T  |\partial_{t}\psi_{t,T}^Z(\xi^\star f)|\,\d t                                                          \\
             & \leq \frac1{C_1^Z}\int_0^T\left(C_8^Z\frac{\e^{-\gamma^X(T-t)}}{\zeta} + C_8^{'Z}\frac{\e^{-\gamma^X (T-t)-\gamma^Z t}}{\zeta}\right)\e^{-\gamma^Z t}\,\d t\|f\|                                        \\
            & \leq \frac{C_8^Z +  C_8^{'Z}}{\zeta C_1^Z}\int_0^T\e^{-\gamma^X(T-t)-\gamma^Z t}\,\d t\|f\|                                        \\
             & \leq\frac{C_8^Z +  C_8^{'Z}}{\zeta C_1^Z} T\e^{-(\gamma^X\land \gamma^Z)T}\|f\|,
        \end{aligned}
    \end{equation}
    using~\eqref{eq:convolution_inequality} in the final line.

    Controlling the contribution of the error measure \(\psi_{t,T}^X\) requires more care. We first decompose
    \begin{equation}
        \label{eq:error_measure_derivative_X_component}
        \partial_{t}\psi_{t,T}^X(\xi^\star f) = \frac{\eta_0^X Q_t^X P^X A^X (\Pi_1+\Pi_0) Q_{T-t}^X r_{t,T}^X(f)}{\eta_0^X Q_t^X \Pi_0 Q_{T-t}^X \1_{\cX}}.
    \end{equation}
    To exploit the separation of timescales property, we first eliminate the contribution of~$\Pi_0$ in the numerator.
    To this end, we introduce the linear functional
    \begin{equation}
        \label{eq:fast_propagator}
        \Phi(t,v) = \eta_0^X Q_t^X\Pi_1 Q_{T-t}^X v,\qquad \forall\, (t,v)\in [0,T] \times \ell^\infty(\cX),
    \end{equation}
    which is the analog of the unnormalized flow~$\eta_0^X Q_t^X\Pi_0 Q_{T-t}^X v$  defining~$\psi_{t,T}^X(v)$, with the substitution~$\Pi_0\leftarrow\Pi_1$. 

    Differentiating with respect to the time variable, we get
    \begin{equation}
        \partial_t \Phi(t,v) = \eta_0^X Q_t^X \left[\cL^X,\Pi_1\right]Q_{T-t}^X v = -\eta_0^X Q_t^X A^X Q_{T-t}^X v
    \end{equation}

    Splitting~$A^X = A^X\Pi_0+A^X\Pi_1$, it follows that
    \begin{equation}
        \label{eq:phi_derivative}
        \begin{aligned}
            \eta_0^X Q_t^X A^X \Pi_0 Q_{T-t}^X r_{t,T}^X(f) & = -\eta_0^X Q_t^X A^X \Pi_1 Q_{T-t}^X r_{t,T}^X(f) - \partial_t \Phi(t,r_{t,T}^X(f))      \\
                                                            & = -\eta_0^X Q_t^X A^X \Pi_1 Q_{T-t}^X r_{t,T}^X(f) -\frac{\d}{\d t}  \Phi(t,r_{t,T}^X(f)) \\
                                                            & \qquad+ D_v \Phi(t,r_{t,T}^X(f))\left(\frac{\d}{\d t}r_{t,T}^X(f)\right)                  \\
                                                            & = -\eta_0^X Q_t^X A^X \Pi_1 Q_{T-t}^X r_{t,T}^X(f) -\frac{\d}{\d t}  \Phi(t,r_{t,T}^X(f)) \\
                                                            & \qquad - \Phi(t,\1_{\cX})\partial_{t}\psi_{t,T}^X(\xi^\star f),                           \\
        \end{aligned}
    \end{equation}
    where \(\partial_t\Phi(\cdot,\cdot)\) denotes the partial time derivative and \(\frac{\d}{\d t}\Phi(\cdot,\cdot)\) denotes the total time derivative.
    Therefore, the term involving~$\Pi_0$ can be eliminated from the numerator in~\eqref{eq:error_measure_derivative_X_component}, leaving terms containing~$\Pi_1$ through the definition of~$\Phi$.

    Substituting~\eqref{eq:phi_derivative} into~\eqref{eq:error_measure_derivative_X_component} yields
    \begin{equation}
        \partial_{t}\psi_{t,T}^X(\xi^\star f) = -\left(\frac{\frac{\d}{\d t}  \Phi(t,r_{t,T}^X(f)) + \Phi(t,\1_{\cX})\partial_{t}\psi_{t,T}^X(\xi^\star f)}{\eta_0^X Q_t^X \Pi_0 Q_{T-t}^X \1_{\cX}}\right).
    \end{equation}

    The first term is handled using integration by parts
    \begin{equation}
        \label{eq:ibp_error_measure}
        \begin{aligned}
             & \int_0^T \frac1{\eta_0^X Q_t^X \Pi_0 Q_{T-t}^X \1_{\cX}}\left(\frac{\d}{\d t}  \Phi(t,r_{t,T}^X(f))\right)\,\d t                                                                                                                                                                              \\
             & = \frac{\Phi(T,r_{T,T}^X(f))}{\eta_0^X Q_T^X \Pi_0 \1_{\cX}} - \frac{\Phi(0,r_{0,T}^X(f))}{\eta_0^X\Pi_0 Q_{T}^X \1_{\cX}} +\int_0^T \frac{\Phi(t,r_{t,T}^X(f))}{\eta_0^X Q_t^X \Pi_0 Q_{T-t}^X \1_{\cX}} \partial_t\log(\eta_0^X Q_t^X \Pi_0 Q_{T-t}^X \1_{\cX})\,\d t                       \\
             & = \frac{\eta_0^X Q_T^X \Pi_1 r_{T,T}^X(f)}{\eta_0^X Q_T^X \Pi_0 \1_{\cX}} - \frac{\eta_0^X \Pi_1 Q_T^X r_{0,T}^X(f)}{\eta_0^X\Pi_0 Q_{T}^X \1_{\cX}} +\int_0^T \frac{\Phi(t,r_{t,T}^X(f))}{\eta_0^X Q_t^X \Pi_0 Q_{T-t}^X \1_{\cX}} \partial_t\log(\eta_0^X Q_t^X \Pi_0 Q_{T-t}^X \1_{\cX})\,\d t \\
             & = - \frac{\eta_0^X \Pi_1Q_T^X r_{0,T}^X(f)}{\eta_0^X\Pi_0 Q_{T}^X \1_{\cX}}  +\int_0^T \frac{\Phi(t,r_{t,T}^X(f))}{\eta_0^X Q_t^X \Pi_0 Q_{T-t}^X \1_{\cX}} \partial_t\log(\eta_0^X Q_t^X \Pi_0 Q_{T-t}^X \1_{\cX})\,\d t,
        \end{aligned}
    \end{equation}

    using \(r^X_{t,T}(f)\in \Pi_0 \ell^\infty(\cX)\) for any \(t\in [0,T]\) to cancel the first boundary term.

    Using a bound identical to~\eqref{eq:residual_duhamel_estimate} on the numerator of the remaining boundary term, we get
    \begin{equation}
        \label{eq:boundary_term_gap_estimate}
        \begin{aligned}
             & |\eta_0^X \Pi_1 Q_T^X r_{0,T}^X(f)|  \leq \|\eta_0^X\|_*\|\Pi_1 Q_T^X r_{0,T}^X(f)\|                                               \\
             & \leq \left(C_7^X\frac{\e^{-(\lambda^X+\gamma^X)T}}{\zeta} + C_7^{'X}\frac{\e^{-(\lambda^X+\gamma^X)T}}{\zeta}\right)\|f\|,
        \end{aligned}
    \end{equation}
    where, by convention, each constant \(C_j^X\) is defined by substituting \(Z\leftarrow X\) in the definition of \(C_j^Z\). The boundary term is then controlled using the lower bound~\eqref{eq:survival_bound} by
    \begin{equation}
        \label{eq:boundary_term_final_bound}
        \begin{aligned}
            \left|\frac{\eta_0^X \Pi_1 Q_T^X r_{0,T}^X(f)}{\eta_0^X\Pi_0 Q_{T}^X \1_{\cX}}\right|\leq \e^{-\gamma^X T}\frac{C_7^X+C_7^{'X}}{C_1^X \zeta}\|f\|.
        \end{aligned}
    \end{equation}

    We now control the integrand in~\eqref{eq:ibp_error_measure} pointwise. Using the canonical modeling identity \( \nu^X \Pi_0 = \nu^X\), which implies \(\Pi_1 = P^X \Pi_1\), as well as an identical estimate to~\eqref{eq:duhamel_estimate}, we have
    \begin{equation}
        \label{eq:ibp_integral_term_duhamel_estimate}
        \begin{aligned}
            |\Phi(t, r_{t,T}^X(f))| & = |\eta_0^X Q_t^X P^X\Pi_1 Q_{T-t}^X r_{t,T}^X(f)|                                                                                                                                  \\
                                    & \leq  C_\gamma^X\e^{-(\lambda^X+\gamma^X)t}\|\Pi_1 Q_{T-t}^X r_{t,T}^X(f)\|                                                                                                         \\
                                    & \leq  C_\gamma^X\e^{-(\lambda^X+\gamma^X)t} \left(C_7^X\frac{\e^{-(\lambda^X+\gamma^X)(T-t)}}{\zeta} + C_7^{'X}\frac{\e^{-(\lambda^X+\gamma^X)(T-t)-\gamma^X t}}{\zeta}\right)\|f\| \\
                                    & := \e^{-\lambda^X T}\left(C_9^X\frac{\e^{-\gamma^X t}\e^{-\gamma^X(T-t)}}{\zeta} + C_9^{'X}\frac{\e^{-2\gamma^X t}\e^{-\gamma^X(T-t)}}{\zeta}\right)\|f\|.
        \end{aligned}
    \end{equation}

    Dividing by~\eqref{eq:survival_bound}, combining with the pointwise bound on the logarithmic derivative~\eqref{eq:survival_probability_derivative_bound} and integrating for \(t\in(0,T)\), we get
    \begin{equation}
        \label{eq:ibp_integral_term_final_bound}
        \begin{aligned}
             & \left|\int_0^T \frac{\Phi(t,r_{t,T}^X(f))}{\eta_0^X Q_t^X \Pi_0 Q_{T-t}^X \1_{\cX}} \partial_t\log(\eta_0^X Q_t^X \Pi_0 Q_{T-t}^X \1_{\cX})\,\d t\right|                                                                   \\
             & \leq \int_0^T \frac{C_5^X}{C_1^X} \e^{-\gamma^X t} \left(C_9^X\frac{\e^{-\gamma^X t}\e^{-\gamma^X(T-t)}}{\zeta} + C_9^{'X}\frac{\e^{-2\gamma^X t}\e^{-\gamma^X(T-t)}}{\zeta}\right)\|f\| \,\d t                            \\
             & \leq \left(\frac{C_5^X C_9^{X}}{C_1^X\zeta \gamma^X}\e^{- \gamma^X T} + \frac{C_5^X C_9^{'X}}{C_1^X\zeta 2\gamma^X}\e^{-\gamma^X T}\right)\|f\| = \frac{C_5^X(C_9^X+C_9^{'X}/2)}{C_1^X\zeta\gamma^X}\e^{-\gamma^X T}\|f\|,
        \end{aligned}
    \end{equation}
    using~\eqref{eq:convolution_inequality} to estimate \(\int_0^T \e^{-K\gamma^X t}\e^{-\gamma^X(T-t)}\,\d t \leq \frac1{(K-1)\gamma^X}\e^{-\gamma^X T}\) in the last line.

    It remains to treat the integral term
    \begin{equation}
        \int_{0}^T \frac{\Phi(t,\1_{\cX})\partial_{t}\psi_{t,T}^X(\xi^\star f)}{\eta_0^X Q_t^X \Pi_0 Q_{T-t}^X \1_{\cX}}\, \d t.
    \end{equation}

    The contribution of~\(|\partial_{t}\psi_{t,T}^X(\xi^\star f)|\) is controlled using the coarse estimate~\eqref{eq:time_derivatives_first_estimates}.
    For~\(\Phi(t,\1_{\cX})\), we use the Duhamel--Dyson formula~\eqref{eq:duhamel_formula} with no spectral decomposition of~\(\1_{\cX}\), noting~\(\Pi_1\1_{\cX}=0\) and inserting \(\Pi_1=P^X\Pi_1\), and estimate
    \begin{equation}
        \begin{aligned}
            |\Phi(t,\1_{\cX})| & = |\eta_0^X Q_t^X P^X\Pi_1 Q_{T-t}^X\1_{\cX}|                                                                                                     \\
                               & \leq C_\gamma^X \e^{-(\lambda^X+\gamma^X)t} \|\Pi_1 Q_{T-t}^X \1_{\cX}\|                                                                          \\
                               & \leq C_\gamma^X \e^{-(\lambda^X +\gamma^X)t}C_\zeta C_{\lambda}^X\int_0^{T-t}\e^{-\zeta (T-t-u)}\|\cL^X_{10}\|\e^{-\lambda^X u}\|\1_{\cX}\|\,\d u \\
                               & = C_\gamma^X \e^{-(\lambda^X +\gamma^X)t}C_\zeta C_{\lambda}^X\|\cL^X_{10}\| \frac{\e^{-\lambda^X(T-t)}-\e^{-\zeta(T-t)}}{\zeta-\lambda^X}        \\
                               & \leq \frac{2C_\gamma^X C_\zeta C_{\lambda}^X\|\cL^X_{10}\|}{\zeta}\e^{-\lambda^X T}\e^{-\gamma^X t}                                               \\
                               & := \frac{C_{10}^X}{\zeta}\e^{-\lambda^X T}\e^{-\gamma^X t}.
        \end{aligned}
    \end{equation}

    Therefore, substituting the pointwise estimate~\eqref{eq:time_derivatives_first_estimates} for the time derivative~\(\partial_t\psi_{t,T}^X(\xi^\star f)\), the final integral is bounded by
    \begin{equation}
        \label{eq:ibp_bootstrap_term_final_bound}
        \begin{aligned}
             & \left|\int_{0}^T \frac{\Phi(t,\1_{\cX})\partial_{t}\psi_{t,T}^X(\xi^\star f)}{\eta_0^X Q_t^X \Pi_0 Q_{T-t}^X \1_{\cX}}\, \d t\right|                                                 \\
             & \leq \int_0^T \frac{C_{10}^X}{C_1^X\zeta}\e^{-\gamma^X t}\left(\frac{C_2^X}{C_1^X} \e^{-\gamma^X t} + \frac{C_3^X C_4^X}{C_1^X}\e^{-2\gamma^X t}\right)\e^{-\gamma^X(T-t)}\|f\| \d t \\
             & \leq \e^{-\gamma^X T}\frac{C_{10}^X}{C_1^X\zeta}\left(\frac{C_2^X}{\gamma^X C_1^X}+\frac{C_3^X C_4^X}{2\gamma^X C_1^X}\right)\|f\|.
        \end{aligned}
    \end{equation}

    Summing the estimates~\eqref{eq:Z_contribution_final_bound},~\eqref{eq:boundary_term_final_bound},~\eqref{eq:ibp_integral_term_final_bound} and~\eqref{eq:ibp_bootstrap_term_final_bound}, we arrive at
    \begin{equation}
        \label{eq:final_bound}
        |\xi_\star \eta_T^X(f) - \eta_T^Z(f)| \leq \zeta^{-1}\left(\widetilde{C}\e^{-\gamma^X T}+\widetilde{C}'T\e^{-(\gamma^Z\land\gamma^X)T}\right)\|f\|,
    \end{equation}
    with explicit constants
    \begin{equation}
        \label{eq:prefactors}
        \begin{aligned}
            \widetilde{C}  & = \frac{C_7^X+C_7^{'X}}{C_1^X}
            + \frac{C_5^X\left(C_9^{X}+C_9^{'X}/2\right)}{\gamma^X C_1^X}
            + \frac{C_{10}^X}{\gamma^X C_1^X}\left(\frac{C_2^X}{C_1^X}+\frac{C_3^X C_4^X}{2C_1^X}\right),
            \\
            \widetilde{C}' & = \frac{C_8^Z + C_8^{'Z}}{C_1^Z}.
        \end{aligned}
    \end{equation}

    To make the constants more manageable, we can bound them further.
    Since \(\|\Pi^Z\| = \|\Pi^X\| = 1\), we also have \(\|A^Z\|=\|\Pi^Z A^X\|\leq\|A^X\|\) by~\eqref{eq:intertwining_defect_control}. Similarly, a direct computation gives
    \(\cL_{10}^X = A^X\Pi^X\) and~\(\cL_{01}^X = -\Pi^X A^X\), therefore \(\|\cL_{10}^X\|,\|\cL_{01}^X\|\leq \|A^X\|\). Finally we note the estimates following from Proposition~\ref{prop:spectral_properties}:
    \begin{equation}
        \label{eq:PX_estimate}
        \|P^X\| = \|Q_0^X P^X\|\leq \underset{t\geq 0}{\sup}\,\|Q_t^X P^X\|\e^{(\lambda^X+\gamma^X)t}\leq C_\gamma^X,
    \end{equation}
    \begin{equation}
        \label{eq:nhX_estimate}
        \|h^X\| = \|h^X\|\|\nu^X\|_* =\|h^X\nu^X\| = \|\underset{t\to\infty}{\lim}\e^{\lambda^X t}Q_t^X\| =\underset{t\to\infty}{\lim}\|\e^{\lambda^X t}Q_t^X\|\leq C_\lambda^X,
    \end{equation}
    where we view \(h^X\nu^X\) as a rank-one operator.

    Substituting the definitions of the constants in~\eqref{eq:prefactors}, estimating using~\eqref{eq:PX_estimate} and~\eqref{eq:nhX_estimate} and simplifying leads, with constants~\(C=\widetilde C/\|A^X\|\) and~\(C'=\widetilde{C}'/\|A^X\|\), to the slightly more manageable form
    \begin{equation}
        \label{eq:prefactors_expanded}
        \begin{aligned}
            C' & \leq \frac{4 C_\lambda^X C_\gamma^X C_\gamma^Z C_\zeta \|A^X\|}{C^Z(\eta_0)}\left(C_\gamma^X+\frac{(C_\lambda^X)^3 C_\gamma^Z}{C^Z(\eta_0)}\right),\\                                                               
            C  & \leq \frac{4C_\lambda^X(C_\gamma^X)^2 C_\zeta}{C^X(\eta_0)}\left(1+\frac{(C_\lambda^X)^3}{C^X(\eta_0)}\right)                                 \\
               & \qquad + \frac{4(C_\lambda^X)^3(C_\gamma^X)^4C_\zeta\|A^X\|}{\gamma^X C^X(\eta_0)^2}\left(1+\frac{(C_\lambda^X)^3}{2C^X(\eta_0)}\right)          \\
               & \qquad + \frac{4(C_\lambda^X)^3(C_\gamma^X)^3C_\zeta\|A^X\|}{\gamma^X C^X(\eta_0)^2}\left(1+\frac{(C_\lambda^X)^3C_\gamma^X}{2C^X(\eta_0)}\right)
        \end{aligned}
    \end{equation}
    where the constants appearing in the denominators are
    \begin{equation}
        \label{eq:denominator_constants}
        C^Z(\eta_0) = \eta_0^Z(h^Z)\underset{\cZ}{\min} \frac{\Pi^Z h^X}{h^Z},\qquad C^X(\eta_0) = \eta_0^X(h^X)\underset{\cX}{\min} \frac{\Pi^X h^X}{h^X}.
    \end{equation}
    Note that the \(\ell^\infty\) quasimode estimates~\eqref{eq:X_quasimode_estimate} and~\eqref{eq:Z_quasimode_estimate} do not provide further control of these constants. Doing so would require instead \(h^Y\)-norm estimates.

    The total variation bound~\eqref{eq:tv_bound} then follows from~\eqref{eq:final_bound} by the duality formula~\eqref{eq:tv_dual_formula} taking the supremum over \(f\) such that \(\|f\|\leq 1\) in~\eqref{eq:final_bound}.

    \subsubsection{Proof of the quasimode estimates~\eqref{eq:X_quasimode_estimate} and~\eqref{eq:Z_quasimode_estimate}}
    We now show the second item in Theorem~\ref{thm:cg_existential_result}. Using the formula~\eqref{eq:duhamel_formula} on \(v=h^X\), we get, for any \(t\geq 0\),
    \begin{equation}
        \Pi_1 h^X = \e^{\lambda^X t}\left(\e^{t\cL^X_{11}}\Pi_1 h^X +  \int_0^t \e^{(t-s)\cL^X_{11}}\cL^X_{10} \e^{-\lambda^X s}h^X \,\d s \right).
    \end{equation}
    Estimating, this gives
    \begin{equation}
        \begin{aligned}
            \|\Pi_1 h^X\| & \leq C_{\zeta}\e^{-(\zeta-\lambda^X)t}\|\Pi_1 h^X\| + C_\zeta\frac{1-\e^{-(\zeta-\lambda^X)t}}{\zeta-\lambda^X}\|\cL_{10}^X\|\|h^X\|,
        \end{aligned}
    \end{equation}
    and the bound~\eqref{eq:X_quasimode_estimate} follows by letting \(t\to \infty\), using \(\|\cL_{10}^X\|\leq \|A^X\|\) and the estimate~\eqref{eq:nhX_estimate}.

    To obtain the second estimate, we write
    \begin{equation}
        \begin{aligned}
            (\cL^Z + \lambda^Z)(\Pi^Z h^X - h^Z) & = \cL^Z\Pi^Z h^X + \lambda^X\Pi^Z h^X      \\
                                                 & = \left[\cL^Z\Pi^Z - \Pi^Z \cL^X\right]h^X \\
                                                 & = A^Z h^X.
        \end{aligned}
    \end{equation}
    Noting that \(\nu^Z(\Pi^Z h^X) = \nu^X(h^X) = \nu^Z(h^Z) = 1\) and using \(\nu^Z A^Z=0\), we may insert \(P^Z\) on both sides, giving
    \begin{equation}
        (\cL^Z + \lambda^Z)P^Z(\Pi^Z h^X - h^Z) = P^Z A^Z h^X,
    \end{equation}
    which is a Poisson equation on \(P^Z \ell^\infty(\cZ)\). Since~\(\lambda^Z\) is a simple eigenvalue by Perron--Frobenius, this equation is well-posed on~\(P^Z \ell^\infty(\cZ)\).
    Applying
    \begin{equation}
        \label{eq:partial_inverse}
        \left.\left(\cL^Z + \lambda^Z\right)\right|_{P^Z \ell^\infty(\cZ)}^{-1} = -\left.\int_0^{\infty}\e^{\lambda^Z t}Q_{t}^Z\,\d t\right|_{P^Z \ell^\infty(\cZ)},
    \end{equation}
    we obtain
    \begin{equation}
        \Pi^Z h^X - h^Z =P^Z(\Pi^Z h^X - h^Z) = -\int_0^{\infty}\e^{\lambda^Z t}Q_{t}^Z P^Z A^Z h^X\,\d t,
    \end{equation}
    which gives the estimate
    \begin{equation}
        \|\Pi^Z h^X - h^Z\| \leq \frac{C_{\gamma}^Z}{\gamma^Z}\|A^Z h^X\|.
    \end{equation}
    At this point, we use~\eqref{eq:AZ_fast} to insert \(A^Z h^X = A^Z \Pi_1 h^X\) to estimate
    \begin{equation}
        \|\Pi^Z h^X - h^Z\| \leq \frac{C_{\gamma}^Z}{\gamma^Z}\|A^Z\|\|\Pi_1 h^X\| \leq  \frac{C_{\gamma}^Z}{\gamma^Z}\|A^X\|\|\Pi_1 h^X\|,
    \end{equation}
    which gives~\eqref{eq:Z_quasimode_estimate} upon substituting the estimate~\eqref{eq:X_quasimode_estimate} on the right-hand side of this inequality.

    \subsubsection{Proof of the two-sided estimate~\eqref{eq:two_sided_error_estimate}}
    We finally show the third item in Theorem~\ref{thm:cg_existential_result}. By the duality formula~\eqref{eq:tv_dual_formula}, we have
    \begin{equation}
        \label{eq:tracking_bound_A}
        \d_{\mathrm{TV}}(\xi_\star \eta_t^X,\xi_\star\nu^X) \leq \d_{\mathrm{TV}}(\eta_t^X,\nu^X).
    \end{equation}

    Now, for any \(f\in \ell^\infty(\cX)\), we may write
    \begin{equation}
        \label{eq:tv_decomposition}
        \begin{aligned}
            \eta_t^X(f) - \nu^X(f) & = \eta_t^X(\Pi_0 f) - \nu^X(\Pi_0 f) + \eta_t^X(\Pi_1 f) - \nu^X(\Pi_1 f) \\
                                   & =\xi_\star\eta_t^X(\Pi^Z f) - \xi_\star\nu^X(\Pi^Z f)+\eta_t^X(\Pi_1 f),
        \end{aligned}
    \end{equation}
    where we used~\eqref{eq:qsd_projection_invariance_X} to deduce \(\nu^X\Pi_1 = 0\).

    Since \(\|\Pi^Z f\|\leq \|f\|\), we have
    \begin{equation}
        \label{eq:tracking_bound_B}
        \begin{aligned}
            \underset{\|f\|\leq 1}{\sup}\, \left[\xi_\star\eta_t^X(\Pi^Z f) - \xi_\star\nu^X(\Pi^Z f)\right] & \leq \underset{\|g\|\leq 1}{\sup}\, \xi_\star\eta_t^X(g) - \xi_\star\nu^X(g) \\
                                                                                                & =\d_{\mathrm{TV}}(\xi_\star\eta_t^X,\xi_\star\nu^X),
        \end{aligned}
    \end{equation}
    which leaves a term
    \begin{equation}
        \eta_t^X(\Pi_1 f) = \frac{\eta_0^X(Q_t^X\Pi_1 f)}{\eta_0^X(Q_t^X\1_{\cX})}= \frac{\eta_0^X(Q_t^XP^X\Pi_1 f)}{\eta_0^X(Q_t^X\1_{\cX})}
    \end{equation}
    to estimate. Using the commutation \(\cL^X P^X = P^X\cL^X\) and the Feynman--Kac formula~\eqref{eq:feynman_kac}, we split
    \begin{equation}
        \partial_t\eta_0^X(Q_t^X P^X\Pi_1) = \eta_0^X(Q_t^X P^X \cL^X\Pi_1) = \eta_0^X(Q_t^X P^X\Pi_1\cL_{11}^X)+\eta_0^X(Q_t^X P^X \cL_{01}^X),
    \end{equation}
    which by Duhamel--Dyson's formula gives, after application to \(f\),
    \begin{equation}
        \eta_0^X(Q_t^X P^X \Pi_1 f ) = \eta_0^X(P^X \e^{t\cL_{11}^X}\Pi_1 f ) + \int_0^t  \eta_0^X(Q_s^X P^X \cL_{01}^X \e^{(t-s)\cL_{11}^X}\Pi_1 f)\,\d s,
    \end{equation}
    using \(\Pi_1\e^{t\cL_{11}^X} = \e^{t\cL_{11}^X}\Pi_1\). Estimating, we find
    \begin{equation}
        \begin{aligned}
            |\eta_0^X(Q_t^X P^X \Pi_1 f )| & \leq \|\eta_0^X\|_*\|P^X\|C_\zeta\e^{-\zeta t}\|f\|+\int_0^t\|\eta_0^X\|_*C_{\gamma}^X\e^{-(\lambda^X+\gamma^X)s}\|\cL_{01}^X\|C_\zeta\e^{-\zeta (t-s)}\|f\|\,\d s \\
                                           & \leq \|P^X\|C_\zeta \e^{-\zeta t}\|f\| + C_\gamma^X \|\cL_{01}^X\|C_\zeta \frac{\e^{-(\lambda^X+\gamma^X)t} - \e^{-\zeta t}}{\zeta -\lambda^X-\gamma^X}\|f\|       \\
                                           & \leq C_\gamma^XC_\zeta\left(\e^{-\zeta t} + 2\|A^X\|\frac{\e^{-(\lambda^X+\gamma^X)t}}{\zeta}\right)\|f\|,
        \end{aligned}
    \end{equation}
    where we used~\eqref{eq:PX_estimate} and~$\|\cL_{01}^X\|\leq\|A^X\|$ in the last line.

    On the other hand, we have the lower bound
    \begin{equation}
        \eta_0^X(Q_t^X \1_{\cX}) \geq \frac1{\|h^X\|}\eta_0^X(Q_t^X h^X) = \frac{\e^{-\lambda^X t}}{\|h^X\|}\eta_0^X(h^X),\,
    \end{equation}
    which gives
    \begin{equation}
        \label{eq:tracking_bound_C}
        \begin{aligned}
            \underset{\|f\|\leq 1}{\sup}\,\eta_t^X(\Pi_1 f) & \leq \frac{\|h^X\|}{\eta_0^X(h^X)}C_\gamma^XC_\zeta\left(\e^{-(\zeta-\lambda^X) t} + 2\|A^X\|\frac{\e^{-\gamma^X t}}{\zeta}\right)       \\
                                                            & \leq \frac{C_\lambda^X C_\gamma^XC_\zeta}{\eta_0^X(h^X)}\left(\e^{-(\zeta-\lambda^X) t} + 2\|A^X\|\frac{\e^{-\gamma^X t}}{\zeta}\right).
        \end{aligned}
    \end{equation}

    Taking the supremum with respect to~\(\|f\|\leq 1\) in~\eqref{eq:tv_decomposition} and using the estimates~\eqref{eq:tracking_bound_B} and~\eqref{eq:tracking_bound_C}, we get estimate~\eqref{eq:two_sided_error_estimate} when combined with~\eqref{eq:tracking_bound_A}.
\end{proof}

\subsection{Uniform coarse-graining estimates in the adiabatic limit}
\label{sec:perturbation_theory}
We describe the behavior of the estimate~\eqref{eq:tv_bound} provided by Theorem~\ref{thm:cg_existential_result} on a more specific class of models, in which the timescale separation \(\zeta\) is tuned by a small adiabatic parameter \(\varepsilon\).

Let us fix a partition of the state space
\begin{equation}
    \label{eq:partition}
    \cX = \underset{1\leq m\leq M}{\bigsqcup} \cX_m,
\end{equation}
where each \(\cX_m\) is a fiber, that is a level set of the collective variable
\begin{equation}
    \label{eq:cv_discrete}
    \xi(x) = \sum_{m=1}^M m\1_{\cX_m}(x)
\end{equation}
from \(\cX\) to \(\cZ=\{1,\dots,M\}\).

We consider the case where the killed semigroup is generated by the sum of three operators
\begin{equation}
    \label{eq:jp_generator}
    \cL^X_\varepsilon = \cL_{\rm mac} -K + \varepsilon^{-1}\cL_{\mathrm{mic}},
\end{equation}
where \(\cL_{\rm mac}\) is the generator of an irreducible mass-conserving jump process on \(\cX\), \(K\) is a nonnegative, nonzero diagonal operator encoding the jump rates to the boundary (i.e. killing rates),
and \(\cL_{\rm mic}\) is the generator of a fiber-reducible jump process, that is, a direct sum
\begin{equation}
    \label{eq:block_diagonal}
    \cL_{\rm mic} = \bigoplus_{m=1}^M A_m,
\end{equation}
where \(A_m\) generates an irreducible mass-conserving jump process on~\(\cX_m\).

We call the limit \(\varepsilon\to 0\) the~\emph{adiabatic regime}, and indicate the~\(\varepsilon\)-dependence of various quantities with subscripts.
The initial distributions~\(\eta_0^X\) and~\(\eta_0^Z\) are fixed independently from~\(\varepsilon\).
We denote by~\(\mu_m\in\cP(\cX_m)\) the unique \(A_m\)-invariant probability measure. Define the operators
\begin{equation}
    \label{eq:spectral_projector_two_scale}
    P_0^{\rm s} = \sum_{m=1}^M \1_{\cX_m}\mu_m,\qquad \cL_{\rm avg} = P_0^{\rm s}(\cL_{\rm mac}-K)P_0^{\rm s}.
\end{equation}
The operator \(P_0^{\rm s}\) is a rank-\(M\) projector onto \(\Pi_\varepsilon^X \ell^\infty(\cX)\). In general \(\Pi_\varepsilon^X\neq P_0^{\rm s}\), since $\nu^X_\varepsilon$-conditional averages over~$\cX_m$ may not agree with averages under~$\mu_m$.

Denote by~\((\lambda_0^{\rm s},h_0^{\rm s},\nu_0^{\rm s}) \in (0,\infty)\times \ell^\infty(\cX)\times \cP(\cX)\) the principal eigentriplet of~\(\cL_{\rm avg}\) as an operator on~$P_0^{\rm s}\ell^\infty(\cX)$, normalized so that~\(\nu_0^{\rm s}(h_0^{\rm s})=1\).
Its uniqueness and the strict positivity of~\(h_0^{\rm s}\) and~\(\nu_0^{\rm s}\) will be checked in the proof of Theorem~\ref{thm:perturbation_theory}.

Using spectral perturbation theory~\cite[Chapter 2]{K95}, we show the following result.
\begin{theorem}
    \label{thm:perturbation_theory}
    Assume that the compatibility condition~\(\xi_\star \eta_0^X = \eta_0^Z\) holds. Then, there exist~\(C,\varepsilon_0>0\) such that, for any~\(\varepsilon\in(0,\varepsilon_0)\),
    \begin{equation}
        \label{eq:adiabatic_limit}
        \left\|\Pi_\varepsilon^X - P_0^{\rm s}\right\|,\,\left|\lambda_{\varepsilon}^X- \lambda_0^{\rm s}\right|,\,\left|\underset{\cX}{\min}\,\frac{\Pi_\varepsilon^X h_\varepsilon^X}{h_\varepsilon^X}-1\right|,\,\left|\underset{\cZ}{\min}\frac{\Pi_\varepsilon^Z h_\varepsilon^X}{h_\varepsilon^Z}-1\right|\leq C\varepsilon,
    \end{equation}
    and the Assumptions of Theorem~\ref{thm:cg_existential_result} are verified, with constants
    \begin{equation}
        \label{eq:adiabatic_constants_control}
        \gamma_{\varepsilon}^X,\gamma_{\varepsilon}^Z\geq 1/C,\qquad C_{\lambda,\varepsilon}^X,C_{\lambda,\varepsilon}^Z,C_{\gamma,\varepsilon}^X,C_{\gamma,\varepsilon}^Z,C_{\zeta,\varepsilon},\|A_\varepsilon^X\|\leq C,\qquad \zeta_\varepsilon\geq 1/(C\varepsilon).
    \end{equation}

    Moreover,
    \begin{equation}
        \label{eq:initial_condition_control}
        \left|\eta_0^X(h_\varepsilon^X)-\eta_0^X(h_0^{\rm s})\right|,\,\left|\eta_0^Z(h_\varepsilon^Z)-\eta_0^X(h_0^{\rm s})\right|\leq C\varepsilon,\qquad \eta_0^X(h_0^{\rm s})\geq 1/C.
    \end{equation}
\end{theorem}
A self-contained proof is given in Appendix~\ref{sec:proof_perturbation_theory}.

Theorem~\ref{thm:perturbation_theory} then implies uniform-in-\(\varepsilon\) versions of Theorem~\ref{thm:cg_existential_result} and Corollary~\ref{cor:diagnostic_bound}.
\begin{corollary}
    \label{cor:uniform_coarse_graining}
    Assume that the compatibility condition~\(\xi_\star \eta_0^X =\eta_0^Z\) holds. There exist~\(C,\varepsilon_0>0\) such that for all~\(\varepsilon\in(0,\varepsilon_0)\), it holds
    \begin{equation}
        \label{eq:uniform_flow_approximation}
        \underset{t\geq 0}{\sup}\, \d_{\rm TV}\left(\eta_{t,\varepsilon}^Z,\xi_\star\eta_{t,\varepsilon}^X\right)\leq C\varepsilon,
    \end{equation}
    and for~\(t_*(\varepsilon) = 2C\varepsilon|\log\varepsilon|\),
    \begin{equation}
        \label{eq:uniform_diagnostic_bound}
        \underset{t\geq t_*(\varepsilon)}{\sup}\, \left|\d_{\rm TV}\left(\eta_{t,\varepsilon}^Z,\nu_\varepsilon^Z\right)- \d_{\rm TV}\left(\eta_{t,\varepsilon}^X,\nu_\varepsilon^X\right)\right| \leq C\varepsilon.
    \end{equation}
\end{corollary}
\begin{proof}
    We begin by proving~\eqref{eq:uniform_flow_approximation}.  From~\eqref{eq:tv_bound} and Theorem~\ref{thm:perturbation_theory}, there exist~$C,\varepsilon_0>0$ such that for all~$\varepsilon\in(0,\varepsilon_0)$, we have, for any~$T\geq 0$, the bound
    \begin{equation}
        \d_{\rm TV}(\xi_\star\eta_{T,\varepsilon}^X,\eta_{T,\varepsilon}^Z)\leq C\varepsilon(1+T)\e^{-T/C}\leq C^2\e^{-(C-1)/C}\varepsilon,
    \end{equation}
    maximizing the middle expression with respect to~$T$ to obtain the final bound. This proves~\eqref{eq:uniform_flow_approximation} upon increasing~$C$. The uniform control on the constants appearing in~\eqref{eq:tv_bound} can be verified using their explicit form~(see Equations~\eqref{eq:prefactors_expanded} and~\eqref{eq:denominator_constants}).

    The estimate~\eqref{eq:uniform_diagnostic_bound} requires the burn-in time~\(t_*(\varepsilon)>0\) to control a boundary layer term~\(\e^{-(\zeta_\varepsilon-\lambda_\varepsilon^X)t}\) appearing in~\eqref{eq:diagnostic_bound}.
    By~\eqref{eq:diagnostic_bound} and Theorem~\ref{thm:perturbation_theory}, there exist~$C,\varepsilon_0$ such that, for all~$\varepsilon\in(0,\varepsilon_0)$,~\(\zeta_\varepsilon \geq 1/(C\varepsilon)\),~\(\lambda_\varepsilon^X\leq C\), and that for all~$T\geq 0$,
    \begin{equation}
        \left|\d_{\rm TV}(\eta_{T,\varepsilon}^Z,\nu_\varepsilon^Z)-\d_{\rm TV}(\eta_{T,\varepsilon}^X,\nu_\varepsilon^X)\right|\leq C\left(\e^{-(1/(C\varepsilon)-C)T} + \varepsilon\e^{-T/C}\right) + \varepsilon C(1+T)\e^{-T/C}.
    \end{equation}
    For~$\varepsilon< 1/(2C^2)$,~$1/(C\varepsilon)-C>1/(2C\varepsilon)$, which implies that for~$T\geq 2C\varepsilon|\log\varepsilon|$,
    \begin{equation}
        \e^{-(1/(C\varepsilon)-C)T}\leq \e^{-T/(2C\varepsilon)} \leq \e^{-|\log\varepsilon|} = \varepsilon,
    \end{equation}
    provided~$\varepsilon\leq 1$. This leaves us, for~$T\geq t_*(\varepsilon)$, with
        \begin{equation}
            \begin{aligned}
                &\left|\d_{\rm TV}(\eta_{T,\varepsilon}^Z,\nu_\varepsilon^Z)-\d_{\rm TV}(\eta_{T,\varepsilon}^X,\nu_\varepsilon^X)\right|\leq \varepsilon C\left(1 + (2+T)\e^{-T/C}\right)\\
                &\qquad \leq \varepsilon C(1+C\e^{-(C-2)/C}),
            \end{aligned}
    \end{equation}
    which concludes the proof upon increasing~$C$ and decreasing~$\varepsilon_0$.
\end{proof}

We illustrate the adiabatic limit numerically in Section~\ref{sec:num_perturbation_theory}, on a simple killed jump process with four states.

\subsection{Particle approximations of non-linear semigroups}
\label{sec:fv}
Online definitions of the decorrelation time in the parallel replica method require access to numerical approximations of the non-linear semigroup~$\eta^X_t$.
For many systems of practical interest, approximations based on the formula~\eqref{eq:non_linear_flow} are intractable, as they require direct propagation of the linear semigroup~\(Q_t^X\). This may not be possible when~\(\cX\) is large and~\(\cL^X\) does not have a convenient sparse representation.

For such cases, which appear for instance in molecular dynamics, one can still approximate~$\eta^X_t$ using interacting particle systems, which is the choice made in the generalized parallel replica method of~\cite{BLS15}.
The simplest approximation is given by the Fleming--Viot process, a type of branching process defined recursively as follows.

\begin{definition}[Fleming--Viot process]
    \label{def:fleming_viot}
    Fix an integer~\(N\geq 1\). The \(N\)-particle Fleming--Viot process~\({\bf X} = (X^{(n)})_{1\leq n\leq N}\) is an~\(\cX^N\)-valued stochastic process, whose paths are constructed as follows.
    \begin{itemize}
        \item[{\bf A:}]{ {\bf Initialization.} Fix the initial time~\(T_0=0\), branching count~\(k=0\) and initial point~\({\bf X}_0 = \left(X^{(j),0}_0\right)_{1\leq j\leq N}\).}
        \item[{\bf B:}]{{\bf Evolution.} For any~\(k\geq 0\), sample on the time interval~\([T_k,T_{k+1})\) (with \(T_{k+1}\) defined in~\eqref{eq:branching_rule}), independent Markovian trajectories~\((X^{(j),k})_{1\leq j\leq N}\) on~\(\overline{\cX}\) with infinitesimal generator~\(\cL^X\), starting from~\(X^{(j),k}_{T_k}\).
              Here we define
              \begin{equation}
                  \label{eq:branching_time}
                  T_{k+1}^j = \inf\,\left\{t>T_k:\,X_t^{(j),k}=\partial^\cX\right\},\qquad J_k = \underset{1\leq j\leq N}{\rm argmin}\,T_{k+1}^j.
              \end{equation}
              Note that the~\(J_k\) are almost-surely uniquely defined.
              }
        \item[{\bf C:}]{{\bf Branching and iteration.} Define the branching time
                  \begin{equation}
                      \label{eq:branching_rule}
                      T_{k+1}:= T^{J_k}_{k+1},
                  \end{equation}
                  and set~\(X_t^{(j)} = X^{(j),k}_t\) for~\(t\in (T_k,T_{k+1})\).

                  Then, set
                  \begin{equation}
                      \forall\,j\neq J_k,\qquad X_{T_{k+1}}^{(j),k+1}= X_{T_{k+1}^-}^{(j),k} \in \cX\,\text{ almost surely},
                  \end{equation}
                  and sample independently
                  \begin{equation}
                      X_{T_{k+1}}^{(J_k),k+1} \widesim{} \frac1{N-1}\sum_{j\neq J_k} \delta_{X^{(j),k}_{T_{k+1}^-}},
                  \end{equation}
                  where~\(Y_{t^-} = \underset{s\uparrow t}{\lim}\, Y_s\) is well-defined whenever \(Y\) is c\`adl\`ag.

                  Set~\(X^{(j)}_{T_{k+1}}=X_{T_{k+1}}^{(j),k+1}\) for all~\(1\leq j\leq N\), and proceed from step {\bf B.} with~\({k\leftarrow k+1}\).
              }
    \end{itemize}
\end{definition}
In plainer terms, trajectories of the original jump process~\(X\) are simulated independently, and each time a replica reaches the boundary~\(\partial^{\cX}\), its state is branched from one of the surviving replicas, sampled independently and uniformly at random.
We stress that a killed replica only inherits the state of a surviving replica during a branching event, but retains its past trajectory.

The same construction may be attempted for more general choices of one-particle c\`adl\`ag dynamics (i.e. \(\cX\) need not be finite nor~\(\cL^X\) generate a jump process).

By viewing the Fleming--Viot process~\(\bf X\) as a~\(\cP(\cX)\)-valued process via the dynamics of the empirical measure
\begin{equation}
    \label{eq:fv_empirical_measure}
    \widehat{\eta}_{t,N}^X = \frac1N \sum_{n=1}^N \delta_{X_t^{(n)}}\in\cP(\cX),
\end{equation}
we obtain a particle approximation of the non-linear flow~\(\eta_t^X\). It is indeed expected that~\(\widehat{\eta}_{t,N}^X\) converge to the non-linear flow~\(\eta_t^X\) in the limit~\(N\to\infty\) for many choices of~\(\cL^X\) (and some appropriate choice of topology).

When such a property is verified, one says that~\emph{propagation of chaos} holds.
In~\cite{V14}, such approximation results have been obtained for general choices of the one-particle Markovian dynamics in Definition~\ref{def:fleming_viot}, but these only prove propagation of chaos at fixed times~\(t>0\).
For uniform-in-time results, we refer to~\cite{CT16} in the case of jump processes. The case of killed diffusion processes is still open in generality, but partial results have been obtained, see for instance~\cite{JM22,JR24,JM25,CCJS26}.

The Fleming--Viot process is a component of the generalized parallel replica algorithm proposed in~\cite{BLS15,HL19}, where it is used to generate approximately independent samples from the QSD, as well as to define the decorrelation time~\(t_{\rm corr}\) as a stopping time for the natural filtration of~\(\bfX\).

\section{On-the-fly quasi-stationarity diagnostics}
\label{sec:qs_criteria}
In this section, we present the main methodological contribution of this paper, which is a simulation-based inference~\cite{CBL20,DBSMMGRLLMal25} method to define on-the-fly decorrelation times for Fleming--Viot processes.
The method combines the theoretical results of Section~\ref{sec:theory} with a Bayesian inference approach to the estimation of reference decorrelation times for the surrogate processes~$Z$ identified in the previous section (for some notion of reference decorrelation times to be defined below, see Equation~\eqref{eq:time_true}).

These two components play complementary roles: i) the results of Section~\ref{sec:theory} suggest that a good choice of collective variable allows to deduce Yaglom convergence~\eqref{eq:decorrelation_quantitative_definition} from the quasi-stationarity of a simpler process (the canonical model~\(Z\) on the smaller space~\(\cZ\)),
and ii) the Bayesian framework places a prior on the set of possible~\(Z\), and the decorrelation time is defined using posterior probabilities of decorrelation, given observations of Fleming--Viot approximations of~$\eta_t^Z \approx \xi_\star\eta_t^X$.

The only on-the-fly quasi-stationarity diagnostic proposed in the literature is, to the best of our knowledge, the MCMC-inspired diagnostic of~\cite{BLS15}.
As such, it will be our main point of comparison in our numerical experiments.

We first recall how this diagnostic is defined in Section~\ref{sec:gr}, and then present our simulation-based inference method in Section~\ref{sec:sbi}. We finally discuss our methodological assumptions in Section~\ref{sec:discussion_of_assumptions}.
\subsection{Gelman--Rubin diagnostic}
\label{sec:gr}
Given a real-valued collective variable \({\xi:\cX\to\cZ\subset\R}\) and a Fleming--Viot approximation~${\bf X} = (X^{(n)})_{1\leq n\leq N}$ of~$\eta_t^X$, the Gelman--Rubin diagnostic is defined using the variance ratio
\begin{equation}
    \label{eq:gelman_rubin}
    R_{T,\xi}^X = \frac{\mathrm{Var}\left(\xi_\star \frac1T\int_0^T \widehat{\eta}_{t,N}^X\,\d t\right)}{\frac1N\sum_{n=1}^N\mathrm{Var}\left(\xi_\star \frac1T\int_0^T \delta_{X_t^{(n)}}\,\d t\right)},
\end{equation}
denoting by~${\rm Var}(\pi)$ the variance of~$X\sim\pi$ for any univariate distribution~$\pi\in\cP(\R)$, and where we recall~$\widehat{\eta}_{t,N}^X$ is the empirical measure~\eqref{eq:fv_empirical_measure}.
The ratio~\eqref{eq:gelman_rubin} contains the variance of \(\xi\) under the ergodic average of the empirical measure (therefore the total occupation measure in~\(\xi\)) in the numerator, and the mean variance over one-replica occupation measures (within-trajectory occupation measures) in the denominator.
It is therefore simple to verify that~\(R_{T,\xi}^X\geq 1\), as the mean-of-variances in the denominator is necessarily smaller than the pooled variance in the numerator, by the law of total variance.

A closely related statistic was proposed by~\cite{GR92} to diagnose the convergence of Markov chain Monte Carlo algorithms in Bayesian inference, using multiple independent copies of the Markov chain instead of the Fleming--Viot process.

To define the Gelman--Rubin decorrelation time, we choose a finite set~\(\cO\) of collective variables, and for~\(\xi\in\cO\), we define the stopping time
\begin{equation}
    \label{eq:gr_cv_decorr_time}
    t^{\rm GR}_{{\rm corr},\xi}(\alpha) = \inf\{t\geq \tau:\,R_{t,\xi}^X-1<\alpha\},
\end{equation}
where~\(\alpha,\tau>0\) are user-provided parameters.

The Gelman--Rubin decorrelation time is finally defined as the maximum of these stopping times:
\begin{equation}
    \label{eq:gr_decorrelation_time}
    t^{\rm GR}_{\rm corr}(\alpha) = \underset{\xi\in \cO}{\max}\,t^{\rm GR}_{\rm corr,\xi}(\alpha).
\end{equation}

The parameter~\(\alpha\) is a tolerance threshold, which~\cite{BLS15,HL19} suggest should be taken such that~\(\alpha\in [0.01,0.1]\) for most systems, based on the results of some numerical experiments.
The parameter~\(\tau>0\) is a burn-in time, which is necessary to avoid spurious convergence: when the Fleming--Viot process starts from a deterministic state (\(\widehat{\eta}_{0,N}^X = \delta_{X_0}\)), this forces~\(R_{t,\xi}^X\approx 1\) for small times~\(t>0\).

Turning~\eqref{eq:gr_decorrelation_time} into a principled decorrelation time would rely on the following assumptions (which are adapted from those outlined in~\cite[Section 3.1]{BLS15}).
\begin{itemize}
    \item[\tagdef{A1}]{For large~\(N\) and~\(\xi\in\cO\), the empirical measure~\(\xi_\star\widehat{\eta}_{t,N}^X\) is, uniformly in time, a good approximation of~\(\xi_\star\eta_{t}^X\). This assumption is actually shared with our method (see Assumption~\tagref{B2} for a more formal discussion).}
    \item[\tagdef{A2}]{The trajectories~\(\xi(X^{(n)}_t)\) for~\(\xi\in\cO\) can be effectively treated as independent Markovian trajectories. This is the property which would justify the use of any Markov chain diagnostic.}
    \item[\tagdef{A3}]{There exists a slow~\(\xi\in\cO\), meaning that convergence of~\(\xi_\star\eta_t^X\) towards its quasi-stationary states implies convergence of~\(\eta_t^X\). This assumption is also shared by our method (and stated more precisely in Assumption~\tagref{B1}).}
\end{itemize}

The main assumption we avoid is~\tagref{A2}, which will be replaced by an assumption about prior knowledge of a set of plausible surrogate processes~$Z$ (this will be made precise in Assumptions~\tagref{B3},~\tagref{B4} below).

Even in the ideal situation where Assumptions~\tagref{A1}--\tagref{A3} hold, the stopping rule~\eqref{eq:gr_decorrelation_time} is known to have the wrong qualitative behavior. Whereas~\(\eta_T^X\) is generally known to converge to~\(\nu^X\) asymptotically exponentially fast as~\(T\to\infty\) (when a quantitative Yaglom limit such as~\eqref{eq:decorrelation_quantitative_definition} holds), the variance ratio~\eqref{eq:gelman_rubin} converges algebraically to~\(1\) as~\(T\to\infty\) (with a rate proportional to~\(1/T\)).
Therefore, when the tolerance threshold~\(\alpha\) is small, we expect the stopping rule~\eqref{eq:gr_decorrelation_time} to be overly conservative. This is broadly in line with what we observe numerically in Section~\ref{sec:num}, and the experiments of~\cite{BLS15,HL19}.

\subsection{Simulation-based inference of decorrelation times}
\label{sec:sbi}
In Section~\ref{sec:theory}, we identify a regime in which a coarse-grained Markov process, the canonical model~\(Z\) from Definition~\ref{def:perfect_model}, provides an accurate description of the convergence of~\(X\) to its QSD within~\(\cX\).
It is therefore tempting to use~\(Z\) to define a quasi-stationarity diagnostic. This is not a practical idea, as the very definition of \(Z\) requires knowledge of \(\nu^X\) through the definition of the fiber measures \(\cF\). As the purpose of a quasi-stationary diagnostic is ultimately to sample from \(\nu^X\), we face an apparently circular problem.

We argue that this circularity can be avoided by combining the approximation results of Section~\ref{sec:theory} with simulation-based Bayesian inference.
Our proposed method can be described loosely as follows. The parameter of interest we would like to infer is the ``honest'' decorrelation time~$t_0(\cL^Z,Z_0)$ of~$Z$ (to be defined precisely in Proposition~\ref{prop:exit_bias} below).
\begin{itemize}
    \item[$\circ$]{We place a prior distribution~$\rho_{\rm synth}$ on the dynamics of the surrogate process~$Z$, specified by an infinitesimal generator and an initial point~$(\cL^Z,Z_0)$, and estimate posterior probabilities~$h_t(F_t^Z):=\P_\rho{t>t_0(\cL^Z,Z_0)}[F_t^Z]$ for~$t\geq 0$, where~$F_t^Z$ are featurized observations of the paths~$(\widehat\eta_{s,N}^Z)_{0\leq s\leq t}$ of the particle-approximated probability flow of~$\eta_t^Z$.}
    \item[$\circ$]{Since~$(\cL^Z,Z_0)$ is marginalized over rather than inferred directly, it is a nuisance parameter in the sense of~\cite[Section 3.1]{GCSR95}, rather than the parameter of interest, which is limited to~$t_0(\cL^Z,Z_0)$.} 
    \item[$\circ$]{Because the conditional likelihood of $F_t^Z$ given $(\cL^Z,Z_0)$ is intractable, we train a machine learning model to approximate the posterior probabilities with a parametric model~$h_{\theta,t}$, following the standard simulation-based inference methodology~\cite{CBL20}.}
    \item[$\circ$]{In deployment, we evaluate the trained model on $F_t^{\xi,X}$, a featurized observation of $(\xi_\star\widehat\eta_{s,N}^X)_{0\leq s\leq t}$. That is, we use~$h_{\theta,t}(F_t^{\xi,X})$ as if~$F_t^{\xi,X}$ had been generated from the synthetic prior. This substitution is motivated by the analysis of Section~\ref{sec:theory}, and the underlying assumptions are made explicit in Section~\ref{sec:discussion_of_assumptions}.
    In particular, paths of~$X$ are never simulated during training, and paths of~$Z$ are never simulated during deployment.}
\end{itemize}

\subsubsection{Honest decorrelation times}
 The parameter of interest is the~\emph{honest decorrelation time}
\begin{equation}
    \label{eq:time_true}
    t_0(\cL^X,\eta_0^X) = \inf\{t>0:\,\d_{\rm TV}(\eta_t^X,\nu^X)<\varepsilon_{\rm corr}\},
\end{equation}
where~\(\varepsilon_{\rm corr}>0\) is a tolerance threshold, which will be fixed throughout this work. Note this is a function of the dynamical rule and initial distribution, not of the sample paths.

A direct estimator of~\(t_0(\cL^X,\eta_0^X)\) would be very valuable, as the knowledge of this time gives direct error control on sampled exit events starting from~\(t_0(\cL^X,\eta_0^X)\). We have the following result.
\begin{proposition}
    Writing~\(t_0^X:=t_0(\cL^X,\eta_0^X)\), we have, for any~\(t\geq 0\) and~\(A\subseteq\partial\cX\),
    \begin{equation}
        \left|\P_{\eta_{t_0^X}^X}{X_{\tau^X}\in A,\tau^X>t} - \P_{\nu^X}{X_{\tau^X}\in A,\tau^X>t}\right| \leq \varepsilon_{\rm corr}/2.
    \end{equation}
\end{proposition}
We omit the proof, as it is similar to the proof of Proposition~\ref{prop:exit_bias} below.

This quantity is however intractable for most systems of interest, as it requires knowledge of~\(\eta_t^X\) and~\(\nu^X\).
We therefore introduce a first layer of approximation, consisting in fixing a collective variable~\(\xi:\cX\to\cZ\) and considering instead the coarse-grained decorrelation time~\(t_0(\cL^Z,\eta_0^Z)\).
The analysis performed in Section~\ref{sec:theory} allows us to estimate the bias on the exit distribution~\eqref{eq:exit_distribution} resulting from this choice.
\begin{proposition}
    \label{prop:exit_bias}
    Under the Assumptions and notations of Theorem~\ref{thm:cg_existential_result}, writing~\(t_0^Z = t_0(\cL^Z,\eta_0^Z)\), we have, for any~\(t\geq 0\) and~\(A\subseteq\partial\cX\),
    \begin{equation}
        \begin{aligned}
             & \left|\P_{\eta_{t_0^Z}^X}{X_{\tau^X}\in A,\tau^X>t} - \P_{\nu^X}{X_{\tau^X}\in A,\tau^X>t}\right| \leq \frac{\varepsilon_{\rm corr}}{2}                   \\
             & \qquad\qquad+\frac{\|A^X\|}{2\zeta}\left(C\e^{-\gamma^X t_0^Z}+C't_0^Z\e^{-(\gamma^Z\land\gamma^X)t_0^Z}\right)                                           \\
             & \qquad\qquad+ \frac{C_\lambda^X C_\gamma^XC_\zeta}{2\eta_0^X(h^X)}\left(\e^{-(\zeta-\lambda^X) t_0^Z}+ 2\frac{\|A^X\|}{\zeta}\e^{-\gamma^X t_0^Z}\right).
        \end{aligned}
    \end{equation}
\end{proposition}
In the above,~\(t_0^Z=t_0(\cL^Z,\eta_0^Z)\) is of course the honest decorrelation time for the canonical model~\(Z\) associated to the choice of~\(X\) and~\(\xi\), given in Definition~\ref{def:perfect_model}.

\begin{proof}[Proof of Proposition~\ref{prop:exit_bias}]
    The function~\(f(x) = \P_x{X_{\tau^X}\in A,\tau^X>t}\) takes values in~\([0,1]\), which allows us to write
    \begin{equation}
        \begin{aligned}
            |\eta^X_{t_0^Z}(f)-\nu^X(f)| & = \left|\int_0^1 \left[\eta^X_{t_0^Z}(f>s) - \nu^X(f>s)\right]\,\d s\right|          \\
                                       & \leq \int_0^1 \underset{A\subseteq \cX}{\sup}\, |\eta^X_{t_0^Z}(A) - \nu^X(A)|\,\d s \\
                                       & = \frac12 \d_{\rm TV}(\eta_{t_0^Z}^X,\nu^X).
        \end{aligned}
    \end{equation}

    From here, the estimate~\eqref{eq:diagnostic_bound} from Corollary~\ref{cor:diagnostic_bound} gives
    \begin{equation}
        \begin{aligned}
             & \d_{\rm TV}(\eta_{t_0^Z}^X,\nu^X)\leq \d_{\rm TV}(\eta_{t_0^Z}^Z,\nu^Z) + \frac{C_\lambda^X C_\gamma^XC_\zeta}{\eta_0^X(h^X)}\left(\e^{-(\zeta-\lambda^X) t_0^Z}+ 2\frac{\|A^X\|}{\zeta}\e^{-\gamma^X t_0^Z}\right) \\
             & \qquad\qquad\qquad\quad+\frac{\|A^X\|}\zeta\left(C\e^{-\gamma^X t_0^Z}+C't_0^Z\e^{-(\gamma^Z\land\gamma^X)t_0^Z}\right)                                                                                             \\
             & \qquad\qquad\qquad\leq \varepsilon_{\rm corr} + \frac{C_\lambda^X C_\gamma^XC_\zeta}{\eta_0^X(h^X)}\left(\e^{-(\zeta-\lambda^X) t_0^Z}+ 2\frac{\|A^X\|}{\zeta}\e^{-\gamma^X t_0^Z}\right)                           \\
             & \qquad\qquad\qquad\quad+\frac{\|A^X\|}\zeta\left(C\e^{-\gamma^X t_0^Z}+C't_0^Z\e^{-(\gamma^Z\land\gamma^X)t_0^Z}\right),
        \end{aligned}
    \end{equation}
    by definition of~\(t_0^Z\) for the last inequality.
\end{proof}

This surrogate criterion is still impractical, since we do not know how to compute~\(\cL^Z\) given~\(\cL^X\) and~\(\xi\). Besides, we are not overly interested in measuring~\(\cL^Z\) directly. The key idea in this section is to treat~\(\cL^Z\) as a nuisance parameter in a probabilistic model.

\subsubsection{Synthetic decorrelation ensembles}

Our goal is to infer~\(t_0(\cL^Z,\eta_0^Z)\), which is equivalent to inferring the values of the variables
\begin{equation}
    \label{eq:decorr_events}
    D_t(\cL^Z,\eta_0^Z) = \{t>t_0(\cL^Z,\eta_0^Z)\},\qquad t\geq 0.
\end{equation}

We define a probabilistic model, specified by the following.
\begin{itemize}
    \item[\(\circ\)]{A prior~\(\rho_{\rm dyn} \in \cP({\rm Gen}(\cZ))\) over generators~\(\cL^Z\) of the coarse-grained dynamics~\(Z\), where~\({\rm Gen}(\cZ)\) denotes the set of generators of killed, conditionally irreducible continuous-time Markov jump processes on~\(\cZ\).}
    \item[\(\circ\)]{A prior~\(\rho_{0}\) over an additional nuisance parameter~\(Z_0\in\cZ\), whose role will be discussed below.}
    \item[\(\circ\)]{A choice of design variables, in this work an integer~\(N\geq 1\) and a lag time~\(\tau>0\).}
    \item[\(\circ\)]{An observation process, carrying information about the non-linear flow~\(\eta_t^Z\), in the form of a sequence of parametrized random variables
              \begin{equation}
                  \label{eq:noisy_observation}
                  (F(\cL^Z,Z_0,N,k\tau))_{1\leq k\leq \ell}.
              \end{equation}
              This observation provides the data in our Bayesian framework. The nature of this data is made precise below.
          }
\end{itemize}

Given these definitions, our goal is to estimate the posterior probabilities
\begin{equation}
    \label{eq:posterior_probabilities}
    h_k^*(\scF;N,\tau) := \P_{\rho_{\rm synth}}{D_{k\tau}(\cL^Z,\eta_0^Z)}[\left(F(\cL^Z,Z_0,N,j\tau)\right)_{1\leq j\leq k}=\scF],
\end{equation}
where~\(\rho_{\rm synth}\) is the joint prior over the nuisance parameters~\((\cL^Z,Z_0)\). Let us be more specific about the role played by these parameters and the observation process~\(F\).

\begin{itemize}
    \item[\(\circ\)]{The parameter~\(\cL^Z\) is the generator of the canonical model from Definition~\ref{def:perfect_model}, and the role of the prior~\(\rho_{\rm synth}\) is to model our uncertainty about which \(Z\) adequately approximates the decorrelation behavior of~\(\xi(X)\). The synthetic ensemble~\(\rho_{\rm dyn}\) should cover the set of plausible surrogate processes, and this is the main modeling input in our framework. It should ideally be calibrated with domain-specific knowledge to avoid overly broad priors.}
    \item[\(\circ\)]{The additional nuisance parameter is limited in this work to~\(Z_0 \in \cZ\), corresponding to the initial state of the surrogate process. The prior~\(\rho_0\) is therefore an element of~\(\cP(\cZ)\), and in this work we take a uniform prior~\(\rho_{0}=\cU(\cZ)\), with~\(\rho_{\rm synth}= \rho_{\rm dyn}\rho_{0}\), i.e. the dynamics and initial point are sampled independently. One could instead treat the initial distribution~\(\eta_0^Z \in \cP(\cZ)\) as a nuisance parameter, in which case~\(\rho_{0}\in\cP(\cP(\cZ))\), with the goal of making the model's predictions robust to changes of the initial distribution. Here we restrict ourselves to the special case \(\eta_0^Z = \delta_{Z_0}\) for simplicity.}
    \item[\(\circ\)]{The variable~\(F\) is a sequence of indirect and noisy observations related to the non-linear flow~\(\eta_t^Z\). The design parameter~\(N\) counts the number of replicas in a Fleming--Viot process (see Definition~\ref{def:fleming_viot}) on~\(\cZ\), killed at the boundary~\(\partial^\cZ\), with one-particle generator~\(\cL^Z\). The Fleming--Viot process is initialized at time~\(t=0\) from the state~\({\bfZ}_0=(Z_0,\dots,Z_0)\). The sample path of this particle process is then featurized into a sequence of probability distributions
              \begin{equation}
                  \label{eq:featurization}
                  F(\cL^Z,Z_0,N,k\tau) = {\rm Hist}\left(\frac1{\tau}\int_{(k-1)\tau}^{k\tau} \widehat{\eta}_{t,N}^Z\,\d t\right),
              \end{equation}
              where~\({\rm Hist}:\cP(\cZ)\to\cP(\cZ)\) is a histogram map associated to a fixed partition of~\(\cZ\). While other featurizations are possible, particle permutation-invariant features preserve the exchangeability symmetry of Fleming--Viot trajectories, which is likely important for parameter efficiency. The total observed data is therefore the sequence
              \begin{equation}
                  \label{eq:total_observation}
                  \left(F(\cL^Z,Z_0,N,k\tau)\right)_{1\leq k\leq \ell}\in \bigcup_{\ell\geq 1} \cP(\cZ)^\ell.
              \end{equation}
              Note that each observation sequence is finite, corresponding to the fact that Fleming--Viot simulations can only be performed over finite time horizons. The length~\(\ell\) of each observation is determined by the learning objective, which will be discussed later.

              In practice, trajectories and time integrals are discretized consistently with a numerical scheme, and~\(\tau = N_{\rm lag}\Delta t\) is an integer multiple of the timestep~$\Delta t$, so that each histogram in~\eqref{eq:featurization} aggregates~\(NN_{\rm lag}\) sample points in total.
          }
\end{itemize}

Let us make some remarks on this model.
\begin{itemize}
    \item[\(\circ\)]{An alternative (which we leave to future work) to the hand-calibration of~\(\rho_{\rm synth}\) is to learn a generative model~\(\rho_{\rm synth}\) for~\((\cL^Z,Z_0)\) from real simulation data. More precisely, one would tune~\(\rho_{\rm synth}\) to make the paths of generated features~\(F\) indistinguishable from those obtained from high-dimensional Fleming--Viot processes (i.e. derived from low-dimensional traces \((\xi_\star \widehat{\eta}_{t,N}^X)_{t\geq 0}\)), for practical choices of the collective variable~\(\xi\). Classification-based approaches for calibrating parameters in synthetic-to-real distribution-matching problems are available in the literature, see~\cite{GDKC18} and references therein.
              Doing so with gradient-based methods requires computing derivatives of Fleming--Viot trajectories with respect to~\(\cL^Z\).}
    \item[\(\circ\)]{The total observed sequence~\(F\) is a noisy and biased observation of the non-linear flow~\(\eta_t^Z\), polluted by several sources of error, first and foremost stochastic error terms and biases related to the finite particle number~\(N\) in the Fleming--Viot process. Training the inference model using imperfect features is consistent with the operational setting of molecular simulation, where these error sources are also present.}
    \item[\(\circ\)]{The likelihood of~\(F\) under~\(\rho_{\rm synth}\) is analytically intractable, and therefore we rely on simulation-based (also known as likelihood-free) estimators of the posterior probabilities~\eqref{eq:posterior_probabilities} as will be explained in Section~\ref{sec:training}, which require access to the ground-truth labels~\(D_{k\tau}(\cL^Z,\eta_0^Z)\). This implies that the distribution of synthetic dynamics~\(\rho_{\rm dyn}\) should be supported on generators~\(\cL^Z\) for which computing these labels is tractable, at least as an offline data-generation cost.}
    \item[\(\circ\)]{Instead of fixing the tolerance parameter~\(\varepsilon_{\rm corr}>0\), it could be treated as a design parameter in future work, which could be useful to tune the error level with no need for retraining.}
\end{itemize}

\subsubsection{Recurrent model}
\label{sec:rnn}
To approximate the posterior probabilities~\eqref{eq:posterior_probabilities}, we propose a simple class of parametric models acting on the space of sequences~\eqref{eq:featurization}, which will be used to detect quasi-stationarity on-the-fly, as the Fleming--Viot simulation proceeds.
The simulation produces a sequence of histograms in~$\xi(\cX)=\cZ$ (the collective variable space), defined in~\eqref{eq:featurization}. Denoting the observation~\eqref{eq:total_observation}
\begin{equation}
    F = (H_k)_{1\leq k\leq \ell},\qquad H_k = F(\cL^Z,Z_0,N,k\tau),
\end{equation}
this sequence has important properties informing our choice of architecture.

First, it is not known in full at deployment time, but streamed as the simulation proceeds and samples are aggregated. For this reason, the final length~\(\ell\) is not known a priori, and moreover may vary over several orders of magnitude throughout the synthetic ensemble (in accordance with the fact that real systems do decorrelate on a range of different timescales).

A naive approach consists in padding the sequence (say with zeros) into a fixed-size~\(\ell_{\max}\)-length buffer, and feeding a flattened~\(\ell_{\max}d_{\rm hist}\)-dimensional feature to a parametric model operating on fixed-size inputs (where~\(d_{\rm hist}\) gives the dimension of a single histogram), for instance
\begin{equation}
    \underbrace{F}_{\ell\times d_{\rm hist}} \to \underbrace{F_{\rm padded}}_{\ell_{\max}\times d_{\rm hist}} \to \underbrace{h_\theta(F_{\rm padded};N,\tau)}_{\in [0,1]^{\ell_{\rm max}}},
\end{equation}
where~\(h_\theta\) is a dense neural network parametrized by some~\(\theta\in \R^p\). The first~\(\ell\) elements of the output can be taken as predictors of the posterior probabilities~\eqref{eq:posterior_probabilities}. This approach has several drawbacks: since inference cost for dense neural networks is quadratic in the dimensionality of input features, and inference has to be run every timestep (or every few timesteps), this strategy leads in total to a~\(\bigo(\ell_{\rm max} d_{\rm hist})\) memory cost and~\(\bigo(\ell\ell_{\rm max}^2 d_{\rm hist}^2)\) total inference cost for a length-$\ell$ sequence.
This may become very costly for long Fleming--Viot traces, and besides requires fixing~\(\ell_{\max}\) a priori (and pessimistically), which leads to wasted computation on short traces.
Moreover, we would like our model for~\(h^*\) to be causal, meaning that future observations may not influence past predictions. Indeed, it is clear from the definition of~\(F\) and the posterior probabilities~\eqref{eq:posterior_probabilities} that
\begin{equation}
    \label{eq:causality}
    \forall 1\leq k\leq \ell,\qquad h^*_k(F) = g_k(H_1,\dots,H_k)
\end{equation}
for some measurable functions~\(g_k:\R^{d_{\rm hist}\times k} \to [0,1]\). This property has to be enforced manually in training this type of model. While more advanced architectures (such as autoregressive Transformers) naturally address some of these issues, these are still limited by memory constraints to relatively short sequences.
Such models are besides poorly suited to the operational constraints of quasi-stationarity diagnostics, which should remain lightweight and deployable on standard CPU computing clusters, with negligible memory cost.

We propose using recurrent neural networks (RNNs), a classical approach in dealing with sequence-to-sequence learning tasks, and which address these shortcomings by construction. These allow for~\(\bigo(d_{\rm hist})\) memory cost and~\(\bigo(\ell d_{\rm hist}^2)\) total inference cost. Since the inference cost per simulation step is constant, they can feasibly be used on arbitrarily long sequences, and besides satisfy the causality property~\eqref{eq:causality} automatically. Recurrent neural networks process the sequence element-by-element, building a sequence of latent states~\((L_k(\theta;N,\tau))_{0\leq k\leq\ell}\) recursively from~\((H_k)_{1\leq k\leq \ell}\) as follows:
\begin{equation}
    \label{eq:recurrent_update}
    \forall\,1\leq k\leq\ell,\qquad L_k(\theta;N,\tau) = r_\theta(H_k,L_{k-1};N,\tau),
\end{equation}
where the recurrent layer is a parametrized map~\(r_\theta:\R^{d_{\rm in}}\times\R^{d_{\rm rnn}}\to\R^{d_{\rm rnn}}\), and~\(L_0\in \R^{d_{\rm rnn}}\) is an arbitrary latent initial state, which we take to be~\(L_0=0\) (\(L_0\) could also be treated as a trainable function of~$Z_0$). Our recurrent model for~\(h^*\) is finally defined as
\begin{equation}
    \label{eq:classification_head}
    \forall\,1\leq k\leq\ell,\qquad h_{k,\theta}(F) = c_\theta(L_k(\theta;N,\tau)),
\end{equation}
where~\(c_\theta:\R^{d_{\rm rnn}}\to [0,1]\) is a classifier acting on RNN latent states. In this work, \(c_\theta\) is a shallow, dense neural network.

We define the recurrent component of the network by
\begin{equation}
    \label{eq:recurrent_layer}
    r_{\theta}(H,L;N,\tau) = {\rm LSTM}_{\theta}(f_\theta(H)\star m(N,\tau),L),
\end{equation}
where~\(f_\theta:\R^{d_{\rm hist}}\to \R^{d_{\rm cnn}}\) is a~\((N,\tau)\)-independent featurizer,~\(m(N,\tau):\R^2\to \R^{d_{\rm meta}}\) is a hand-defined feature encoding design information, \({\rm LSTM}_\theta:\R^{d_{\rm cnn} + d_{\rm meta}}\times\R^{d_{\rm rnn}}\to \R^{d_{\rm rnn}}\) is a stack of one or several long short-term memory (LSTM) layers, and \(\star : \R^{d_{\rm cnn}}\times \R^{d_{\rm meta}}\to \R^{d_{\rm cnn}+d_{\rm meta}}\) is the concatenation operator.
LSTMs are a classical class of recurrent models introduced for processing sequential data, and are an available option in most neural network packages today. Although modern alternatives to LSTMs exist, we found these to be sufficient for the purposes of this work.

We choose~\(f_\theta\) to be a convolutional neural network (CNN) operating on histograms, which is a parameter-efficient architecture suitable for feature-extraction in spatially structured low-dimensional signals.
The fact that our histogram features are spatially structured is not yet obvious: this will become apparent when we consider choices of~\(\xi\) related to molecular collective variables (see Section~\ref{sec:num} below).

Finally, the recurrent layer is informed of the design variables only through the scalar feature~\(m(N,\tau)=\sqrt{N\tau}\), so that~$d_{\rm meta}=1$. This choice is physically motivated by the fact that~\(\sqrt{N\tau}\) is a natural precision scaling for the histogram features~\eqref{eq:featurization}. We therefore expect that supplying this quantity to the recurrent model will make it more effective at filtering the finite-sample statistical noise out of its latent state trajectories.
One could also supply~$m(N,\tau)=(N,\tau)$ directly, but we deliberately chose not to do this in order to avoid overfitting to the elapsed time of the simulation. The influence of this choice has not been tested systematically.

One could also ask whether it is possible to directly approximate the posterior probability~$\P_{\rho_{\rm synth}}{D_{k\tau}(\cL^Z,\eta_0^Z)}[H_k]$ rather than~\eqref{eq:posterior_probabilities}. This target, however, discards our knowledge of~$(H_1,\dots,H_{k-1})$ when we have observed~$H_k$. In Section~\ref{sec:num_synth}, we test a diagnostic based on the joint observation of~$(H_k,\nu^Z)$, and find that it performs poorly when compared to the recurrent model. Although~$H_k$ does depend implicitly on~$(H_1,\dots,H_{k-1})$, it may be too noisy in practice to be a useful predictor. 

\subsubsection{Training objective}
\label{sec:training}
For any set~\(E\), we denote~\(\cS(E)=E^\N\) the set of~\(E\)-valued sequences. The posterior probabilities~\eqref{eq:posterior_probabilities} we wish to estimate define a causal, measurable function~\(h^*(\cdot\,;N,\tau):\cS(\R^{d_{\rm hist}})\to \cS([0,1])\) mapping sequences to sequences.

For any summable weighting function~\(w\in \ell^1(\N,\R_+)\), we introduce a weighted cross-entropy functional
\begin{equation}
    \label{eq:conditional_loss_functional}
    \scL_w[h;N,\tau] = \E_{\rho_{\rm synth}}{\E_{{\bf Z}}{\sum_{k\geq 1} w(k)\bbH\left\{D_{k\tau}(\cL^Z,Z_0),h_k(F)\right\}}},
\end{equation}
where the outer expectation is with respect to the prior~\(\rho_{\rm synth}\) over~\((\cL^Z,Z_0)\), while the inner expectation is over the distribution of features~\(F\) induced by the Fleming--Viot process~\(\bfZ\).
We recall that the integer~\(\ell\) is the length of the feature sequence~\(F\), and that the variables~\(D_{k\tau}\) defined in~\eqref{eq:decorr_events} are~\(\{0,1\}\)-valued. We write
\begin{equation}
    \label{eq:bce}
    \bbH\left\{p,q\right\} = -p\log\,q -(1-p)\,\log(1-q)
\end{equation}
for the cross-entropy of a Bernoulli distribution with respect to another, with parameters~\(p\) and~\(q\).

This class of objectives is naturally motivated by the following lemma.
\begin{lemma}
    \label{lemma:ideal_loss_minimizer}
    For any choice of weight~\(w\in\ell^1(\N,\R_+)\), design parameters~\(N\geq 1\),~\(\tau>0\), and any causal~\(h:\cS(\R^{d_{\rm hist}})\to \cS([0,1])\) (that is, verifying~\eqref{eq:causality}), it holds that
    \begin{equation}
        \scL_w[h;N,\tau]\geq \scL_w[h^*;N,\tau]
    \end{equation}
    with equality if and only if~\(h_k = h_k^*\) almost surely for every~\(k\) such that~\(w(k)>0\).
\end{lemma}
\begin{proof}
    Denote~\(\bbE = \bbE_{\rho_{\rm synth}} \bbE_{\bfZ}\), and~\(\cF_{k} = \sigma(H_j,\,1\leq j\leq k)\) be the natural filtration for the observed data. Let~\(h:\cS(\R^{d_{\rm hist}})\to \cS([0,1])\) be a causal measurable function. Then,
    \begin{equation}
        \begin{aligned}
            \scL_w[h;N,\tau] & = \E{\sum_{k\geq 1} w(k)\bbH\left\{D_{k\tau}(\cL^Z,Z_0),h_k(F)\right\}}              \\
                             & = \sum_{k\geq 1} w(k)\E{\E{\bbH\left\{D_{k\tau}(\cL^Z,Z_0),h_k(F)\right\}}[\cF_{k}]} \\
                             & = \sum_{k\geq 1} w(k)\E{\bbH\left\{\E{D_{k\tau}(\cL^Z,Z_0)}[\cF_{k}],h_k(F)\right\}} \\
                             & = \sum_{k\geq 1} w(k)\E{\bbH\left\{h_k^*(F),h_k(F)\right\}}                          \\
                             & \geq \sum_{k\geq 1}w(k)\E{\bbH\left\{h_k^*(F),h_k^*(F)\right\}}                     \\
                             & =\scL_w[h^*;N,\tau],
        \end{aligned}
    \end{equation}
    where we used the causality property~\eqref{eq:causality} and the fact that~\(\bbH\) is affine in its first argument for the third equality, the definition of~\(h^*\) for the fourth equality, and Gibbs' cross-entropy inequality in the fifth line. This inequality saturates if and only if~\(h_k=h_k^*\) whenever~\(w(k)>0\), which is our claimed equality condition.
\end{proof}

When~\(w\) has infinite support, evaluating this functional requires simulating infinitely long Fleming--Viot trajectories. A simple idea to avoid this consists in truncating the sum using a deterministic weight~\(w(k)=\1_{k\leq \ell_{\max}}\) for some large~\(\ell_{\max}\gg 1\).
This requires setting a time-horizon a priori for the diagnostic, which we wished to avoid by using a recurrent architecture.

We propose another approach using a randomized time-horizon. Given~\((\cL^Z,Z_0)\sim\rho_{\rm synth}\), we sample a random variable~\(L\sim \pi_L\left(\cdot\,\middle|t_0(\cL^Z,Z_0),N,\tau\right)\in \cP(\N)\), and consider the objective
\begin{equation}
    \label{eq:randomized_loss_functional}
    \scL_{L}[h;N,\tau] = \E_{\rho_{\rm synth}}{\E_L{\E_{{\bf Z}}{\sum_{1\leq k\leq L} \bbH\left\{D_{k\tau}(\cL^Z,Z_0),h_k(F)\right\}}}}.
\end{equation}
Then we have the following result.
\begin{lemma}
    \label{lemma:randomized_loss_minimizer}
    For~\(N\geq 1\),~\(\tau>0\), and any causal~\(h:\cS(\R^{d_{\rm hist}})\to \cS([0,1])\), it holds that
    \begin{equation}
        \label{eq:randomized_loss_inequality}
        \scL_L[h;N,\tau]\geq \scL_L[h^*_L;N,\tau],
    \end{equation}
    where
    \begin{equation}
        \label{eq:randomized_loss_minimizer}
        h^*_{L,k}(F) = \P_{\rho_{\rm synth},L}{\tau k > t_0(\cL^Z,Z_0)}[H_1,\dots,H_k,k\leq L].
    \end{equation}

    Moreover, if~\(L\) is chosen such that~\(\tau L > t_0(\cL^Z,Z_0)\) almost surely, then
    \begin{equation}
        \label{eq:risk_inequality}
        h_{L,k}^*(F) \leq h_k^*(F)\qquad \bbP_{\rho_{\rm synth},L,\bfZ}\text{-almost surely}.
    \end{equation}
\end{lemma}
\begin{proof}
    The inequality~\eqref{eq:randomized_loss_inequality} is proven in the same way as Lemma~\ref{lemma:ideal_loss_minimizer}, by writing~\(\sum_{1\leq k\leq L} = \sum_{k\geq 1}\1_{k\leq L}\) and conditioning on~\(\sigma(H_1,\dots,H_k,\1_{k\leq L})\) rather than~\(\cF_k\).

    For the inequality~\eqref{eq:risk_inequality}, write, with~$\rho=\rho_{\rm synth}$,
    \begin{equation}
        \begin{aligned}
            1-h_{L,k}^*(F) & = \P_{\rho,L}{\tau k \leq t_0(\cL^Z,Z_0)}[H_1,\dots,H_k,k\leq L]                                       \\
                           & = \frac{\P_{\rho,L}{\tau k \leq t_0(\cL^Z,Z_0),\tau k\leq \tau L}[\cF_k]}{\P_{\rho,L}{k\leq L}[\cF_k]} \\
                           & =\frac{\P_{\rho}{\tau k \leq t_0(\cL^Z,Z_0)}[\cF_k]}{\P_{\rho,L}{k\leq L}[\cF_k]}                      \\
                           & \geq \P_{\rho}{\tau k \leq t_0(\cL^Z,Z_0)}[\cF_k]                                                      \\
                           & =1-h_k^*(F),
        \end{aligned}
    \end{equation}
    using~\(\tau L > t_0(\cL^Z,Z_0)\) in the third line, and~\(0\leq\P_{\rho_{\rm synth},L}{k\leq L}[\cF_k]\leq 1\) in the fourth line.
\end{proof}
The inequality~\eqref{eq:risk_inequality} means that the minimizer~\(h_{L}^*\) is conservative, as it underestimates the posterior probabilities~\eqref{eq:posterior_probabilities} of successful decorrelation.
In practice, we find that enforcing the condition~\(\tau L>t_0(\cL^Z,Z_0)\) via the rule
\begin{equation}
    \label{eq:horizon_sampling_rule}
    L = \lceil R t_0(\cL^Z,Z_0)/\tau \rceil,\qquad R\sim \cU(1,2)\text{ independently}
\end{equation}
is reasonable, as it results in a fairly balanced set of positive and negative observations, and naturally scales the simulation cost with the physical decorrelation time, thereby avoiding the cost of running long stationary simulations.

It is also simple to show that the ideal predictor~\(h^*_L\) keeps the prior-averaged risk of false decorrelation below~\(\alpha\), conditionally on a finite detection time.
\begin{lemma}
    \label{lemma:risk_control}
    Let~\(h^*_L\) be defined as in~\eqref{eq:randomized_loss_minimizer} (with the randomized horizon~\(L\) satisfying the condition for~\eqref{eq:risk_inequality}).
    Define the decorrelation frame
    \begin{equation}
        K(\alpha) = \inf\left\{k\geq 1:\,\,1-h^*_{L,k}(F) < \alpha\right\}
    \end{equation}
    and the false decorrelation event
    \begin{equation}
        A(\alpha) = \left\{D_{K(\alpha)\tau}(\cL^Z,Z_0)=0\right\},
    \end{equation}
    recalling the definition~\eqref{eq:decorr_events}. Then
    \begin{equation}
        \label{eq:false_detection_inequality}
        \P_{\rho_{\rm synth}}{A(\alpha)}[K(\alpha)<\infty]\leq \alpha.
    \end{equation}
\end{lemma}
\begin{proof}
    We use the notation introduced in the proofs of Lemmas~\ref{lemma:ideal_loss_minimizer} and~\ref{lemma:randomized_loss_minimizer}, and denote~$\rho=\rho_{\rm synth}$. Note that~\(K(\alpha)\) is an~\(\cF_k\)-stopping time, and write
    \begin{equation}
        \begin{aligned}
            \P_{\rho}{A(\alpha),K(\alpha)<\infty} & = \sum_{k\geq 1}\E_\rho{\1_{K(\alpha)=k} \1_{A(\alpha)}}                                     \\
                                                  & = \sum_{k\geq 1}\E_\rho{\1_{K(\alpha)=k}\E_\rho{\1_{D_{K(\alpha)\tau}(\cL^Z,Z_0)=0}}[\cF_k]} \\
                                                  & = \sum_{k\geq 1}\E_\rho{\1_{K(\alpha)=k}\P_\rho{D_{k\tau}(\cL^Z,Z_0)=0}[\cF_k]}              \\
                                                  & = \sum_{k\geq 1}\E_{\rho}{\1_{K(\alpha)=k}\left(1-h_k^*(F)\right)}                           \\
                                                  & \leq \sum_{k\geq 1}\E_{\rho}{\1_{K(\alpha)=k}\left(1-h_{L,k}^*(F)\right)}                    \\
                                                  & \leq \alpha\sum_{k\geq 1}\P_\rho{K(\alpha)=k}                                                \\
                                                  & \leq \alpha\P_\rho{K(\alpha)<\infty},
        \end{aligned}
    \end{equation}
    and dividing gives the result. We use the stopping-time property in the second line, the definition of~\(h^*\) in the fourth line, the inequality~\eqref{eq:risk_inequality} in the fifth line, and the definition of~\(K(\alpha)\) in the sixth line.
\end{proof}

To make the predictions robust to the choice of lag time and number of replicas, we finally introduce an additional distribution over the design variables~\((N,\tau)\), and following the standard machine learning approach, we replace the function~\(h\) by the parametric ansatz of the previous subsection, which leads to the problem
\begin{equation}
    \label{eq:ml_problem}
    \underset{\theta\in\R^p}{\min}\,\E_{N,\tau}{\scL_L[h_\theta ; N,\tau]}.
\end{equation}

Given an \emph{i.i.d.} training batch~\(B=\left(t_0(\cL^Z_i,Z_{0,i}),F_i,L_i,N_i,\tau_i\right)_{1\leq i\leq b}\), we approximate a descent direction for the problem~\eqref{eq:ml_problem} with
\begin{equation}
    \label{eq:stochastic_gradient_estimator}
    \widehat{g}_B(\theta) = \frac1{C_B}\sum_{1\leq i\leq b}\sum_{1\leq k\leq L_i}\nabla_\theta \bbH\left\{D_{k\tau_i}(\cL^Z_i,Z_{0,i}),h_{\theta,k}(F_i)\right\},
\end{equation}
where~\(C_B = \sum_{1\leq i\leq b} L_i\) is a consistent estimator of
\begin{equation}
    \frac1{\E_{\rho_{\rm synth}}{L}}\nabla_\theta\E_{N,\tau}{\scL_L[h_\theta ; N,\tau]}
\end{equation}
in the large batch limit~\(b\to\infty\). The estimator~\(\widehat{g}_B(\theta)\) is then used in a stochastic gradient method (the Adam optimizer~\cite{KB14} in this work) to iteratively update the parameters.

\subsubsection{On-the-fly diagnostic}
Once the parameter~\(\theta\in\R^p\) is chosen by the optimization procedure, the quasi-stationarity diagnostic is deployed as follows. The parameters~\(N,\tau\) are set by the practitioner, and therefore known.
We construct, as the simulation of Fleming--Viot process~\(\bf X\) proceeds, and at each multiple~\(k\tau\) (\(k\geq 1\)) of the lag time, the histogram
\begin{equation}
    \label{eq:deployment_histograms}
    H_k^{\xi} =  \mathrm{Hist}\left( \frac1{\tau}\int_{(k-1)\tau}^{k\tau} \xi_\star\widehat{\eta}_{t,N}^X\,\d t\right) \in \R^{d_{\rm hist}},
\end{equation}
where we recall~\(\widehat{\eta}_{t,N}^X\) is the empirical measure of~\(\bf X\).
This only requires accumulating bin counts for values of~\(\xi\) over the~\(N\) replicas of~\(\bf X\) and the time interval~\([(k-1)\tau,k\tau]\).

We then use this histogram to update the model's latent state and predict, according to Equations~\eqref{eq:recurrent_update} and~\eqref{eq:classification_head},
\begin{equation}
    \label{eq:predicted_probabilities}
    h_{k,\theta}^{\xi} = c_{\theta}(L_k(\theta;N,\tau)),\qquad L_k(\theta;N,\tau) = r_{\theta}(H_k^{\xi},L_{k-1};N,\tau).
\end{equation}

The RNN decorrelation time is finally defined as
\begin{equation}
    \label{eq:rnn_decorrelation_time}
    t_{\rm corr}^{\rm RNN}(\alpha) = \tau\times\inf\,\left\{k\geq 1:\,h_{k,\theta}^{\xi}>1-\alpha\right\},
\end{equation}
where~\(\alpha\in (0,1)\) is a user-defined threshold. As in the Gelman--Rubin approach, one could conceivably apply the RNN to multiple collective variables~$\xi\in\cO$ (see Equation~\eqref{eq:gr_decorrelation_time}).
This is however only heuristic if the boundary compatibility condition~$\xi(\partial^\cX)=\partial^\cZ$ is not enforced simultaneously for all~$\xi\in\cO$ (we perform such an experiment involving multiple collective variables in Section~\ref{sec:num_dimer} below).

We conclude this section by clearly stating the set of assumptions justifying the deployment of the synthetically-trained model on the non-linear, non-Markovian particle approximation~\(\xi_\star\widehat{\eta}_{t,N}^X\).

\subsection{Discussion of underlying assumptions}
\label{sec:discussion_of_assumptions}
The transferability of the RNN diagnostic rests on several approximations, which we make explicit, as some are only empirically supported at this stage.
\begin{itemize}
    \item[\tagdef{B1}]{The collective variable~\(\xi\) is slow, i.e. the assumptions of Theorem~\ref{thm:cg_existential_result} are satisfied with~\(\|A^X\|/\zeta\) small.}
    \item[\tagdef{B2}]{The histogram trajectories of the Fleming--Viot approximations~\(\left(\rm{Hist}\left(\xi_\star \widehat{\eta}_{t,N}^X\right)\right)_{t\geq 0}\) converge to those of~\((\xi_\star \eta_t^X)_{t\geq 0}\) as~\(N\to\infty\), uniformly in time. The same property holds for the~\(Z\)-counterparts of these objects.}
    \item[\tagdef{B3}]{The decorrelation time~\(t_0(\cL^Z,Z_0)\) is, on average over the prior~\(\rho_{\rm synth}\), accurately predicted by the histogram trajectory~\(\left(\rm{Hist}(\eta_t^Z)\right)_{t\geq 0}\).}
    \item[\tagdef{B4}]{The synthetic prior places sufficient mass near the canonical model~\((\cL^Z,Z_0)\) of the system at deployment time.}
\end{itemize}

Assumption~\tagref{B1} justifies the approximation~\(\xi_\star\eta_t^X\approx \eta_t^Z\), uniformly in~\(t\geq 0\), in the sense described by the first item in Theorem~\ref{thm:cg_existential_result}.
In a second step,~\tagref{B2} then justifies the approximation~\({{\rm Hist}\left(\xi_\star\widehat{\eta}_{t,N}^X\right)\approx {\rm Hist}\left( \xi_\star\eta_t^X\right)\approx {\rm Hist}\left(\eta_t^Z\right)}\), again uniformly in time.
Assumptions~\tagref{B3} and~\tagref{B4} then imply that trajectories of \({\rm Hist}\left(\xi_\star\widehat{\eta}_{t,N}^X\right)\) are highly informative about~\(t_0(\cL^Z,Z_0)\), and therefore about the sampling bias on the exit event, according to Proposition~\ref{prop:exit_bias} and~\tagref{B1}.

The validity of the approximation warranted by Assumption~\tagref{B1} is settled for the case of finite-state jump processes~\(X\) by the results of Section~\ref{sec:theory}. The extension to the case of diffusion processes~\(X\) killed at the boundary of a domain (which is the relevant setting for MD) is still open, and will be studied in an upcoming work. The case of jump processes however already covers relevant physical models from atomistic simulation, such as kinetic Monte Carlo models.
Similarly, the justification of~\tagref{B2} requires uniform-in-time propagation of chaos results for Fleming--Viot processes. While such results are known to hold when the underlying generator is related to a jump process (see for example~\cite{CT16}), the strongest results available for diffusion processes are for now restricted to particular regimes~\cite{JR24,JM25,CCJS26}, and current mathematical understanding falls short of treating the case of general diffusions killed at a boundary.
We expect similar results to hold in very general situations.

The training procedure described in Section~\ref{sec:sbi} is designed, to some degree, to address the finite-\(N\) sampling approximation. Indeed, the recurrent model is trained to predict properties of~\(\eta_t^Z\) using observations of~\(\widehat{\eta}_{t,N}^Z\), and can therefore be understood as a particle filter with a latent model of this error, fitted during training by observing a range of~\(N\) and~\(\tau\).
We observe empirically that the optimized RNN is quite robust to the choice of~\((N,\tau)\), and that the performance of the model improves with~\(N\) (see Figure~\ref{fig:synth_auc_grid}).
This does not immediately dissolve the need for Assumption~\tagref{B2}, as it is unclear whether this finite~\(N\)-error model, fitted on~\(\widehat{\eta}_{t,N}^Z\to\eta_t^Z\) discrepancies, is well calibrated to the~\(\xi_\star \widehat{\eta}_{t,N}^X\to\xi_\star\eta_t^X\) discrepancy faced at deployment. One way to address this is to prove coarse-graining results at the level of the Fleming--Viot process~\(\bf X\) itself, rather than in the mean-field description of Section~\ref{sec:theory}.
Another way is to assume~\tagref{B2} and take~\(N\) sufficiently large so that the mean-field description is accurate.

In Section~\ref{sec:perspectives}, we discuss some strategies for relaxing some of these assumptions in future studies (in particular those pertaining to the choice of collective variable~\(\xi\) \tagref{B1} and prior~\(\rho_{\rm synth}\) in \tagref{B4}).

Regardless of whether~\tagref{B1}--\tagref{B4} hold, the stopping times~\eqref{eq:gr_decorrelation_time} and~\eqref{eq:rnn_decorrelation_time} are well-defined diagnostics: each one aggregates the history of the Fleming--Viot process into a decision to stop the simulation, and its usefulness can be assessed empirically.
For instance: what is the bias in sampled exit events, starting from a Fleming--Viot particle stopped at the diagnostic time, relative to a true sample from the QSD? How much does the diagnostic overestimate the honest decorrelation time, compared to some idealized ``ground truth'' estimate? How sensitive is the diagnostic to the choice of parameters~\((N,\tau,\alpha)\), and can these be chosen a priori over a wide physical range, without requiring per-system calibration?
We will explore these questions numerically in Section~\ref{sec:num} below.

When Assumptions~\tagref{B1}--\tagref{B4} do hold, the threshold~\(\alpha\) becomes interpretable in the RNN decision rule. As~\(h_{k,\theta}^{\xi}\) approximates a lower bound on the posterior probability~\eqref{eq:posterior_probabilities} according to Lemma~\ref{lemma:randomized_loss_minimizer}, the diagnostic rule~\eqref{eq:rnn_decorrelation_time} sets a prior-averaged risk of false convergence less than~\(\alpha\) (see Lemma~\ref{lemma:risk_control} above). This in turn gives (through Proposition~\ref{prop:exit_bias}) control on the average bias in sampled exit events under this rule. In this ideal situation,~\(\alpha\) can therefore be selected a priori from a target error level rather than tuned empirically. This is in contrast to the threshold~\(\alpha\) in~\eqref{eq:gr_cv_decorr_time}, where the variance ratio~\eqref{eq:gelman_rubin} has no comparable probabilistic interpretation, and may require system-by-system tuning.

We finally note that a caveat applies when using the RNN diagnostic to treat the decorrelated Fleming--Viot population as independent samples from the QSD (as one does in the generalized parallel replica algorithm), as the particles are correlated. However, these correlations are known to be weak if propagation of chaos holds. Besides, the same caveat applies to the Gelman--Rubin diagnostic (we refer to~\cite[Section 3.1]{BLS15} for various strategies addressing this concern).

\section{Numerical experiments}
\label{sec:num}
We present various experiments to delineate the scope of the method developed in Section~\ref{sec:qs_criteria}. In Section~\ref{sec:num_perturbation_theory}, we present direct numerical evidence supporting the theoretical findings of Section~\ref{sec:theory}. In Section~\ref{sec:num_implementation}, we provide details on our implementation of the RNN diagnostic of Section~\ref{sec:qs_criteria}. In Section~\ref{sec:num_synth}, we benchmark the performance of our RNN diagnostic on synthetic examples, comparing it to the Gelman--Rubin diagnostic from~\cite{BLS15}.
We also explore the reliability of the diagnostic out-of-distribution, when the training prior is deliberately misspecified. We finally investigate in Section~\ref{sec:num_dimer} how the diagnostic behaves in transfer, by applying it to a high-dimensional molecular system, contrasting again its behavior with the Gelman--Rubin diagnostic, in a case where~\(\xi\) is a slow variable of the system, and in a case where it is not.
All the code required to reproduce these results is available in a public repository~\cite{repo}, and public copies of processed simulation data will be made available in the short future.

\subsection{Validation of theoretical predictions}
\label{sec:num_perturbation_theory}

We validate the theoretical claims on the simplest non-trivial example, where \(|\cX|=4\) and the fibers of~\(\xi\) (as defined in~\eqref{eq:cv_discrete}) satisfy \(|\cX_1| = |\cX_2| = |\cZ|= 2\). We consider the small-parameter chain whose generator is given by~\eqref{eq:jp_generator}.

To investigate the effect of timescale separation on decorrelation, we randomly sample~\(\cL_{\rm mac},\cL_{\rm mic}\) and~\(K\) in~\eqref{eq:jp_generator}, and let the scale-parameter~\(\varepsilon\) vary.

Every jump rate (prior to any~\(\varepsilon\)-rescaling) is chosen independently at random, uniformly in the interval~\([1,5]\).
The prefactors \(C \in\{C_\zeta,C_\lambda^Y,C_\gamma^Y\}\), associated respectively with the propagators
\begin{equation}
    P_t\in\{\e^{t\Pi_1\cL^X\Pi_1}\Pi_1,\e^{t\cL^Y}, \e^{t\cL^Y}P^Y\}
\end{equation}
are estimated as \(\underset{t\in[0,T_{\max}]}{\sup} \e^{\mu t} \|P_t\|\), where each \(\mu\in\{\zeta,\lambda^Y,\lambda^Y+\gamma^Y\}\) is computed with a numerical eigensolver (this example could conceivably be solved analytically).
For \(C\in\{C_\lambda^Y,C_\gamma^Y\}\), a discrete timestepping scheme with \(\Delta t=0.01\) is used. For \(C_\zeta\), a finer timestep \(\varepsilon\Delta t\) is used instead to resolve fast transient effects. The value of \(T_{\max}\) is determined dynamically, using the criterion \(T_{\max}=\inf\{t>0:
\, \|P_t\| < \omega_{\mathrm{tol}} \sup_{s<t} \|P_s\|\}\), where we take \(\omega_{\mathrm{tol}}=10^{-12}\). We find this is necessary to avoid divergences associated with loss of numerical precision.

For propagation of the non-linear flow~\eqref{eq:non_linear_flow}, we consider three types of initial conditions: i) drawing \(\eta_0^X\) uniformly from the probability simplex \(\cP(\cX)\) (\textit{random} initialization), ii) setting \(\eta_0^X = \delta_P\) where \(P\sim \cU(\cX)\) (\textit{Dirac} initialization), and iii) \(\eta_0^X = \sum_{z\in \cZ} W_z \pi_z\) where \(W\) is drawn uniformly from the simplex \(\cP(\cZ)\) (\textit{fiber-equilibrated} initialization).
In each case, \(\eta_0^Z\) is determined by the compatibility condition \(\eta_0^Z = \xi_\star \eta_0^X\) of Theorem~\ref{thm:cg_existential_result}.

\begin{figure}
    \centering
    \includegraphics[width=\linewidth]{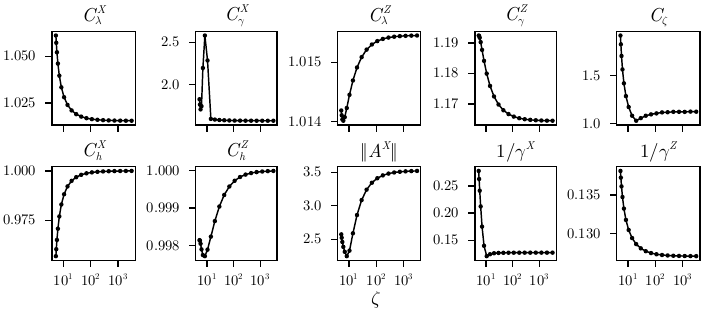}
    \caption{Various terms appearing in the prefactors \(C,C'\) from~\eqref{eq:tv_bound}, as a function of the timescale separation \(\zeta\).}
    \label{fig:cg_theory_prefactor}
\end{figure}

In Figure~\ref{fig:cg_theory_prefactor}, we show one realization of the various components appearing in the expression~\eqref{eq:prefactors_expanded} for the prefactors~\(C,C'\) of~\eqref{eq:tv_bound} against the timescale separation parameter \(\zeta\). All terms converge to a finite value as \(\zeta\to\infty\), showing that the prefactors are to leading order independent of \(\zeta\) in this limit, in accordance with the conclusion of Theorem~\ref{thm:perturbation_theory}. The overall dependence of the constants on~\(\zeta\) is mild for this system.

\begin{figure}
    \centering
    \includegraphics[width=0.8\linewidth]{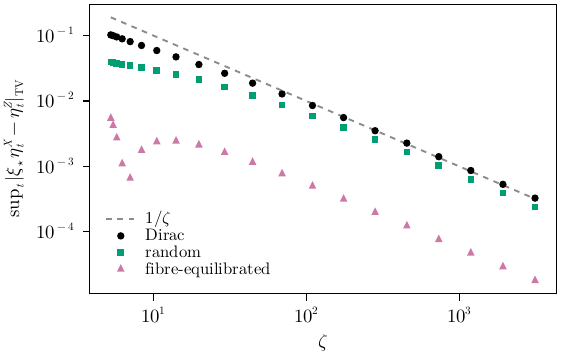}
    \caption{Supremum in time of the total variation bias as a function of the timescale separation \(\zeta\). The theoretical scaling in~\(1/\zeta\) expected from Theorem~\ref{thm:cg_existential_result} is indicated with a dashed line.}
    \label{fig:cg_theory_max_error}
\end{figure}

In Figure~\ref{fig:cg_theory_max_error}, we show one realization of the supremum in \(t\) of the total variation bias \(\underset{t\geq 0}{\sup}\,\d_{\mathrm{TV}}(\xi_\star\eta_t^X,\eta_t^Z)\) against \(\zeta\). The supremum scales as \(1/\zeta\) as predicted by the bound~\eqref{eq:tv_bound}. Generally we observe, uniformly in time, excellent agreement between the coarse-grained canonical description~\(\eta_t^Z\) and the true flow. We also observe that fiber-equilibrated initial conditions lead to lower bias, which is not apparent in our analysis. This will be further investigated for the case of killed diffusion in future work.

\begin{figure}
    \centering
    \includegraphics[width=\linewidth]{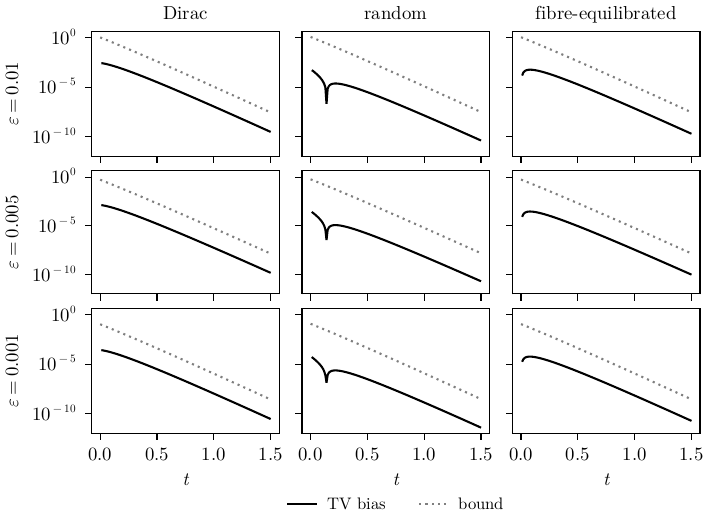}
    \caption{Total variation bias (solid lines) and theoretical bound from Theorem~\ref{thm:cg_existential_result} (dotted lines) for various values of \(\varepsilon\) and choices of initial conditions.}
    \label{fig:cg_theory_bound}
\end{figure}

In Figure~\ref{fig:cg_theory_bound}, we show one realization of the long-time decay of the total variation bias, for various choices of initialization and parameter \(\varepsilon\).
The scaling is consistent with the rate predicted by~\eqref{eq:tv_bound}, while the prefactors \(C\) and \(C'\) from~\eqref{eq:tv_bound} are far from sharp.

\subsection{Implementation of the RNN diagnostic}
\label{sec:num_implementation}

We now investigate the behavior of the RNN on-the-fly diagnostic described in Section~\ref{sec:sbi}, and we compare it to the Gelman--Rubin diagnostic described in Section~\ref{sec:gr}.

In view of applications to MD, we keep in mind that~\(\xi\) is related to a molecular collective variable~\(\overline{\xi}:\Omega\to\R^m\), where~\(\Omega\) is an open and bounded fixed metastable state of the molecular system at hand. In fact we will assume in this work that~\(\overline{\xi}\) is one-dimensional (i.e. \(m=1\)) and bounded (i.e. \(\overline{\xi}(\Omega)\) is bounded), so that without loss of generality we can assume~\(\overline{\xi}(\Omega)=(0,1)\) after a change of gauge and unit. As the framework described in Section~\ref{sec:sbi} does not rely on this assumption, the extension to higher-dimensional CVs is a natural next step, which we will pursue in future work.
The one-dimensional case covers many practical choices of molecular collective variables: bond lengths, dihedral angles, similarity measures, path collective variables, etc. It is also standard practice to define the metastable state in terms of the CV, so that~\(\Omega = {\overline\xi}^{-1}(0,1)\) in our context.

When~\(\overline{\xi}\) is a slow variable of the molecular system, a natural ansatz class for low-dimensional surrogate dynamics is the set of Markovian diffusions over~\(\R^m\). Such processes have been extensively studied in the literature (see for example~\cite{LL10,ZHS16,LLO17,DLPSS18}) as effective dynamics for~\(\overline{\xi}(\overline{X})\), where the molecular dynamics~\(\overline{X}\) is defined as an elliptic diffusion process.
For one-dimensional CVs, these effective dynamics assume the form
\begin{equation}
    \label{eq:effective_diffusions}
    \d \overline{Z}_t = -D(\overline{Z}_t)U'(\overline{Z}_t)\,\d t + \frac1\beta D'(\overline{Z}_t)\,\d t + \sqrt{\frac{2 D(\overline{Z}_t)}{\beta}}\,\d B_t,
\end{equation}
where the coefficients~\(U:[0,1]\to \R\) and~\(D:[0,1]\to (0,\infty)\) are assumed to be smooth, and correspond physically to free energy and diffusivity profiles respectively. The parameter~\(\beta = 1/kT >0\) is proportional to the inverse temperature, and~\(B\) is a standard Brownian motion on~\(\R\).

In this work, our aim is to use a surrogate process for the decorrelation of~\(\overline{X}\) within~\(\Omega\). We take as an ansatz the class of probability flows associated to dynamics of the form~\eqref{eq:effective_diffusions}, killed at the boundary~\(\{0,1\}\), and conditioned on non-absorption. Indeed, we expect this class to contain the canonical diffusion associated with the molecular dynamics within~$\Omega$, that is the analog of the canonical model~\(Z\) from Definition~\ref{def:perfect_model}.
Verifying this rigorously requires extending the coarse-graining results of Section~\ref{sec:theory} to killed diffusions, but this ansatz will nevertheless motivate our construction of the synthetic prior~\(\rho_{\rm synth}\).

\subsubsection{Construction of the prior ensemble}

As the framework described in Section~\ref{sec:sbi} only handles discrete surrogate models, we naturally define \(\cZ \subset (0,1)\) on a finite grid, and take~\(\xi:\cX\to\cZ\) to be an associated quantization of~\(\overline{\xi}\).
Correspondingly, the prior ensemble~\(\rho_{\rm dyn}\) is supported on generators of discrete dynamics, which are jump-process discretizations of the continuous dynamics~\eqref{eq:effective_diffusions}. More precisely,~\(\rho_{\rm dyn}\) is supported on the set of generators of nearest-neighbor jump-processes, on a regular grid~\(\cZ\subset(0,1)\), killed at the boundary~\(\partial^{\cZ}=\{0,1\}\). The size of the grid~\(|\cZ|=d_{\rm grid}=100\) will be fixed throughout.

The relationship between~\(Z\) and~\(\overline{Z}\) is the following: the jump rates of~\(\cL^Z\) are given, for all~\(1\leq k\leq d_{\rm grid}\), by
\begin{equation}
    \label{eq:jump_rates}
    \begin{aligned}
         & \cL^Z_{k,k\pm 1} = \sigma_{d_{\rm grid},\beta}\left(\frac{D(z_{k\pm 1})+D(z_{k})}{2}\right)\e^{-\frac{\beta}{2}(U(z_{k\pm 1})-U(z_{k}))}, \\
         & \cL^Z_{k,k} = -\left(\cL^Z_{k,k-1}+\cL^Z_{k,k+1}\right),
    \end{aligned}
\end{equation}
and zero otherwise, where the energy~\(U\) and diffusion~\(D\) are evaluated on the grid~\(\cZ = \{z_k\}_{k=1,\dots,d_{\rm grid}}\), as well as~\(z_0:=0\) and~\(z_{d_{\rm grid}+1}:=1\). The constant~\(\sigma_{d_{\rm grid},\beta} = (d_{\rm grid}+1)^2/\beta\) is needed to get a consistent approximation of the generator~\(\cL^{\overline{Z}}\) of the continuous dynamics~\eqref{eq:effective_diffusions} in the limit~\(d_{\rm grid}\to\infty\).

The jump process~\(Z\) would be mass-conserving were it not for the contributions~\(-\cL^Z_{1,0}\) and~\(-\cL^Z_{d_{\rm grid},d_{\rm grid}+1}\) to the diagonal coefficients~\(\cL^Z_{1,1}\) and~\(\cL^Z_{d_{\rm grid},d_{\rm grid}}\) respectively, which make~\(\cL^Z\) into the generator of a conditionally irreducible jump process on~\(\cZ\).

Given these definitions, the synthetic prior~\(\rho_{\rm dyn}\) is specified by a sampling procedure for the random physical parameters~\(U,D,\beta\). The prior distributions of~\(U,D\) and~\(\beta\) are described in detail in Appendix~\ref{sec:synth_prior_specification}.
In Figure~\ref{fig:synth_dynamics}, we represent sixteen independent draws of~\(\cL^Z\) from~\(\rho_{\rm dyn}\), by plotting the corresponding physical parameters. The prior covers a broad range of metastable states, including ones containing multiple free energy wells at low temperature.

At low temperatures~($\beta\gg 1$), there exist initial configurations~\(Z_0\) and landscapes~$(U,D)$ for which decorrelation inside~\(\cZ\) is itself a metastable process. We therefore expect the distribution of~\(t_0(\cL^Z,Z_0)\) under~\(\rho_{\rm synth}\) to be broad.
This is what we verify in Figure~\ref{fig:synth_tcorr}, where we plot the density per decade for this distribution, which is defined as
\begin{equation}
    \label{eq:dpd}
    t\mapsto \frac{\d}{\d(\log_{10}\,t)}\P_{\rho_{\rm synth}}{t_0(\cL^Z,Z_0)<t},
\end{equation}
where we recall~\(\rho_{\rm synth}\) is the total prior over~\((\cL^Z,Z_0)\). This function is estimated empirically by the black curve in Figure~\ref{fig:synth_tcorr}. The other lines correspond to density per decade curves of decorrelation times for various diagnostics. As expected, we observe decorrelation times span several orders of magnitude as~\((\cL^Z,Z_0)\) ranges over the ensemble. The procedure to approximate~\(t_0(\cL^Z,Z_0)\) is described in the next subsection.

\begin{figure}
    \centering
    \figorbox{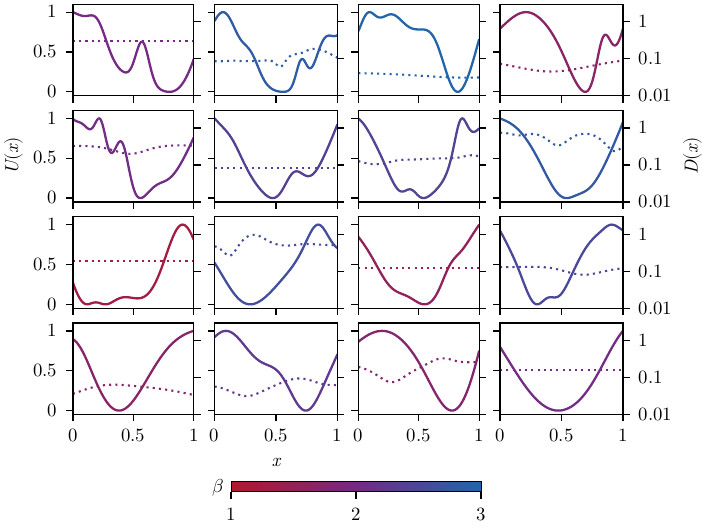}
    \caption{Free energy profiles~\(U\) (solid lines) and diffusion profiles~\(D\) (dotted lines, in logarithmic scale) for sixteen independent samples from the training prior. The value of the inverse temperature~\(\beta\) is indicated with a color gradient.}
    \label{fig:synth_dynamics}
\end{figure}

\begin{figure}
    \centering
    \figorbox{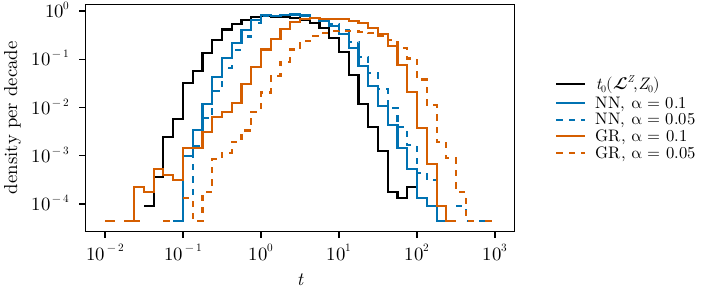}
    \caption{Density per decade~\eqref{eq:dpd} of the honest decorrelation times~$t_0(\cL^Z,Z_0)$ (black), of the RNN decorrelation times (blue) and of the Gelman--Rubin decorrelation times (orange), for decision thresholds~\(\alpha=0.1\) (solid lines) and~\(\alpha=0.05\) (dashed lines).}
    \label{fig:synth_tcorr}
\end{figure}

\subsubsection{Synthetic data generation}
For each sample from the synthetic prior~\((\cL^Z,Z_0)\sim\rho_{\rm synth}\), we approximate the honest decorrelation time~\(t_0(\cL^Z,Z_0)\) as follows. We represent the operator~\(\cL^Z\) defined in~\eqref{eq:jump_rates} as a sparse matrix. This allows~\(t_0(\cL^Z,Z_0)\) to be computed very quickly by~i) solving the sparse eigenproblem defining~\((\lambda^Z,\nu^Z)\) using the Julia interface to Arpack and ii) directly computing the non-linear flow~\eqref{eq:non_linear_flow} in discrete steps of a fixed lag time~\(\tau_{\rm gt}=0.01\), by iteratively applying the adjoint of~\(Q_{k\tau_{\rm gt}}^Z = \e^{k\tau_{\rm gt}\cL^Z}\) to~\(\eta_0^Z\in\cP(\cZ)\), and normalizing the probability flow.
This makes it possible to compute, for~\(\eta_0^Z = \delta_{Z_0}\),
\begin{equation}
    \label{eq:synth_tcorr_ref}
    t_0(\cL^Z,Z_0) \approx \tau_{\rm gt}\min\left\{k\geq 1:\,\d_{\rm TV}\left(\eta_{k\tau_{\rm gt}}^Z,\nu^Z\right) < \varepsilon_{\rm corr}\right\},
\end{equation}
which amounts to counting the number of iterations to convergence of a power iteration algorithm. We fix~\(\varepsilon_{\rm corr}=0.05\) throughout our experiments.

For higher-sdimensional grids~(\(|\cZ| \propto d_{\rm grid}^m\)), which are needed to represent the dynamics of higher-dimensional collective variables, this approach remains tractable if the operations above are implemented with matrix-free solvers. In particular, the propagation of the semigroup could be implemented using exponential-Krylov methods~\cite{TS14}. For our value of~$d_{\rm grid}$, we found a dense representation of~\(\e^{\tau_{\rm gt}\cL^Z}\) to be more efficient.
The computation of~$t_0(\cL^Z,Z_0)$ is affordable enough that it can be performed online during training and evaluation. For this reason, prior samples are never used in more than one training batch (and never thereafter in testing or validation). This specificity of the one-dimensional case likely explains why our model successfully learns in the absence of any explicit regularization strategy.

To compute features, given a choice of number of replicas~\(N\) and lag time~$\tau$, we sample histogram trajectories
\begin{equation}
    \label{eq:feature_traces}
    H_k^{Z,\xi} =  \mathrm{Hist}\left(\frac1{\tau}\int_{(k-1)\tau}^{k\tau} \widehat{\eta}_{t,N}^{\overline{Z}^{\Delta t}}\,\d t\right),
\end{equation}
where~\(\overline{Z}^{\Delta t}\) is an Euler--Maruyama discretization of the continuous diffusion~\eqref{eq:effective_diffusions} corresponding to~\(Z\), and~\(\widehat{\eta}_{t,N}^{\overline{Z}^{\Delta t}}\) is the empirical measure~\eqref{eq:fv_empirical_measure} of an associated Fleming--Viot process (see Definition~\ref{def:fleming_viot}).

This introduces a slight discrepancy with respect to the procedure described in Section~\ref{sec:sbi}, as the Fleming--Viot processes in Section~\ref{sec:sbi} are generated by the discrete dynamics~\(Z\).
The substitution~\(Z\leftarrow \overline{Z}^{\Delta t}\) introduces a new discretization bias (which combines with the finite-\(N\) sampling error motivating Assumption~\tagref{B2}). We expect this discretization error to vanish in the limit~\(d_{\rm grid}\to\infty,\,\Delta t\to 0\), and to be partially filtered out by the histogram featurization.
Controlling this discretization bias rigorously requires new results in the numerical analysis of discretized Fleming--Viot diffusions, which as far as we are aware are not yet available in the literature, and lie outside the scope of this work.

This is also consistent with deployment conditions for the RNN diagnostic, as these discretized Fleming--Viot trajectories are generally those available in a simulation setting.

The length of simulated Fleming--Viot trajectories is randomized in training, according to the rule~\eqref{eq:horizon_sampling_rule}, with the added constraint that~$L\geq 5$, i.e.~$L=\max(5,\lceil Rt_0(\cL^Z,Z_0)/\tau\rceil),\,R\sim\cU(1,2)$. This rule still satisfies the conditions of Lemma~\ref{lemma:randomized_loss_minimizer}. In accordance with the objective~\eqref{eq:ml_problem}, we also randomize~\((N,\tau)\) in training, where~\(N,\tau/\Delta t\sim\,\cU\{10,\dots,200\}\) independently, fixing~\(\Delta t=0.001\) throughout.

To amortize the computation of~\(\nu^Z\) and~\(\e^{\tau_{\rm gt}\cL^Z}\), samples of physical parameters~\((U,D,\beta)\) are used to produce more than one Fleming--Viot simulation. For each~\((U,D,\beta)\) a number of samples of~\((Z_0,N,\tau)\) are drawn, the QSD and semigroup are computed once, and each decorrelation time~\(t_0(\cL^Z,Z_0)\) is computed according to~\eqref{eq:synth_tcorr_ref}. One Fleming--Viot simulation is then performed for each~$Z_0$, and the feature sequences are finally truncated using several samples of the time-horizon~\(L\).
The values we used produced training batches of
\begin{equation}
    \label{eq:training_batch_size}
    \underbrace{5}_{(U,D,\beta)}\, \times\,\underbrace{5}_{(Z_0,N,\tau)}\times \underbrace{2}_{L} = 50
\end{equation}
feature sequences. Note that the batch elements are not \emph{i.i.d.} variables, however their use in the estimator~\eqref{eq:stochastic_gradient_estimator} is still consistent in the limit of many independent samples of~\((U,D,\beta)\).

\subsubsection{Architecture and training details}
We implement the RNN model described in Section~\ref{sec:sbi} using the Julia machine learning library Flux~\cite{flux}.
All parameters were updated using the Adam optimizer with parameters~({\texttt{beta1},\texttt{beta2},\texttt{epsilon}})\(=(0.9,0.999,10^{-8})\).

Training proceeds in two steps. In a pretraining step, the architecture of the model is tuned using a Bayesian optimization algorithm implemented with the NePS~\cite{neps} library.
This hyperparameter search uses a precomputed dataset of~\(1000\) feature sequences, with~\(800\) sequences used for training and~\(200\) sequences used to compute the objective, which consists in the empirical loss for~\eqref{eq:ml_problem} over the held-out set.
The search was specified over the following hyperparameters: input feature dimension~$d_{\rm hist}$, number of channels and depth (number of layers) in the CNN, dimension~$d_{\rm rnn}$ and depth of the RNN, width and depth of the classifier, and learning rate in the optimizer.
The same data was used for different candidate architectures, producing a final model with~\(p=195313\) trainable parameters, most of which are in the recurrent layer. The architecture is described in detail in Appendix~\ref{sec:nn_architecture} (see Table~\ref{tab:architecture}).

In a second step, the resulting model's weights were retrained using the Adam optimizer (with a reinitialized state), using~\(3.75\times 10^5\) feature sequences from~$3.75\times10^4$~physical landscapes~$(U,D,\beta)$, generated during training and assembled into batches according to~\eqref{eq:training_batch_size}.
A cosine learning-rate schedule ($\eta_k = \eta_{\rm min} + \frac12(\eta_{\rm max}-\eta_{\rm min})(1+\cos(k\pi/K))$ with \(\eta_{\rm max}=5\times10^{-4}, \eta_{\rm min}=5\times10^{-5}\)) was applied, where each~$k$ corresponds to an ``epoch'' of~$50$ training batches, for a total of~$K=150$ epochs.
Training was performed on a laptop computer, using~$5$ CPU cores for approximately~$11$ hours.

\subsection{Performance on synthetic data}
\label{sec:num_synth}

We first validate the RNN diagnostic~\eqref{eq:rnn_decorrelation_time} on synthetic data, comparing it to the Gelman--Rubin~\eqref{eq:gr_decorrelation_time} diagnostic.
In this section, samples of~\((\cL^Z,Z_0,N,\tau)\) are drawn from the training distribution, and the corresponding Fleming--Viot simulations are performed on a time-horizon of~\(10t_0(\cL^Z,Z_0)\).
A diagnostic which predicts a decorrelation time larger than this time horizon is declared to ``fail'', as a ten-fold overestimation of the decorrelation time~\eqref{eq:time_true} roughly corresponds to a ten-fold decrease in the parallel efficiency of the parallel replica algorithm (see~\cite[Section 2.6]{PUV15} or~\cite[Appendix B]{BLS26}).

\subsubsection{Binary classification performance}
We first investigate the performance of each diagnostic as a decorrelation classifier in the training distribution. We sample uniformly at random a final observation frame~\(K\) such that~\(\tau\leq K\tau \leq 2t_0(\cL^Z,Z_0)\), and compute the corresponding diagnostic statistics, namely \(h_{K,\theta}(H_1,\dots,H_K)\) for the RNN diagnostic, and~\(R_{\tau K,\xi(z)=z}^Z\) for the Gelman--Rubin diagnostic (see Equation~\eqref{eq:gelman_rubin}).
In this experiment, we will also compare this to the instantaneous total-variation distances
\begin{equation}
    \label{eq:tv_oracle}
    O_k^Z = \d_{\rm TV}\left(H_k,{\rm Hist}\left(\nu^Z\right)\right),\qquad 1\leq k\leq K
\end{equation}
between each Fleming--Viot histogram~\(H_k\) and the histogram featurization of the QSD (using~$d_{\rm hist}=64$ bins). The associated diagnostic is given by
\begin{equation}
    \label{eq:tcorr_tv}
    t_{\rm corr}^{\rm TV}(\alpha) = \tau\times\inf\,\left\{k\geq 1:\,\,O_k^Z < \alpha\right\}.
\end{equation}
This diagnostic is non-recurrent, but directly informed of~\(\nu^Z\), by contrast to the RNN and Gelman--Rubin diagnostics for which the opposite statement holds.

The Gelman--Rubin statistic~\eqref{eq:gelman_rubin} is computed at every multiple of the lag time~\(\tau\), and the reported Gelman--Rubin decorrelation time is therefore
\begin{equation}
    \label{eq:gr_num}
    t_{\rm corr}^{\rm GR}(\alpha) = \tau\times \inf\left\{k\geq 1:\,\, R_{k\tau,\xi}^Z < 1+\alpha\right\}
\end{equation}
rather than~\eqref{eq:gr_decorrelation_time}. Therefore~\(\tau\) acts as a burn-in time and a decision period (which both have to be defined in practical applications of Gelman--Rubin~\cite{BLS15,HL19}).

Each of these diagnostics can be equivalently expressed in terms of an observable~${f:E\to [0,\infty)}$ over a set of sequences~$E$, where~$f\to 0$ as the diagnostic confidently predicts decorrelation.
Given a partition of a sequence dataset~$\cD = \cD_0\cup\cD_1$ into negative and positive examples (corresponding to transient and quasi-stationary endpoints, respectively), the receiver operating characteristic curve~(ROC curve, see for example~\cite[Section 5.7.2.1]{M12}) is a classical measure of classifier performance,
which we estimate as a parametric curve~$\alpha\mapsto ({\rm FPR}(\alpha),{\rm TPR}(\alpha))$, where FPR and TPR are empirical false positive and true positive rates
\begin{equation}
    \label{eq:roc}
    {\rm FPR}(\alpha) = \frac1{|\cD_0|}\sum_{s\in \cD_0}\1_{f(s)<\alpha},\quad {\rm TPR}(\alpha) = \frac1{|\cD_1|}\sum_{s\in \cD_1}\1_{f(s)<\alpha}.
\end{equation}
The area under the ROC curve (AUC) corresponds to the probability that a random pair of negative and positive examples~$(s_0,s_1)\in \cD_0\times\cD_1$ is correctly ranked by the classifier statistic, i.e.~$f(s_0)>f(s_1)$.
We estimate it from the Wilcoxon--Mann--Whitney U statistic (see~\cite{HM82,CJ07})
\begin{equation}
    \label{eq:auc}
    {\rm AUC} = \frac{1}{|\cD_0||\cD_1|} \sum_{s_0\in \cD_0}\sum_{s_1\in \cD_1} \left(\1_{f(s_0)>f(s_1)} + \frac12 \1_{f(s_0)=f(s_1)}\right).
\end{equation}
The AUC score is a popular metric for assessing the overall discriminative quality of a binary classifier, as it is independent of the choice of decision threshold~$\alpha$.

The experiment was performed on a \(6\times 6\) grid of design parameters,~\(N,\tau/\Delta t\in\{10,25,50,100,150,200\}\).
For each choice of design parameter,~\(200\) validation batches were assembled, consisting of~\(5\) draws of physical parameters~\((U,D,\beta)\) times \(5\) draws of initial points~\(Z_0\), each giving one feature sequence (training batches, by contrast, assembled two feature sequences for each~\(Z_0\)).

\begin{figure}
    \centering
    \figorbox[width=0.55\linewidth]{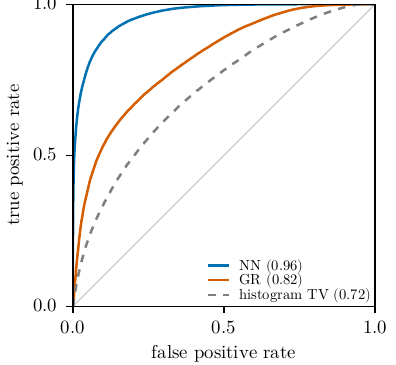}
    \caption{ROC curves for the RNN, Gelman--Rubin and histogram TV~\eqref{eq:tv_oracle} diagnostics, within the training prior. Samples are aggregated over a grid of design parameters~\((N,\tau)\). The ROC-AUC score of each diagnostic is indicated in the legend.}
    \label{fig:synth_roc}
\end{figure}

In Figure~\ref{fig:synth_roc}, we show the ROC curves for the three considered diagnostics, aggregating samples over all choices of design parameters~\((N,\tau)\).
The RNN curve dominates the Gelman--Rubin curve, which itself dominates the total variation curve associated with~\eqref{eq:tv_oracle}, indicating a strict ordering in discriminative quality. This ordering is reflected in the computed AUC scores (\(0.96,0.82\) and \(0.72\) respectively).
The poor performance of the total variation diagnostic reveals the usefulness of a recurrent diagnostic such as the RNN or Gelman--Rubin, since access to a reference QSD and~$H_k$ alone are insufficient to accurately predict decorrelation over a broad range of systems, due to finite-particle noise, even in toy one-dimensional systems.

\begin{figure}
    \centering
    \figorbox[width=1\linewidth]{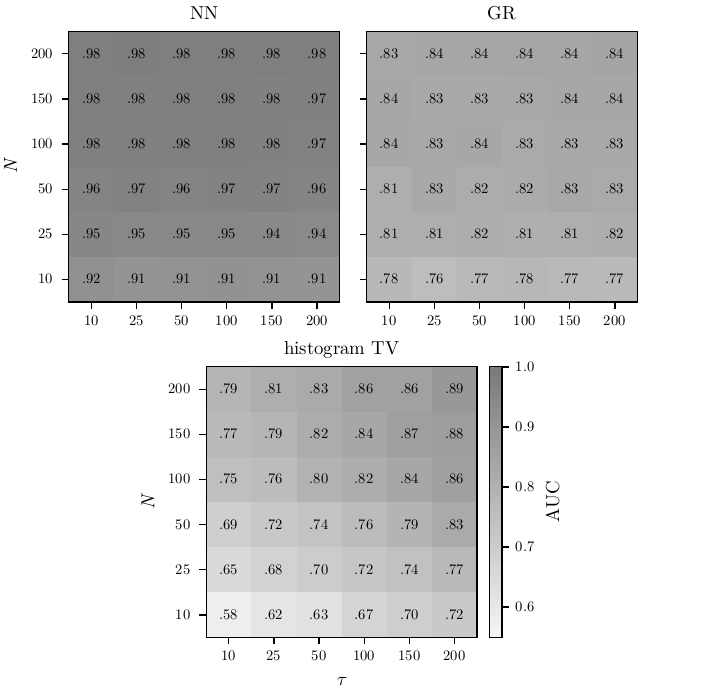}
    \caption{ROC-AUC scores of the RNN, Gelman--Rubin and histogram TV~\eqref{eq:tv_oracle} diagnostics for the sub-classification tasks associated to each design setting~\((N,\tau)\), under the training prior.}
    \label{fig:synth_auc_grid}
\end{figure}

In Figure~\ref{fig:synth_auc_grid}, we report estimated AUC scores for each choice of~\((N,\tau)\) in the parameter grid. We observe that the RNN diagnostic uniformly outscores the Gelman--Rubin diagnostic, in the sense that the worst RNN AUC score is higher than the best Gelman--Rubin AUC score within the grid.
We also observe that both the RNN and Gelman--Rubin diagnostics are relatively robust to the choice of~\(\tau\) on this task (recall that~\(\tau\) plays the role of a burn-in time for the Gelman--Rubin diagnostic), and less so to the choice of~\(N\), with performance improving monotonically for higher numbers of replicas, suggesting that the finite-\(N\) Fleming--Viot particle approximation error plays a non-negligible role.
This is further confirmed by the total variation AUC scores, which show a high sensitivity to both the choice of~\(\tau\) and~\(N\): for low values of~\((N,\tau)\), the total-variation diagnostic is barely more discriminative than a random classifier, with AUC scores increasing monotonically as each frame~\(H_k\) accumulates more samples.
For the higher values of~\((N,\tau)\) we consider, the total-variation diagnostic eventually outscores the Gelman--Rubin diagnostic. The sensitivity of the total variation diagnostic to the number of aggregated samples in each frame can be explained by the fact that the statistic~\eqref{eq:tv_oracle} has no mechanism by which to filter out the finite-$N$ error.
We consider the total variation diagnostic to be impractical on the basis of these results, and therefore exclude it from further benchmarks.

\subsubsection{On-the-fly quasi-stationarity detection}
We next study the behavior of the Gelman--Rubin and RNN diagnostics as on-the-fly predictors of the decorrelation time, for fixed choices of the decision threshold~\(\alpha\). For both diagnostics, we compare a lax decision threshold~\(\alpha=0.1\) and the more conservative threshold~\(\alpha=0.05\). For the Gelman--Rubin diagnostic, these values are those prescribed in~\cite{BLS15,HL19}, and for the RNN, these are motivated by the average posterior risk estimates of Lemmas~\ref{lemma:randomized_loss_minimizer} and~\ref{lemma:risk_control}.
The Fleming--Viot simulations are the same as those used in the previous subsection (corresponding in particular to many values of~$(N,\tau)$), which were simulated up to~\(10t_0(\cL^Z,Z_0)\) to allow diagnostics to declare very late decorrelation times if necessary. If a diagnostic does not declare decorrelation within this time-frame, it is deemed to fail on this trajectory. 

\begin{figure}
    \centering
    \figorbox{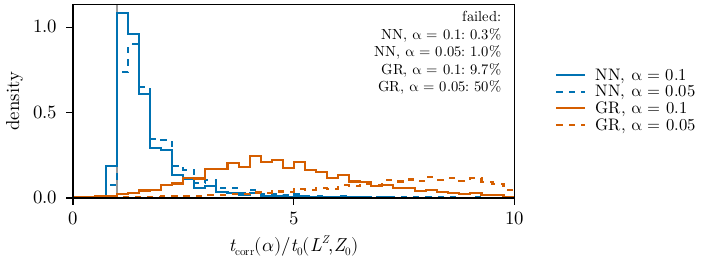}
    \caption{Distributions of the declared decorrelation times~\(t_{\rm corr}(\alpha)\) in the training prior, in units of~\(t_0(\cL^Z,Z_0)\), for the RNN (blue) and Gelman--Rubin (orange) diagnostics, for~\(\alpha=0.1\) (solid lines) and~\(\alpha=0.05\) (dashed lines). The proportion of failed trajectories for each diagnostic is indicated in the legend.}
    \label{fig:synth_detection}
\end{figure}

In Figure~\ref{fig:synth_detection}, we show empirical distributions of the diagnostic decorrelation times~\eqref{eq:rnn_decorrelation_time} and~\eqref{eq:gr_num}, in units of the honest decorrelation time~\(t_0(\cL^Z,Z_0)\). The area under each curve is given by the proportion of Fleming--Viot trajectories for which the diagnostic did not fail.
The distribution of~\(t^{\rm RNN}_{\rm corr}(\alpha)\) has a sharp mode around~\(t_{0}(\cL^Z,Z_0)\) (for both choices of~\(\alpha\)), whereas \(t^{\rm GR}_{\rm corr}(\alpha)/t_{0}\) has a much wider distribution with a mode around~\(4\) at the lax setting~\(\alpha=0.1\), and~\(8\) at the conservative setting~\(\alpha=0.05\).
The Gelman--Rubin diagnostic fails on a substantial proportion of trajectories (\(9.7\%\) for \(\alpha=0.1\) and \(50\%\) at \(\alpha=0.05\)), which biases the empirical distribution towards earlier Gelman--Rubin decorrelation times. By contrast, only~\(0.3\) and~\(1\%\) of trajectories cause the RNN diagnostic to fail for the lax and conservative thresholds, respectively.
We note that the Gelman--Rubin diagnostic is much more sensitive to the choice of decision threshold, as indicated by the distribution of decorrelation times shifting by an approximate factor of two when \(\alpha\) is halved (\(0.1\to 0.05\)). This behavior is consistent with an algebraic time-convergence of the variance ratio in~\eqref{eq:gelman_rubin} at a rate~\(1/t\).

\subsubsection{Posterior risk calibration and prior misspecifications}
We next study the probabilistic interpretability of the decision threshold~\(\alpha\) in the RNN diagnostic, as introduced in Lemmas~\ref{lemma:randomized_loss_minimizer} and~\ref{lemma:risk_control}.
We investigate the calibration of our learned approximation~\(h_\theta\) to the posterior-averaged risk of false decorrelation.

\begin{figure}
    \centering
    \figorbox[width=0.5\linewidth]{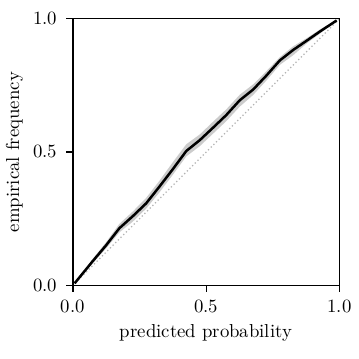}
    \caption{Empirical decorrelation frequency as a function of empirical RNN predictions under the training prior. Perfect calibration is indicated with a dotted line. Confidence bands are computed with a bootstrap estimator, treating different Fleming--Viot trajectories on the same landscape~$(U,D,\beta)$ as independent samples.}
    \label{fig:synth_calibration}
\end{figure}

A total of~\(600\) batches were drawn from the training distribution, representing \(15000\) Fleming--Viot trajectories, each simulation lasting for~\(2t_0(\cL^Z,Z_0)\). At every frame, the output of the RNN~\(h_\theta\) is recorded, giving~\(1.24\times 10^6\) predictions in total. In Figure~\ref{fig:synth_calibration}, we plot the model's predicted probabilities against the empirical proportion of decorrelated Fleming--Viot trajectories, according to the reference definition~\eqref{eq:synth_tcorr_ref}. Predictions are grouped into~\(20\) bins of equal width (the smallest bin containing~\(16262\) predictions).
We observe that the model underestimates decorrelation probabilities, on average over the prior (the curve lies above the diagonal), as expected from Lemma~\ref{lemma:randomized_loss_minimizer}. Generally, this under-confidence is mild, and the model appears well-calibrated, particularly for high predicted probabilities (corresponding to low values of~\(\alpha\)).
Similarly, and consistently with Lemma~\ref{lemma:risk_control}, the risk of false decorrelation appears well-calibrated on average over the training prior, as the empirical proportion of falsely-decorrelated trajectories is below~$\alpha$~(as indicated by the~``early'' columns in the first row of Table~\ref{tab:robustness_nn}).

We next study the behavior of these empirical risks on purposefully misspecified priors, obtained by shifting the distributions of physical parameters~\((U,D,\beta)\) away from the training distribution. We experiment with the following transformations.
\begin{itemize}
    \item[\(\circ\)] \textbf{high \(D\)}: diffusion coefficients are larger than those observed in training.
    \item[\(\circ\)] \textbf{low \(T\)}: temperatures are lower (higher~\(\beta\) parameters) than those observed in training.
    \item[\(\circ\)] \textbf{low \(T\), high \(D\)}: a combination of the two previous transformations.
    \item[\(\circ\)] \textbf{low \(T\)/one-well}: temperatures are lower, but \(U\) has only shallow energy barriers within the state, so decorrelation is not a metastable process.
\end{itemize}
Precise definitions for these shifts are provided in Appendix~\ref{sec:synth_prior_specification}.

In Figure~\ref{fig:ood_calibration}, we repeat the calibration study on each of these shifted priors. The least populated bins contained respectively~\(1899\),~\(40637\),~\(3824\) and~\(27693\) samples for each of the transformations listed above.
Beyond what could be expected from our methodology, the model remains reliable on average in terms of false-decorrelation risk (see the~``early'' columns in Table~\ref{tab:robustness_nn}), and generally conservative on average, as the predicted risk typically overestimates the empirical risk.
A mild miscalibration occurs at low temperatures (the ``low~$T$'' ensemble). We can speculate that this is due to overconfident predictions in cases where there are metastable substates, within which the particle approximation becomes quasi-stationary, and is mistakenly labeled as stationary.
Any nonconvergence diagnostic may fail this way. If such metastable substates are excluded from the ensemble~(low~\(T\), one well), this overconfidence appears to be largely corrected. 

In the first lines of Table~\ref{tab:robustness}, we collect some additional summary statistics associated with Fleming--Viot trajectories sampled under these prior shifts, reporting, for both the RNN and Gelman--Rubin diagnostics: AUC scores (for samples aggregated over time and parameters~\(N,\tau\)), median normalized decorrelation time~\(t_{\rm corr}(\alpha)/t_0(\cL^Z,Z_0)\) and~$25\%,75\%$ quantiles, empirical false decorrelation risks, and empirical failure risks for each of the decision thresholds~\(\alpha\in\{0.1,0.05\}\).
The RNN diagnostic outscores the Gelman--Rubin in AUC under all prior shifts. We observe, in line with Figure~\ref{fig:synth_detection}, that the Gelman--Rubin diagnostic generally overestimates the reference decorrelation time, with a larger interquartile range than for the RNN. The Gelman--Rubin diagnostic fails on a significant number of trajectories, sometimes on a majority of them.
On the other hand, the Gelman--Rubin diagnostic has a lower empirical risk of false decorrelation compared to the RNN diagnostic, in line with a general conservativeness at these choices of~$\alpha$. Beyond what Lemma~\ref{lemma:risk_control} allows, the empirical risk of false decorrelation is controlled by~\(\alpha\) for the RNN diagnostic, over all the synthetic prior misspecifications we considered.

\begin{figure}
    \centering
    \figorbox{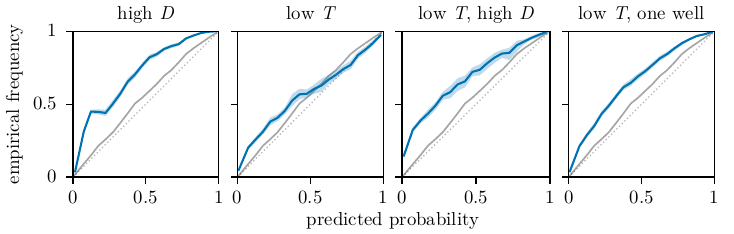}
    \caption{Empirical frequency of decorrelation as a function of the probability predicted by the RNN diagnostic, under each prior shift (blue lines) and under the training prior (grey lines). Perfect calibration is indicated with a dotted line.}
    \label{fig:ood_calibration}
\end{figure}

\subsection{Transfer to a solvated dimer}
\label{sec:num_dimer}

We now compare the behavior of the RNN diagnostic to the Gelman--Rubin diagnostic on a solvated dimer, a toy model of biochemical systems.
We describe the system in Section~\ref{sec:dimer_system}, the dynamics and reference data in Sections~\ref{sec:dimer_md} and~\ref{sec:dimer_ref}, and the results of our three experiments in Sections~\ref{sec:dimer_otf},~\ref{sec:dimer_bias} and~\ref{sec:dimer_mitigation}.

\subsubsection{Dimer system and collective variables}
\label{sec:dimer_system}
The system is composed of \(N_{\rm at}=512\) atoms placed in a cubic periodic domain~\(\cQ = L\bbT^{3N_{\rm at}}\) ($\bbT=\R/\bbZ$ is the one-dimensional torus), with~$L=(N_{\rm at}/\rho)^{1/3}$ enforcing a particle density of~\(\rho=0.693\), and is simulated at an inverse temperature~\(\beta=1/(k_{\rm B} T)=1\). All quantities in this experiment are given in reduced Lennard--Jones units, see~\cite[Appendix B]{AT17}.
To model the liquid phase typical of biochemistry, the solvent particles interact via the WCA pair potential
\begin{equation}
    \label{eq:wca}
    v_{\rm{WCA}}(r) =\left\{
    \begin{aligned}
         & 4\epsilon\left[\left(\frac{r}{\sigma}\right)^{-12} - \left(\frac{r}{\sigma}\right)^{-6}\right] & \text{if }0<r<2^{1/6}\sigma,   \\
         & -\epsilon                                                                                      & \text{if }r\geq 2^{1/6}\sigma.
    \end{aligned}
    \right.
\end{equation}
This potential is simply the repulsive component of the classical Lennard--Jones potential, where we take~\(\epsilon=\sigma=1.0\).

A selected pair of atoms (say atoms~\(1\) and~\(2\)) acts as a solvated dimer. Between these atoms, the WCA interaction is turned off, and replaced by a potential of their interatomic distance. This potential is defined for~\(r\in(0.68,2.22)\) by
\begin{equation}
    \label{eq:dimer_potential}
    W(r) = K_W\left(\frac{s(r)^6}{6} - (1+C_W^2)\frac{s(r)^4}{4} + C_W^2\frac{s(r)^2}{2}\right) + E_W s(r),
\end{equation}
where~\(s(r) = (r-1.45)/0.7\) and~\((K_W,C_W,E_W)=(78,0.6,0.49)\), and is extended linearly beyond this range.

The total interaction potential of the system is therefore
\begin{equation}
    \label{eq:dimer_total_energy}
    V(q)=\sum_{\underset{(i,j)\neq (1,2)}{1\leq i<j\leq N_{\rm at}}}v_{\rm{WCA}}(r_{ij}) + W(r_{12}),
\end{equation}
where~\(q=(q_1,\dots,q_{N_{\rm at}})\) is the vector of atomic positions and~\(r_{ij}\) denotes the periodic distance between atoms~\(i\) and~\(j\), which is their geodesic distance in~$L\bbT^3$.
The system's equilibrium distribution is the canonical Gibbs measure
\begin{equation}
    \label{eq:gibbs_measure}
    \mu(\d q) = \frac{\e^{-\beta V(q)}}{\dsint_{\cQ}\e^{-\beta V}}\,\d q \in \cP(\cQ).
\end{equation}

The potential~\(W\) induces three metastable states in the system, corresponding to three conformations of the dimer molecule: a compact conformation, an intermediate conformation, and a stretched conformation.
In this experiment, we consider decorrelations of the system within the intermediate state, defined by
\begin{equation}
    \label{eq:intermediate_state}
    \Omega=\left\{r_{\rm L}<r_{12}<r_{\rm R}\right\},\qquad (r_{\rm L},r_{\rm R}) = (1.0396,1.8795).
\end{equation}
To monitor the system, we consider two collective variables:
\begin{equation}
    \label{eq:cvs}
    \overline{\xi}(q) = \frac{r_{12}-r_{\rm L}}{r_{\rm R}-r_{\rm L}},\qquad n_{\rm coord}(q) = \sum_{j=3}^{N_{\rm at}} \scS(\min(r_{1j},r_{2j})),\qquad {\scS}(r)=\frac{1}{1+e^{\frac{r-r_c}{w}}},
\end{equation}
where~$r_c=2^{1/6}\sigma$ is the interaction radius of the WCA potential and~$w=0.2$.
The variable~\(\overline{\xi}\) is simply the dimer length, normalized so that~\(\Omega = \overline{\xi}^{-1}(0,1)\). The variable~\(n_{\rm coord}\) is a soft solvent coordination number.
Atoms at a distance of at least~$1,1.5,2 r_c$ from the dimer contribute respectively at most~$0.5,0.06$ and~$0.004$ to the sum in~\eqref{eq:cvs}, so that~$n_{\rm coord}$ roughly counts the number of atoms in the first solvation shell (the value~\(1.5 r_c\) being a standard shell radius for Lennard--Jones fluids, see~\cite[Figure 2.2]{AT17}).

\begin{figure}
    \centering
    \figorbox{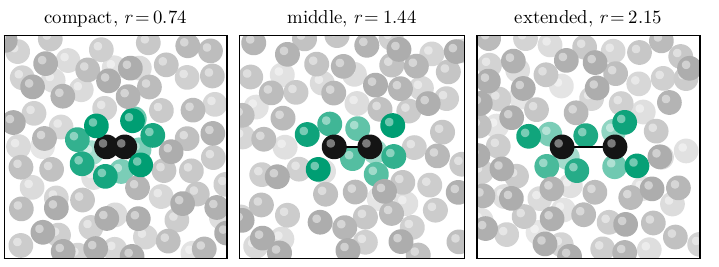}
    \caption{Configurations of the solvated dimer in each of its three metastable conformations. The dimer atoms are shown in black, and the solvent atoms within~$1.5r_c$, contributing the bulk of~\(n_{\rm coord}\) in green.}
    \label{fig:dimer_snapshots}
\end{figure}

In Figure~\ref{fig:dimer_snapshots}, we represent three sample configurations of the solvated dimer system, within each one of its metastable states. The dimer is represented as a pair of bonded atoms in black, while the solvent molecules at a distance at most~$1.5r_c$ from the dimer are shaded in green.

\begin{figure}
    \centering
    \figorbox{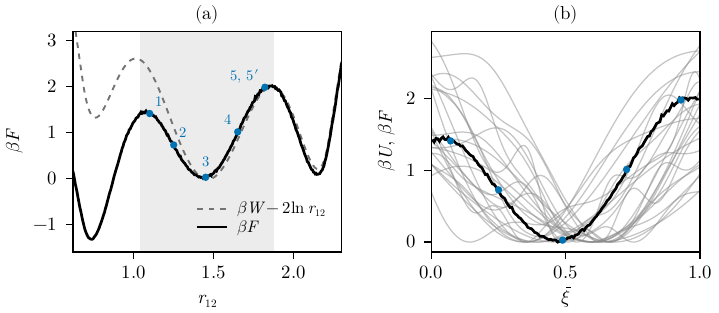}
    \caption{(a) Free energy of the dimer length~\(r_{12}\) (solid line) and unsolvated free energy (dashed line). The state~\(\Omega\) is shaded, and the initial points are indicated in blue. (b) Free energy as a function of~\(\overline\xi\) in~\(\Omega\), with draws~\(\beta U\) from the training prior (grey lines).}
    \label{fig:dimer_system}
\end{figure}

In Figure~\ref{fig:dimer_system}, we show free energy profiles for the system in the interatomic distance~\(r_{12}\). The dashed line in the left panel corresponds to the unsolvated free energy (the case~\(N_{\rm at}=2\)) which can be computed in closed form from~\(W\) and a polar change of variables, expressing~\(\mu\) in the coordinate~\(r_{12}\). This change of variables gives an explicit contribution from the Jacobian, which is the entropic correction in the Helmholtz free energy, reflecting the fact that low-$r_{12}$ configurations have low entropy.
The solid curve corresponds to the solvated free energy~\(F\), computed using a simple histogram estimator~(see~\cite[Section 2.5.1.1.]{LRS10}) from a reference trajectory of the equilibrium dynamics (see Equation~\eqref{eq:overdamped_langevin} below). The shaded area corresponds to the metastable state~\(\Omega\), and the blue points~\(1,2,3,4,5,5'\) correspond to reference initial configurations to be used in our experiments. On the right panel, we represent the restriction to~\(\Omega\) of the free energy in the coordinate~\(\overline\xi\) (solid black line), superimposed with several samples of~\(\beta U\) from the training prior (grey lines). This suggests that the distribution of~\((\beta,U)\) is reasonably calibrated to this system, but this is only a heuristic interpretation, as~\(F\) has no simple relationship to the canonical model~\(Z\).

\subsubsection{Molecular dynamics}
\label{sec:dimer_md}
The system evolves under the overdamped Langevin dynamics, which is defined by the SDE
\begin{equation}
    \label{eq:overdamped_langevin}
    \d \overline{X}_t = -D\nabla V(\overline{X}_t)\,\d t + \sqrt{\frac{2}\beta}D^{1/2}\,\d W_t,
\end{equation}
where~\(W\) is a standard~\(3N_{\rm at}\)-dimensional Brownian motion, and the diffusion matrix is a diagonal matrix
\begin{equation}
    D = D_d(\rmI_{1}+\rmI_{2}) + \sum_{i\geq 3} \rmI_i,\qquad D_d>0,
\end{equation}
with~\(I_i\) acting as the identity on the~\(i\)-th atom's degrees of freedom, and zero otherwise. The coefficient~\(D_d>0\) is used to control the mobility of the dimer, which will be useful to control whether~\(\overline{\xi}\) is a slow (low~\(D_d\)) or a fast~(high~\(D_d\)) variable, and therefore whether the system lies in the physical regime described by  the assumptions of Theorem~\ref{thm:cg_existential_result}.
Broadly speaking, the case~\(D_d=0.05\) is a toy model for a heavy molecule in water solution, with the dimer length representing some global geometric feature of the solute, while the case~\(D_d=1\) is rather representative of light solutes.

In all simulations, the trajectories of~\eqref{eq:overdamped_langevin} (and any derived Fleming--Viot process) are discretized using an Euler--Maruyama scheme with a timestep~\(\Delta t = 10^{-4}\).

Molecular dynamics simulations were implemented in Julia, using functionalities of the package Molly~\cite{molly}.

\subsubsection{Reference decorrelation times and scale separation}
\label{sec:dimer_ref}

As the honest decorrelation time~\eqref{eq:time_true} is sensitive to the initial condition~\(\eta_0^X\), we perform our experiments for six deterministic initial conditions~\(q_0\in\Omega\), constructed as follows.
First, the dimer is frozen at a length~\(r_0\in\{1.1,1.25,1.45,1.65,1.82\}\), corresponding to the five blue points depicted in Figure~\ref{fig:dimer_system}. Then the solvent molecules are added to the system, and relaxed under overdamped Langevin dynamics for~\(t_{\rm th}=20\), where the dimer-solvent WCA interaction is treated as an external potential, keeping the dimer configuration fixed.
For the most stretched initial point~\(r_0=1.82\), two initial configurations were sampled in this way. These are labeled points~\(5\) and~\(5'\) in the results, while the others are labeled~\(1,2,3\) and~\(4\).

For each initial configuration, \(M=3000\) independent Fleming--Viot simulations for the one-particle dynamics~\eqref{eq:overdamped_langevin} killed at~\(\partial\Omega\) were performed, using \(N=72\) replicas each.
Estimators~\(\widehat{\eta}_t^{\overline{X}}\) of~$\eta_t^{\overline{X}}$ are constructed by aggregating over independent simulations
\begin{equation}
    \label{eq:nl_flow_estimator}
    \widehat{\eta}_t^{\overline{X}} = \frac1M \sum_{m=1}^M \widehat{\eta}^{\overline{X},(m)}_{t,N},
\end{equation}
where~\(\widehat{\eta}^{\overline{X},(m)}_{t,N}\) denotes the empirical measure of the~\(m\)-th Fleming--Viot simulation. Traces of~\((\overline{\xi},n_{\rm coord})_\star\widehat{\eta}_t^{\overline{X}}\) were recorded during the simulations.
Simulations were run for~\(t_{\rm ref} = 3\), and the second half of these~\(\widehat{\eta}_t^{\overline{X}}\) for~\(t\in[1.5,3]\) was averaged to estimate the QSD
\begin{equation}
    \label{eq:qsd_estimator}
    \widehat{\nu}^{\overline{X}} = \frac{2}{t_{\rm ref}}\int_{t_{\rm ref}/2}^{t_{\rm ref}}\widehat{\eta}_t^{\overline{X}}\,\d t,
\end{equation}
with time-averages replaced by averages over the numerical trajectory.

Reference decorrelation times do not rely on direct estimators~\(\d_{\rm TV}(\eta_t^{\overline{X}},\nu^{\overline X})\), which are computationally intractable in this setting.
Instead, we estimate \(\d_{\rm TV}(f_\star\eta_t^{\overline{X}},f_\star\nu^{\overline X})\) for~\(f\in\{\overline{\xi},n_{\rm coord},(\overline{\xi},n_{\rm coord})\}\). When~\(f=\overline{\xi}\) is a slow variable (in the sense of Theorem~\ref{thm:cg_existential_result}), the generalization of~\eqref{eq:diagnostic_bound} to dynamics such as~\eqref{eq:overdamped_langevin} would imply that quasi-stationarity in~\(\overline\xi\) predicts (approximate) quasi-stationarity in the full system.
In our first two experiments, the role of both on-the-fly diagnostics is to detect quasi-stationarity of~\(\overline{\xi}_\star \eta_t^{\overline X}\).
The cases~\(f\in\{n_{\rm coord},(\overline{\xi},n_{\rm coord})\}\) are included at this point to monitor failures in the assumption that~\(\overline{\xi}\) is a slow variable, and therefore failures of our framework to justify the use of a marginal diagnostic in~$\overline{\xi}$.

Total variation distances between marginals are estimated using histograms. For each~\(f\), we fix a regular partition of the range of~\(f\): \(d_{\rm hist}=64\) equal bins for~$f=\overline\xi,n_{\rm coord}$ (where the range of~$n_{\rm coord}$ is clamped to~$[4,10]$), and a partition of~$(0,1)\times (4,10)$ into~\(8\times 8\) regular bins for~\(f=(\overline\xi,n_{\rm coord})\).
In each case, we denote by~\({\rm Hist}_f\) the corresponding histogram map, which sends a point probability measure to a probability measure with piecewise-constant Lebesgue-density.
We then construct an estimate for~\(\d_{\rm TV}(f_\star\eta_t^{\overline{X}},f_\star\nu^{\overline X})\) via
\begin{equation}
    \label{eq:tv_estimator_raw}
    \widehat{D}_f(t) = \left\|{\rm Hist}_f\left(f_\star\widehat\eta_t^{\overline X}\right)-{\rm Hist}_f\left(f_\star\widehat\nu^{\overline X}\right)\right\|_{L^1},
\end{equation}
and define reference decorrelation times in~\(f\) as
\begin{equation}
    \label{eq:dimer_tcorr_ref}
    t_0(f) = \inf\left\{t\geq 0:\,\,\widehat{D}_f(s)<\varepsilon_{\rm corr}\,\forall\,s\in[t,t+t_{\rm h}]\right\},
\end{equation}
where~\(t_{\rm h}=0.05\) for~\(D_d=1\) and~\(t_{\rm h}=0.2\) for~\(D_d=0.05\).

\begin{figure}
    \centering
    \figorbox{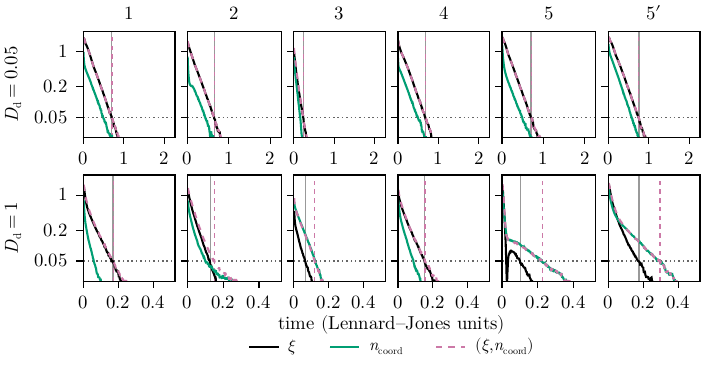}
    \caption{Histogram distances~\(\widehat D_f(t)\)~\eqref{eq:tv_estimator_raw} to their quasi-stationary state as a function of time, for~\(f=\overline\xi,n_{\rm coord}\) and~\((\overline\xi,n_{\rm coord})\). Columns correspond to different initial conditions, and rows to different values of the parameter~\(D_d\). Reference decorrelation times~\(t_0(f)\) defined in~\eqref{eq:dimer_tcorr_ref} are indicated with vertical lines (solid gray:~$f=\overline\xi$, dashed pink:~$f=(\overline{\xi},n_{\rm coord}))$. The tolerance threshold~\(\varepsilon_{\rm corr}\) is indicated with dotted horizontal lines.}
    \label{fig:dimer_fiber}
\end{figure}

In Figure~\ref{fig:dimer_fiber}, we show decorrelation plots of~\(\widehat{D}_{f}(t)\) for each of the three observables~\(f\) we consider, each of the six initial configurations and both values of the diffusion coefficient~\(D_d\in\{0.05,1\}\). We confirm that the decorrelation time is generally sensitive to the choice of initial conditions, with configuration~\(3\), located near the free-energy minimum (see Figure~\ref{fig:dimer_system}) leading to significantly shorter decorrelation times in all cases.
Moreover, we observe that histograms in~\(\overline{\xi}\) routinely converge before those in~\(n_{\rm coord}\) in the case~\(D_d=1\) (and therefore~$\overline{\xi}$ fails to be a slow variable), whereas in the case~\(D_d=0.05\), decorrelation times (in our marginal definition~\eqref{eq:dimer_tcorr_ref}, indicated by vertical lines) are always predicted by~\(\overline{\xi}\). Therefore we informally expect the use of the RNN diagnostic in~$\overline{\xi}$ to be sound in the case~\(D_d=0.05\), but not~\(D_d=1\), as in this latter case,~\(\overline{\xi}_\star\eta_t^{\overline X}\) may not be approximable by a non-linear Markovian probability flow. However we note that scale separation between~\(n_{\rm coord}\) and~\(\overline{\xi}\) is at most mild even in the case~\(D_d=0.05\). Therefore, the system is not deep inside the adiabatic regime of Section~\ref{sec:perturbation_theory}.
The choice of~\(t_{\rm h}>0\) in~\eqref{eq:dimer_tcorr_ref} only has a significant effect when~\(D_d=1\) in initial configuration~\(5\).

\subsubsection{On-the-fly decorrelation detection}
\label{sec:dimer_otf}
For each value~\(D_{d}\in\{0.05,1\}\), each initial point and each design parameter~\((N,\tau)\), with \(N\in\{25,50,100\}\) and \(\tau/\Delta t\in\{100,200,400\}\),~\(20\) independent Fleming--Viot simulations are performed. Each simulation is observed up to~\(10 \max(t_0(\overline\xi),t_0(\overline\xi,n_{\rm coord}) )\).
The RNN model from Section~\ref{sec:num_implementation}, trained on synthetic trajectories, is applied without retraining. Fleming--Viot histogram features in~\(\overline\xi\) are aggregated on-the-fly, and the scalar~\(m(N,\tau)=\sqrt{N\tau}\) is used as input in the recurrent layer without rescaling (the lag time~\(\tau\), in Lennard--Jones units, is therefore directly compared to the synthetic time units used in training). The Gelman--Rubin diagnostic is computed using the collective variable~\(\overline\xi\) only~(i.e. \(\cO=\{\overline{\xi}\}\) in~\eqref{eq:gr_decorrelation_time}).
We note that adding more observables to~\(\cO\) can only make the Gelman--Rubin diagnostic more conservative, as we shall see in Section~\ref{sec:dimer_mitigation} below.

In Figure~\ref{fig:dimer_detection}, we plot distributions of detected decorrelation times for the solvated dimer, in units of~\(t_0(\overline{\xi})\). The nonconvergence of~\(n_{\rm coord}\) in the case~\(D_d=1\) is hidden from both diagnostics in this experiment, and therefore not taken into account for evaluation.
For~$D_d=0.05$, the RNN is well-calibrated with respect to this reference in terms of empirical false decorrelation risk, and both diagnostics display the same overall behavior as the one observed in Figure~\ref{fig:synth_detection}. Namely, the RNN decorrelation times have a sharp mode around~$t_0(\overline\xi)$, while the Gelman--Rubin decorrelation times have a spread-out distribution, are highly sensitive to~$\alpha$, overestimate the reference, and exceed~$10t_0(\overline\xi)$ on a significant portion of trajectories. The RNN never fails to detect decorrelation within this time horizon.
For~\(D_d=1\), we observe a significant miscalibration in the RNN, with an empirical false decorrelation risk of~\(57\%\) for~\(\alpha=0.1\) and~\(41\%\) for~\(\alpha=0.05\). For~$D_d=1$, the distribution of decorrelation times is also shifted to the left for Gelman--Rubin, leading to an overall better diagnostic for~$D_d=1$. The empirical risks of false decorrelation for the Gelman--Rubin diagnostic are~\(11\%\) for~\(\alpha=0.1\)  and~\(0\%\) for~\(\alpha=0.05\), while the corresponding failure rates are~\(0\%\) and~\(2\%\).
Since the false decorrelation risk is defined with respect to the approximate definition~\eqref{eq:dimer_tcorr_ref}, these values should not be interpreted quantitatively. It nevertheless appears that the RNN diagnostic is qualitatively miscalibrated on this system, which is expected from the failure of Assumption~\tagref{B1}. We explore the impact of this miscalibration on exit distribution biases in our next experiment.
The summary statistics reported in Section~\ref{sec:num_synth} are also reported for this experiment, in the last lines of Tables~\ref{tab:robustness_nn} and~\ref{tab:robustness_gr}, along with the ROC AUC scores for both diagnostics, where the RNN again outscores the Gelman--Rubin diagnostic for both~\(D_d=1\) and~\(D_d=0.05\).

We speculate that the risk miscalibration of the RNN could be explained by two factors. A first explanation could be that a higher effective diffusion in~\(\overline{\xi}\) for~$D_d=1$ lies outside the range of the training prior. Such a failure could be addressed directly by changing the prior. However, the results of Table~\ref{tab:robustness_nn} indicate that high \(D\) prior misspecifications do not induce early detections in the synthetic setting.
We therefore believe that the main cause is more fundamental: the fact that~\(\overline{\xi}\) is not slow for~\(D_d=1\) (according to Figure~\ref{fig:dimer_fiber}) may cause non-Markovian features to appear in the~\(\overline{\xi}\) time-marginals of the non-linear probability flow.
Since such features were never observed in training, the recurrent model's inputs are out of distribution, its outputs extrapolative, and its predictions are therefore unreliable.
We believe this failure mode can also be addressed in several ways, to be explored in future work: we will come back to this in Section~\ref{sec:perspectives} below.

It is interesting to note that the Gelman--Rubin diagnostic performs better where its assumptions~\tagref{A1}--\tagref{A3} suggest it should fail. We doubt whether this is a systematic effect, but rather interpret it as an indication that the relationship between the decision threshold~\(\alpha\) and the risk of false decorrelation is system-dependent, which prevents~\(\alpha\) from being selected a priori for the Gelman--Rubin diagnostic over a broad range of systems.

\begin{figure}
    \centering
    \figorbox{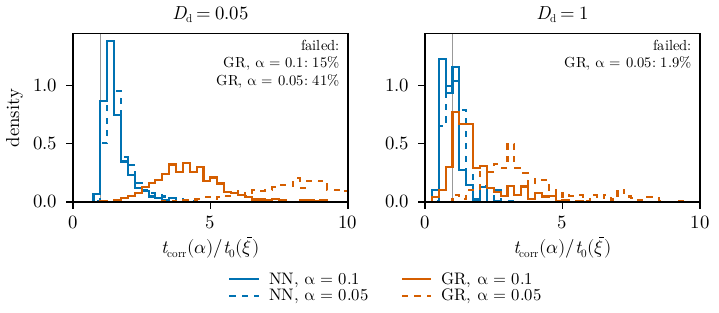}
    \caption{Distributions of the declared decorrelation times~\(t_{\rm corr}(\alpha)\), in units of~\(t_0(\overline\xi)\), for the RNN (blue) and Gelman--Rubin (orange) diagnostics on the solvated dimer, for~\(\alpha=0.1\) (solid lines) and~\(\alpha=0.05\) (dashed lines). The proportion of trajectories on which each diagnostic fails is indicated in each panel.}
    \label{fig:dimer_detection}
\end{figure}

\subsubsection{Bias on the exit event}
\label{sec:dimer_bias}
We now investigate the effect of the choice of on-the-fly diagnostic on the exit distribution~$(\tau^\Omega,\overline{X}_{\tau^\Omega})$ from~\(\Omega\) (see~\eqref{eq:exit_distribution}), starting the dynamics~\eqref{eq:overdamped_langevin} from a replica from the Fleming--Viot state at the decorrelation time of each diagnostic.
When the dynamics starts from an ideal sample from the QSD, exit points and exit times from~\(\Omega\) are independent, and exit times are exponentially distributed. Moreover, the exit point is independent of the starting configuration. Checking these properties empirically provides a way to indirectly validate decorrelation time definitions (on-the-fly or not) without resorting to an approximate reference such as~\eqref{eq:dimer_tcorr_ref}.

At the decorrelation time of each diagnostic (or at time~\(10t_0(\overline\xi)\) when the diagnostic fails) in the experiment of Section~\ref{sec:dimer_otf}, each of the~\(N\) replicas of its Fleming--Viot process is taken as the initial point of an independent trajectory of the dynamics~\eqref{eq:overdamped_langevin}, which is simulated until it exits~\(\Omega\). For each replica, we record the exit time and exit point in this way. As the exit point can only take two values in the continuum limit, we only record whether the dimer exits to the left (to the compact state) or to the right (to the stretched state). Samples are then aggregated over replicas and trajectories.
We also compute a reference exit event distribution in the same way, starting from the final state (after~\(t_{\rm ref}=3\) Lennard--Jones time units, when we assume every system is decorrelated, see Figure~\ref{fig:dimer_fiber}) of~\(48\) reference Fleming--Viot simulations per initial configuration and value of~$D_d$, using~\(72\) replicas each, giving a total of~\(3456\) reference exit events per~$(q_0,D_d)$, and~$41472$ exit events in total.

\begin{figure}
    \centering
    \figorbox{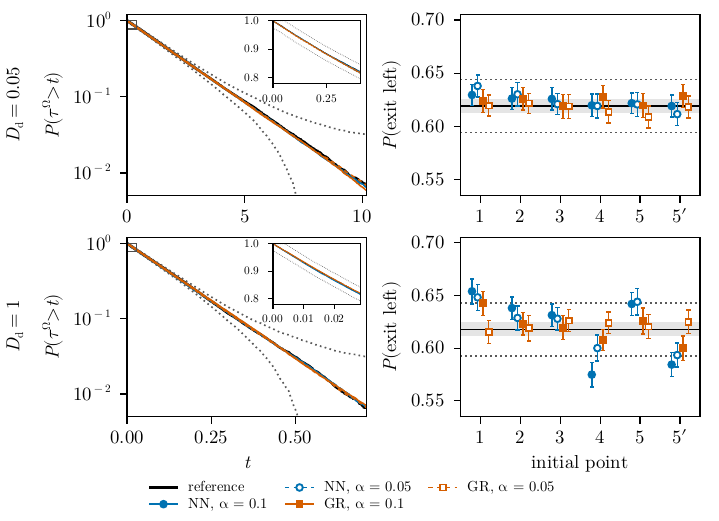}
    \caption{Exit distribution from diagnosed samples, for~\(D_d=0.05\) (top) and~\(D_d=1\) (bottom). The left panels show empirical survival functions for the exit time, with the short-time behavior represented in the insets. Right panels show the empirical probability of exiting through~\(\overline\xi=0\) for each initial point, with~\(95\%\) confidence intervals, treating Fleming--Viot particles as independent. The reference values and its confidence interval are indicated by the black line and grey band, and the heuristic error level~\(\pm\varepsilon_{\rm corr}/2\) by dotted lines.}
    \label{fig:dimer_exits}
\end{figure}

In Figure~\ref{fig:dimer_exits}, we compare empirical exit distributions for the reference initial points and the decorrelated samples according to each diagnostic of the previous subsection.
The left panel compares survival probabilities~\(t\mapsto \P{\tau^\Omega>t}\), while the right panel compares probabilities of exiting through the left point~$(r_{12}({\overline{X}}_{\tau^\Omega})=r_{\rm L}$), which corresponds to a transition to the compact state. In both panels, grey dotted lines correspond to heuristic~\(\pm\varepsilon_{\rm corr}/2\) error bars around the reference mean, corresponding to the method's target error level from Proposition~\ref{prop:exit_bias}. On the right panel, the grey shaded region corresponds to a~\(\pm1.96\sigma\) confidence interval for the reference probabilities.

The choice of diagnostic has no discernible effect on the distribution of exit times at our statistical resolution, and is in every case indistinguishable from the reference distribution, and from one another.

In the case~\(D_d=0.05\), we observe no bias or dependence on the initial point in the exit distribution, at our statistical resolution. This suggests that all four diagnostics are valid decorrelation times in this case. As the RNN diagnostic is less conservative~(Figure~\ref{fig:dimer_detection}), it is the overall better diagnostic for~$D_d=0.05$.
In the case~\(D_d=1\), we observe more dispersion in the exit probabilities for both the RNN and Gelman--Rubin diagnostics. From initial configurations~$1$ and~$4$ at~$\alpha=0.1$, the RNN exit points appear biased: the empirical probability of a compact transition lies slightly outside the target error range of~$\varepsilon_{\rm corr}/2$. This is consistent with the presence of false decorrelations suggested by the experiment of Section~\ref{sec:dimer_otf}. For other choices of initial configuration or threshold~$\alpha$, any bias is not clearly beyond the target error, at our level of statistical resolution.
The Gelman--Rubin exit distribution has no discernible bias.

\begin{figure}
    \centering
    \figorbox{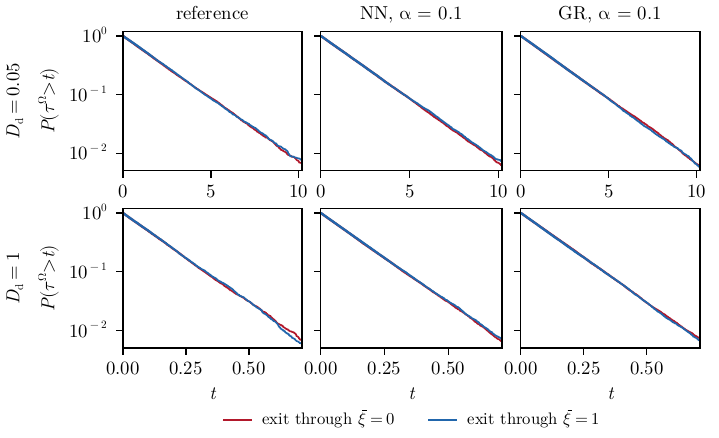}
    \caption{Empirical survival functions of the exit time conditioned on the exit point, through~\(\overline\xi=0\) (red) or~\(\overline\xi=1\) (blue), starting from a Fleming--Viot population. Left panel: reference decorrelated population. Middle panel: population decorrelated according to the RNN. Right panel: population decorrelated according to the Gelman--Rubin diagnostics.}
    \label{fig:dimer_exit_sides}
\end{figure}

In Figure~\ref{fig:dimer_exit_sides}, we investigate the dependence of the distribution of exit times on the exit point, by conditioning the empirical survival probability on the exit point. For the reference population and both diagnostics, for both values of~\(D_d\), the exit times show no dependence on the exit point, consistently with the property~\eqref{eq:exit_distribution} of the QSD.
The plots only show these distributions for the lax setting~\(\alpha=0.1\), as a conservative setting~\(\alpha=0.05\) is expected to decrease any bias.

\subsubsection{Using multiple collective variables}
\label{sec:dimer_mitigation}
We finally address the failure of the RNN diagnostic in the previous section using a simple mitigation strategy. Just as the Gelman--Rubin diagnostic can be defined using multiple collective variables (see~\eqref{eq:gr_decorrelation_time}), we propose applying the RNN diagnostic to multiple CVs at once.
The feature trajectories can be batched to compute predictions efficiently. Here, we apply both diagnostics with~$\cO=\{\overline{\xi},\widetilde n_{\rm coord}\}$, where the coordination number is normalized
\begin{equation}
    \label{eq:ncoord_renormalized}
    \widetilde n_{\rm coord} = \frac16 \texttt{clamp}(n_{\rm coord}-4,[0,6]) \in [0,1].
\end{equation}
Histogram trajectories~$H_k^{\widetilde n_{\rm coord}}$ (see~\eqref{eq:deployment_histograms}) are then recorded, and model predictions~$h_{\theta,k}^{\widetilde n_{\rm coord}}$ (see~\eqref{eq:predicted_probabilities}) are used to define
\begin{equation}
    \widehat{t}_{\rm corr}^{\rm RNN}(\alpha) = \max\left(t_{\rm corr}^{\rm RNN}(\alpha),\tau\times\inf\,\left\{k\geq 1:\,h_{k,\theta}^{\widetilde n_{\rm coord}}>1-\alpha\right\}\right),
\end{equation}
where~$t_{\rm corr}^{\rm RNN}(\alpha)$ is the~$\overline\xi$-marginal RNN decorrelation time. The reference decorrelation time we use is now the joint decorrelation time~$t_0(\overline\xi,n_{\rm coord})$ defined in~\eqref{eq:dimer_tcorr_ref}.

We repeat the experiments of Sections~\ref{sec:dimer_otf} and~\ref{sec:dimer_bias} using these new definitions. Summary statistics are presented in Table~\ref{tab:robustness}.
For~$D_d=0.05$, the behavior of both the RNN and Gelman--Rubin diagnostics shows little difference compared to the single collective-variable experiment.
For~$D_d=1$, the RNN diagnostic becomes more conservative, with an empirical false decorrelation risk of zero compared to the reference, yet still very few failures (the RNN diagnostic shows an empirical failure rate of~$0.1\%$ at~$\alpha=0.05$ in this case).
The Gelman--Rubin diagnostic is again more conservative than the RNN, fails more often, and has a lower ROC-AUC, though only slightly in this case.

\begin{figure}
    \centering
    \figorbox{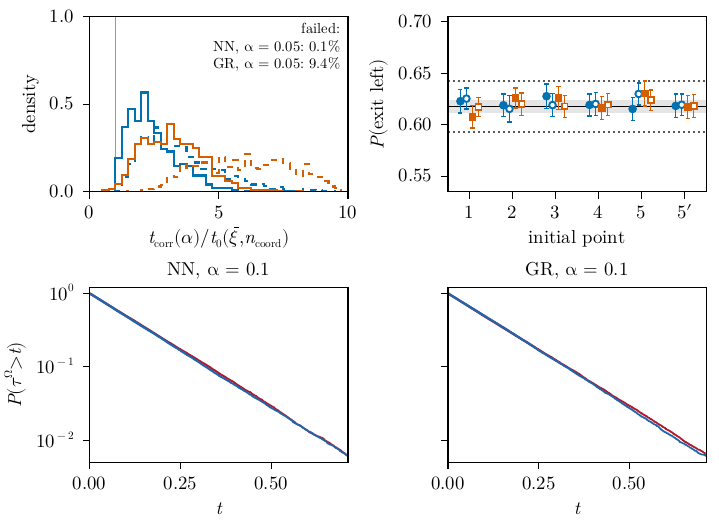}
    \caption{Behavior of the two-variable diagnostics for~$\alpha=0.1$. Top left: distribution of decorrelation times in units of~\(t_0(\overline\xi,n_{\rm coord})\) (see Figure~\ref{fig:dimer_detection}). Top right: exit-side distribution per initial point (see right panels of Figure~\ref{fig:dimer_exits}). Bottom: survival functions conditioned on the exit side, for the RNN (left) and Gelman--Rubin (right, see Figure~\ref{fig:dimer_exit_sides}).}
    \label{fig:dimer_mitigation}
\end{figure}

In Figure~\ref{fig:dimer_mitigation}, we show the equivalents of Figures~\ref{fig:dimer_detection},~\ref{fig:dimer_exits} and~\ref{fig:dimer_exit_sides} using these new definitions of decorrelation times.
The same qualitative behavior as in well-calibrated situations can be observed at the level of decorrelation time distributions. No bias is detectable in the exit distribution.

We stress that this mitigation strategy is heuristic, as marginals of the Fleming--Viot process in~$\widetilde n_{\rm coord}$ are not well-represented by the training prior.
In particular, replicas may be killed at the center of the~$\widetilde{n}_{\rm coord}$ domain, rather than only at the boundary for~$\overline\xi$. It is also possible to include a soft killing rate in the training prior to represent such dynamics.

\subsubsection{Computational cost and summary statistics}
The cost of one RNN inference (on a single histogram feature) is comparable to the cost of one MD step for the dimer system on a laptop.
While we have not performed systematic benchmarks for an optimized implementation, this anecdotal evidence suggests that the RNN represents a negligible portion of the total computational cost, while the Gelman--Rubin diagnostic is cheaper still.
As the cost of RNN inference is independent of the number of atoms in the system and its relative cost decreases as the lag time~\(\tau\) increases, we therefore expect the approach to be scalable to much larger systems with negligible computational cost in deployment.

Table~\ref{tab:robustness} summarizes some of the numerical results of the experiments presented in Sections~\ref{sec:num_synth} and~\ref{sec:num_dimer}.

\begin{table}[H]
    \centering\small\setlength{\tabcolsep}{3.5pt}
    \begin{subtable}{\linewidth}
        \centering
        \IfFileExists{tables/robustness_table_nn.tex}{\begin{tabular}{l c ccc ccc}
\toprule
 & & \multicolumn{3}{c}{$\alpha=0.1$} & \multicolumn{3}{c}{$\alpha=0.05$} \\
\cmidrule(lr){3-5}\cmidrule(lr){6-8}
shift & ROC-AUC & $t_{\mathrm{corr}}/t_0$ & early & failed & $t_{\mathrm{corr}}/t_0$ & early & failed \\
\midrule
training prior & 0.962 & $1.40\ (1.19, 1.83)$ & 5\% & 0\% & $1.59\ (1.29, 2.19)$ & 2\% & 1\% \\
high $D$ & 0.888 & $2.00\ (1.50, 3.00)$ & 1\% & 1\% & $2.50\ (1.80, 4.00)$ & 0\% & 5\% \\
low $T$ & 0.919 & $1.76\ (1.30, 3.12)$ & 5\% & 10\% & $2.17\ (1.46, 4.80)$ & 2\% & 16\% \\
low $T$, high $D$ & 0.847 & $2.50\ (1.67, 4.33)$ & 1\% & 14\% & $3.33\ (2.00, 7.00)$ & 0\% & 22\% \\
low $T$, one well & 0.941 & $1.71\ (1.33, 2.75)$ & 2\% & 8\% & $2.00\ (1.46, 3.81)$ & 1\% & 13\% \\
\midrule
dimer, $D_{\mathrm{d}}=0.05$ & 0.988 & $1.42\ (1.26, 1.72)$ & 2\% & 0\% & $1.55\ (1.34, 1.91)$ & 0\% & 0\% \\
\quad with $n_{\rm coord}$ & 0.975 & $1.44\ (1.28, 1.74)$ & 1\% & 0\% & $1.65\ (1.42, 2.11)$ & 0\% & 0\% \\
dimer, $D_{\mathrm{d}}=1$ & 0.941 & $0.86\ (0.67, 1.00)$ & 57\% & 0\% & $1.00\ (0.81, 1.25)$ & 41\% & 0\% \\
\quad with $n_{\rm coord}$ & 0.933 & $2.20\ (1.69, 2.93)$ & 0\% & 0\% & $3.14\ (2.33, 4.25)$ & 0\% & 0\% \\
\bottomrule
\end{tabular}
}{\fbox{missing table: tables/robustness\_table\_nn.tex}}
        \caption{RNN}
        \label{tab:robustness_nn}
    \end{subtable}

    \vspace{1em}
    \begin{subtable}{\linewidth}
        \centering
        \IfFileExists{tables/robustness_table_gr.tex}{\begin{tabular}{l c ccc ccc}
\toprule
 & & \multicolumn{3}{c}{$\alpha=0.1$} & \multicolumn{3}{c}{$\alpha=0.05$} \\
\cmidrule(lr){3-5}\cmidrule(lr){6-8}
shift & ROC-AUC & $t_{\mathrm{corr}}/t_0$ & early & failed & $t_{\mathrm{corr}}/t_0$ & early & failed \\
\midrule
training prior & 0.817 & $4.78\ (3.60, 6.45)$ & 1\% & 10\% & ${>}10\ (7.50, {>}10)$ & 0\% & 50\% \\
high $D$ & 0.866 & $4.00\ (3.00, 5.33)$ & 0\% & 7\% & $9.00\ (5.93, {>}10)$ & 0\% & 48\% \\
low $T$ & 0.802 & $5.00\ (3.19, 7.25)$ & 2\% & 13\% & ${>}10\ (6.97, {>}10)$ & 0\% & 54\% \\
low $T$, high $D$ & 0.826 & $4.25\ (2.90, 6.08)$ & 1\% & 11\% & ${>}10\ (6.00, {>}10)$ & 0\% & 52\% \\
low $T$, one well & 0.802 & $5.16\ (3.43, 7.35)$ & 1\% & 13\% & ${>}10\ (7.44, {>}10)$ & 0\% & 57\% \\
\midrule
dimer, $D_{\mathrm{d}}=0.05$ & 0.784 & $4.40\ (3.56, 5.63)$ & 0\% & 15\% & $9.18\ (7.61, {>}10)$ & 0\% & 41\% \\
\quad with $n_{\rm coord}$ & 0.805 & $4.39\ (3.55, 5.62)$ & 0\% & 15\% & $9.17\ (7.61, {>}10)$ & 0\% & 41\% \\
dimer, $D_{\mathrm{d}}=1$ & 0.876 & $1.50\ (1.19, 2.00)$ & 11\% & 0\% & $3.24\ (2.56, 4.33)$ & 0\% & 2\% \\
\quad with $n_{\rm coord}$ & 0.923 & $3.11\ (2.27, 4.00)$ & 1\% & 0\% & $6.22\ (4.67, 7.75)$ & 0\% & 9\% \\
\bottomrule
\end{tabular}
}{\fbox{missing table: tables/robustness\_table\_gr.tex}}
        \caption{Gelman--Rubin}
        \label{tab:robustness_gr}
    \end{subtable}
    \caption{Summary statistics for the RNN (a) and Gelman--Rubin (b) diagnostics in the experiments of Section~\ref{sec:num}. The column~\(t_{\rm corr}/t_0\) gives \(50\%\,(25\%,75\%)\) quantiles of~\(t_{\rm corr}(\alpha)/t_0\) over all trajectories. Failed trajectories with~\(t_{\rm corr}(\alpha)>10t_0\) are reported as~\({>}10\). The columns ``early'' and ``failed'' give the empirical false decorrelation risk~(\(t_{\rm corr}(\alpha)<t_0\)) and failure risk~(\(t_{\rm corr}(\alpha)>10t_0\)), respectively.
        The reference~\(t_0\) corresponds to~\eqref{eq:synth_tcorr_ref} for the synthetic experiments, and~\eqref{eq:dimer_tcorr_ref} for the molecular systems (``with~$n_{\rm coord}$'' corresponds to~$f=(\overline\xi,n_{\rm coord})$, or else~$f=\overline\xi$ only).}
    \label{tab:robustness}
\end{table}

\section{Discussion and perspectives}
\label{sec:perspectives}
We conclude by summarizing the main features and limitations of our proposed method, and outlining some perspectives for future work addressing these limitations.

\begin{itemize}
    \item[\(\circ\)]\emph{Deployability in accelerated molecular dynamics algorithms.} The RNN we propose only requires periodic observations of a Fleming--Viot process in the collective variable~\(\overline{\xi}\). As this is precisely the operational setting of the generalized parallel replica algorithm~\cite{BLS15,HL19}, an RNN diagnostic could replace the Gelman--Rubin diagnostic in these algorithms with minor implementation difficulties. Even on the small system considered in Section~\ref{sec:num_dimer}, the computational cost of inference will likely remain negligible compared to the cost of evaluating the interatomic forces.
    \item[\(\circ\)]\emph{Physical information as a modeling input.} Our method uses prior knowledge of the system, which is encoded in the definition of the prior~\(\rho_{\rm synth}\) over plausible surrogate processes (Assumption~\tagref{B4}). Our method therefore occupies a middle ground between MCMC-inspired diagnostics (which require no prior information, but appear conservative and difficult to calibrate) and deterministic definitions of the decorrelation time (which only apply in specific physical regimes). The experiments of Section~\ref{sec:num_synth} show empirically that this prior can be defined rather broadly (see Figures~\ref{fig:synth_tcorr} and~\ref{fig:synth_dynamics}), with the diagnostic nevertheless retaining high discriminative power (see Figure~\ref{fig:synth_roc}) and robustness with respect to the parameters \((N,\tau)\) (see Figure~\ref{fig:synth_auc_grid}). As a broad range of effective models is often known in practice, this middle ground may be ideal for certain applications.
    \item[\(\circ\)]\emph{Interpretability.} The decision threshold~\(\alpha\) can be interpreted probabilistically, if the prior is well-calibrated to the deployment conditions: it controls the average risk of false decorrelation over the prior (Lemma~\ref{lemma:risk_control}), which in turn controls the bias on the sampled exit distribution, via Proposition~\ref{prop:exit_bias}. It can therefore be chosen a priori by setting a target error level, rather than calibrated system-by-system as the Gelman--Rubin~\(\alpha\) seems to require. We stress that this interpretation is heuristic in practice, and this guarantee is lost under prior misspecification (see Figure~\ref{fig:ood_calibration}).
    \item[\(\circ\)] \emph{Failure under misspecification.} When the RNN diagnostic's assumptions do not hold, it may fail optimistically, falsely and confidently declaring decorrelation (see the case~\(D_d=1\) in Figure~\ref{fig:dimer_detection}, where~\(\overline{\xi}\) is not a slow variable of the system). At present, the diagnostic has no mechanism to detect that inputs lie outside the support of its training distribution. If such a mechanism existed, a simple mitigation would be to apply the RNN diagnostic in several collective variables concurrently (just as the Gelman--Rubin diagnostic does, see~\eqref{eq:gr_decorrelation_time}), and discard predictions from out-of-distribution trajectories. As demonstrated in Section~\ref{sec:dimer_mitigation}, a simpler heuristic mitigation consists in taking the most conservative prediction across multiple CVs.
    \item[\(\circ\)] \emph{Choice of collective variables.} In our experiments, we restricted ourselves to one-dimensional collective variables. The same method should nevertheless remain practical as long as the honest decorrelation times \(t_0(\mathcal L^Z,Z_0)\) can be computed cheaply for~\(\cL^Z\) in the training distribution. For grid-based quantizations of an~\(m\)-dimensional CV~\(\overline{\xi}\), the size of \(\mathcal Z\) scales as \(d_{\rm grid}^m\), which should make training a model possible for \(m\in\{2,3,4\}\) at most, if matrix-free methods are used for the computation of reference decorrelation times and coarser values of~$d_{\rm grid}$ are considered. Many complex systems require higher-dimensional CVs to accurately resolve the global landscape of metastable states. However, the CVs used in this method are only required to describe decorrelation within a single, metastable state. In other words, the collective variable~$\xi$ may change from state to state. We therefore expect in many cases this local description to involve fewer effective degrees of freedom than the global landscape. Statistical methods for the automatic detection of slow variables have been developed (in particular TICA~\cite{MS94} and its variants), which may be combined with our approach in future work.
\end{itemize}

Several directions are left open for future work.

\begin{itemize}
    \item[\(\circ\)] \emph{Coarse-graining for killed diffusions.} The analysis of Section~\ref{sec:theory} is at present restricted to finite-state jump processes. A natural extension, relevant to the MD context, would be to extend the result to elliptic diffusions killed at the boundary of a domain, possibly in a small-parameter adiabatic regime analogous to the setting of Section~\ref{sec:perturbation_theory}. We expect this to be a mathematically interesting problem. A subsequent step would be the analogous result for hypoelliptic diffusions, whose spectral properties are much more delicate to analyze.
    \item[\(\circ\)] \emph{Accelerated MD simulations on large systems.} We plan to implement the RNN diagnostic for higher-dimensional CVs, and to deploy it in the generalized parallel replica algorithm to test its usefulness for the accelerated simulation of larger glassy materials and biomolecular systems.
    \item[\(\circ\)] \emph{Relaxing Assumption~\tagref{B1}.} The purpose of Assumption~\tagref{B1} is to justify the use of a prior over Markovian dynamics. It is however possible to enrich the training prior with non-Markovian effective dynamics, in the form of low-dimensional projections \(\xi(Y)\) of some higher-dimensional process \(Y\). If \(Y\) is not very high-dimensional (say of dimension~\(3\) at most), honest decorrelation times for~\(Y\) should remain computable. A few examples from genuinely high-dimensional systems could also be incorporated, where computationally intractable honest decorrelation times could be replaced with expensive approximations, as in Section~\ref{sec:num_dimer}.
    \item[\(\circ\)] \emph{Relaxing Assumption~\tagref{B2}.} The finite-\(N\) implicit error model learned by the RNN is fitted on discrepancies between \(\widehat\eta^Z_{t,N}\) and \(\eta^Z_t\) within the training distribution. Its deployment to \(\xi_\star\widehat\eta^X_{t,N}\) is at present not justified by the mean-field analysis of Section~\ref{sec:theory}. Proving effective dynamics results at the level of paths of the Fleming--Viot process would address this gap.
    \item[\(\circ\)] \emph{Relaxing Assumption~\tagref{B4}.} We plan to explore automatic methods to define the prior \(\rho_{\rm dyn}\), for instance by calibrating it so that synthetically generated features are indistinguishable from traces \(\xi_\star\widehat\eta^X_{t,N}\) observed on the target system, using classification-based methods (see~\cite{GDKC18}). In an accelerated simulation setting, this procedure could be deployed online, recalibrating the prior periodically as new metastable states are visited.
\end{itemize}

\paragraph{Acknowledgements.}
This research was supported by the Swiss National Science Foundation (SNSF, Grant No. 10002757).
The author thanks the members of the CERMICS-LCT working group on accelerated MD methods, whose discussions motivated this work;
in particular, Gabriel Stoltz and Tony Leli\`evre for feedback on early versions of this work, and Pierre Monmarch\'e for remarks which inspired Definition~\ref{def:perfect_model}.
\paragraph{AI statement.}
A conversational AI assistant (Claude Opus 4.8) was used for proofreading (limited to grammar and spelling checks), literature search and auditing mathematical arguments. Mistakes in the constants of Theorem~\ref{thm:cg_existential_result} were detected this way, and were subsequently verified and corrected by the author.
The overall problem statement, mathematical framework and detailed proof strategy were elaborated independently by the author.
AI was not used in writing prose or mathematical proofs: all interpretations and scientific conclusions belong to the author. 

The core numerical methodology (synthetic dynamics generation, RNN architecture and training, Fleming--Viot simulations, RNN inference and Gelman--Rubin diagnostic) was implemented by the author in the form of a Julia module.
AI coding agents (Claude Opus 4.8 and Opus 5 in Claude Code) were granted access to this module to write experimental pipelines, data processing and plotting scripts, following detailed specifications supplied by the author.
All generated code was reviewed and edited by the author, and is publicly available~\cite{repo}.

The author assumes full responsibility for the integrity, correctness and reproducibility of the theoretical and numerical results presented in this manuscript.

\appendix
\section{Proof of Theorem~\ref{thm:perturbation_theory}}
\label{sec:proof_perturbation_theory}
In the following, the constants~\(\varepsilon_0,C>0\) denote~\(\varepsilon\)-independent quantities, whose values may be respectively reduced or increased from line to line. To study~\(\cL_\varepsilon\) perturbatively, we consider the rescaled nonnegative operator
\begin{equation}
    \label{eq:rescaled_operator}
    T(\varepsilon)=-\varepsilon\cL_\varepsilon^X = -\cL_{\rm mic}-\varepsilon\left(\cL_{\rm mac} - K\right) = -(R+\varepsilon S)
\end{equation}
as a perturbation of the reducible operator~\(-\cL_{\rm mic}=-R\). We denote by \((\eta_j(\varepsilon))_{1\leq j\leq |\cX|}\) the eigenvalues of~\(T(\varepsilon)\), enumerated with algebraic multiplicity in order of non-decreasing real part.

\paragraph{Projector and semigroup perturbations.}
We first recall some basic tools from perturbation theory~(see~\cite{K95} for a comprehensive discussion). We consider a continuous family of operators~\((H_\varepsilon)_{0\leq\varepsilon<\varepsilon_0}\) on~\(\ell^\infty(\cX)\) satisfying~\(\|H_\varepsilon-H_0\|\leq C\varepsilon\) for some constants~\(C,\varepsilon_0\) and all~\(0\leq \varepsilon<\varepsilon_0\).

All the results we use can be derived from the Dunford formula~\cite[Section I.5.6]{K95}, which states that for any linear operator~\(H\) and a function~\(f\) holomorphic on a neighborhood of~\(\Gamma\) and its interior,
\begin{equation}
    \label{eq:semigroup_integral_representation}
    f(H)P_H(\Gamma) = \frac1{2i\pi}\oint_{\Gamma} f(z)(z-H)^{-1}\,\d z,
\end{equation}
where~\(\Gamma\) is a closed, positively oriented contour in~\(\bbC\) not intersecting~\(\sigma(H)\), and~\(P_H(\Gamma)\) is the spectral projector associated to the part of~\(\sigma(H)\) enclosed by~\(\Gamma\).
Since~\(\sigma(H_\varepsilon)\) depends continuously on~\(\varepsilon\), we may assume (upon reducing~\(\varepsilon_0\)) that~\(\Gamma\) does not intersect any of the~\(\sigma(H_\varepsilon)\) for~\(\varepsilon<\varepsilon_0\), so that
\begin{equation}
    {\underset{z\in\Gamma,0\leq\varepsilon<\varepsilon_0}{\sup}\,\|(z-H_\varepsilon)^{-1}\|=M<\infty}.
\end{equation}
Writing~\(P_\varepsilon(\Gamma)\) for the spectral projector of~\(H_\varepsilon\), we have
\begin{equation}
    \begin{aligned}
        P_\varepsilon(\Gamma)-P_0(\Gamma) & = \frac1{2i\pi}\oint_{\Gamma}\left[(z-H_\varepsilon)^{-1}-(z-H_0)^{-1}\right]\,\d z      \\
                                          & = \frac1{2i\pi}\oint_{\Gamma}(z-H_\varepsilon)^{-1}(H_\varepsilon-H_0)(z-H_0)^{-1}\,\d z
    \end{aligned}
\end{equation}
so that estimating, we obtain
\begin{equation}
    \label{eq:projector_estimate}
    \|P_\varepsilon(\Gamma)-P_0(\Gamma)\|\leq \frac{CM^2\ell(\Gamma)}{2\pi}\varepsilon
\end{equation}
with~\(\ell(\Gamma)\) being the contour length.
In particular, the projector-valued map~\(\varepsilon\mapsto P_\varepsilon(\Gamma)\) is continuous (even locally Lipschitz) at~\(\varepsilon=0\).
Moreover, the projectors~\(P_\varepsilon(\Gamma)\) and~\(P_0(\Gamma)\) are conjugated for~\(\varepsilon<\varepsilon_0\) via
\begin{equation}
    \label{eq:kato_connection}
    P_\varepsilon(\Gamma) = U_\varepsilon^{-1}P_0(\Gamma)U_\varepsilon,\qquad \|U_\varepsilon-\Id\|\leq C\|P_\varepsilon-P_0\|\leq C\varepsilon,
\end{equation}
where~\(U_\varepsilon\) is an isomorphism mapping~\(P_\varepsilon(\Gamma)\ell^\infty(\cX)\to P_0(\Gamma)\ell^\infty(\cX)\). In particular, \({\rm rank}\,P_\varepsilon(\Gamma) = {\rm rank}\,P_0(\Gamma)\).
The operator~\(U_\varepsilon\) is a construction of Kato, see~\cite[Section I.4.6]{K95}, which we will call Kato's connection.

If~\(\Gamma_1\) and~\(\Gamma_2\) are disjoint closed contours enclosed in~\(\Gamma\), whose union encloses the~\(\Gamma\)-part of~\(\sigma(H_\varepsilon)\), then for~\(\varepsilon<\varepsilon_0\),
\begin{equation}
    \label{eq:sum_of_projectors}
    P_\varepsilon(\Gamma) = P_\varepsilon(\Gamma_1)+P_\varepsilon(\Gamma_2),\qquad P_\varepsilon(\Gamma_1)P_\varepsilon(\Gamma_2)=P_\varepsilon(\Gamma_2)P_\varepsilon(\Gamma_1)=0.
\end{equation}
Taking~\(\Gamma\) large enough for~\(H_\varepsilon\) bounded gives the case~\(P_\varepsilon(\Gamma)=\Id\).

If~\({\Gamma\subset \{z\in\bbC:\,\mathfrak{Re}\,z<-h<0\}}\), we have the semigroup estimate
\begin{equation}
    \label{eq:semigroup_estimate}
    \forall\,t\geq 0,\qquad\left\|\e^{H_\varepsilon t}P_\varepsilon(\Gamma)\right\| = \left\|\frac1{2i\pi}\oint_{\Gamma} e^{zt}(z-H_\varepsilon)^{-1}\,\d z\right\|\leq \frac{M\ell(\Gamma)}{2\pi}\e^{-h t}.
\end{equation}

\begin{figure}[ht]
    \centering
    \begin{tikzpicture}[scale=0.68,
            ev0/.style={circle,fill=black,inner sep=0pt,minimum size=4pt},
            eve/.style={circle,fill=blue!70!black,inner sep=0pt,minimum size=3.2pt},
            ev1/.style={cross out,draw=green!45!black,thick,inner sep=0pt,minimum size=3.5pt},
            contour/.style={thick,red!70!black},
            clab/.style={font=\footnotesize,red!70!black},
            gap/.style={<->,densely dotted,thick,gray!70!black},
            guide/.style={densely dotted,gray},
            lab/.style={font=\footnotesize}]
        \begin{scope}
            \node[lab] at (-7.2,3.2) {(a)};
            \draw[->,gray] (-7.6,0) -- (1.8,0) node[right,black,lab] {\(\mathfrak{Re}\)};
            \draw[->,gray] (0,-2.6) -- (0,2.6) node[above,black,lab] {\(\mathfrak{Im}\)};
            \draw[contour] (0,0) circle (1.3);
            \node[clab,above right] at (0.92,0.92) {\(\Gamma^{\rm s}\)};
            \draw[contour] (-4.85,0) ellipse (2.2 and 1.8);
            \node[clab,above] at (-4.85,1.8) {\(\Gamma^{\rm f}\)};
            \draw[guide] (-3,0) -- (-3,-1.6);
            \draw[gap] (-3,-1.6) -- (0,-1.6) node[midway,below,lab,black] {\(\gamma^{\rm f}\)};
            \foreach \x/\y in {0/0,-3/0,-3.6/0.8,-3.6/-0.8,-4.3/0,-4.8/1.3,-4.8/-1.3,-5.4/0.4,-5.4/-0.4,-6.1/1.0,-6.1/-1.0,-6.6/0}
            \node[ev0] at (\x,\y) {};
            \foreach \x/\y in {-0.4/0,-0.82/0,-0.95/0.25,-0.95/-0.25,-3.15/0,-3.5/0.9,-3.5/-0.9,-4.45/0,-4.7/1.2,-4.7/-1.2,-5.5/0.5,-5.5/-0.5,-6.0/1.1,-6.0/-1.1,-6.75/0}
            \node[eve] at (\x,\y) {};
        \end{scope}
        \begin{scope}[xshift=6.3cm]
            \node[lab] at (-3.0,3.2) {(b)};
            \draw[->,gray] (-3.1,0) -- (2.9,0) node[right,black,lab] {\(\mathfrak{Re}\)};
            \draw[->,gray] (0,-2.9) -- (0,2.9) node[above,black,lab] {\(\mathfrak{Im}\)};
            \draw[contour] (0,0) circle (2.6);
            \node[clab,above right] at (1.84,1.84) {\(\Gamma^{\rm s}\)};
            \draw[contour] (-0.8,0) circle (0.38);
            \node[clab,below] at (-0.8,1.1) {\(\varepsilon\Gamma_1^{\rm s}\)};
            \draw[contour] (-1.9,0) ellipse (0.5 and 0.95);
            \node[clab,above] at (-1.5,0.85) {\(\varepsilon\Gamma_2^{\rm s}\)};
            \draw[guide] (-1.75,0) -- (-1.75,-1.55);
            \draw[guide] (-0.8,0.0) -- (-0.8,-1.55);
            \draw[gap] (-1.75,-1.55) -- (-0.8,-1.55) node[midway,below,lab,black] {\(\varepsilon\gamma^{\rm s}\)};
            \node[ev0] at (0,0) {};
            \foreach \x/\y in {-0.8/0,-1.75/0,-2.05/0.45,-2.05/-0.45}
            \node[ev1] at (\x,\y) {};
            \foreach \x/\y in {-1.0/0,-1.55/0,-1.95/0.68,-1.95/-0.68}
            \node[eve] at (\x,\y) {};
        \end{scope}
        \node[ev0] at (-4.6,-3.5) {};
        \node[lab,right] at (-4.45,-3.5) {\(\sigma(-T(0))\)};
        \node[eve] at (-1.6,-3.5) {};
        \node[lab,right] at (-1.45,-3.5) {\(\sigma(-T(\varepsilon))\), \(\varepsilon\in(0,\varepsilon_0)\)};
        \node[ev1] at (3.4,-3.5) {};
        \node[lab,right] at (3.55,-3.5) {\(\sigma(-\varepsilon T^{(1)})\)};
    \end{tikzpicture}
    \caption{Clustering of~\(\sigma(-T(\varepsilon))\) (blue dots, see~\eqref{eq:rescaled_operator}). Black dots represent~\(\sigma(-T(0))=\sigma(\cL_{\rm mic})\). The right panel details the structure of the spectrum inside~\(\Gamma^{\rm s}\), which comes from the probabilistic nature of~\(T^{(1)}\) (see~\eqref{eq:reduced_perturbation} and the Perron--Frobenius argument above).}
    \label{fig:contours}
\end{figure}
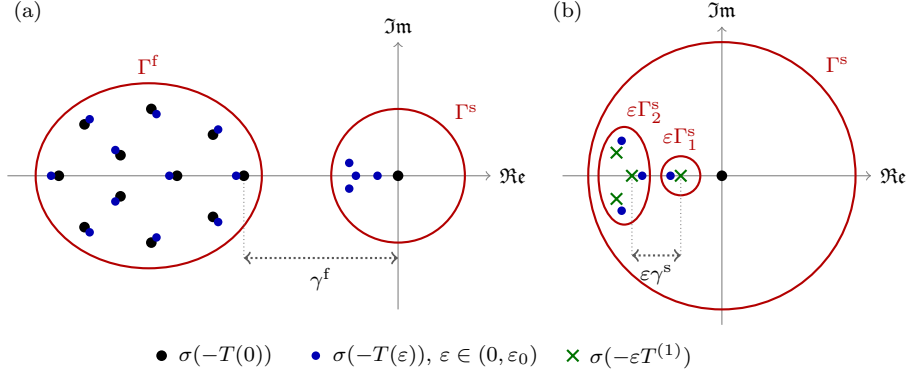

\paragraph{Clustering of~\(\sigma(T(\varepsilon))\).}
We use Kato's reduction process to study perturbations of small eigenvalues of~\(T(\varepsilon)\).
The operator~\(T(0)=-R\) is a direct sum (see Equation~\eqref{eq:block_diagonal}), and each of its blocks~\(A_m\) is the generator of an irreducible Markov jump process. In particular~\(\sigma(-T(0)) = \bigcup_{m} \sigma(A_m)\). The Perron--Frobenius theorem implies that \(0\) is a simple eigenvalue of~\(A_m\), hence a semisimple eigenvalue of \(T(0)\) with multiplicity~\(M\).
We introduce the following quantity
\begin{equation}
    \label{eq:micro_spectral_gap}
    \gamma^{\rm f} = \underset{1\leq m\leq M}{\min}\min\,{\mathfrak{Re}}\left(\sigma(-A_m)\setminus\{0\}\right).
\end{equation}
We introduce two disjoint closed contours~\(\Gamma^{\rm s}\) and~\(\Gamma^{\rm f}\), enclosing~\(0\) and~\({\sigma(-T(0))\setminus\{0\}}\) respectively, such that
\begin{equation}
    \label{eq:micro_contour_inclusions}
    \Gamma^{\rm f} \subset \{z\in\bbC:\,{\mathfrak Re}\, z <-\gamma^{\rm f}/2\}.
\end{equation}
We use the shorthand~\(P_{\varepsilon}^{\rm s}=P_\varepsilon(\Gamma^{\rm s})\) and~\(P_{\varepsilon}^{\rm f}=P_\varepsilon(\Gamma^{\rm f})\).
A schematic representation of~\(\sigma(-T(\varepsilon))\),~\(\Gamma^{\rm s}\) and~\(\Gamma^{\rm f}\) is given in the left panel of Figure~\ref{fig:contours}.

We first study the~\(\Gamma^{\rm s}\)-part of~\(\sigma(-T(\varepsilon))\). Since~\(0\) is a semisimple eigenvalue of~\(-R\),~\cite[Section II.5.4, Theorem 5.4]{K95} shows that \(\eta_j(\varepsilon)\) is positively semi-differentiable at \(\varepsilon=0\) for each~\(1\leq j\leq M\), with an expansion
\begin{equation}
    \label{eq:eigenvalue_expansion}
    \forall\,1\leq j\leq M,\qquad \eta_j(\varepsilon) = \varepsilon \eta_j^{(1)} + \smallo(\varepsilon),
\end{equation}
where the \(\eta_j^{(1)}\) are the eigenvalues of the reduced perturbation
\begin{equation}
    \label{eq:reduced_perturbation}
    T^{(1)} = -P_0^{\rm s} S P_0^{\rm s}.
\end{equation}
Note that the fiber-decomposition~\eqref{eq:block_diagonal} gives in this case an explicit formula
\begin{equation}
    \label{eq:principal_eigenprojector}
    P_0^{\rm s} = \sum_{m=1}^M \1_{\cX_m}\mu_m.
\end{equation}

We order the \((\eta_j^{(1)})_{1\leq j\leq M}\) by non-decreasing real part, and characterize them more precisely by noticing that \(-T^{(1)}\) generates a conditionally irreducible sub-Markovian semigroup on \(P_0^{\rm s}\ell^\infty(\cX)\).
We work in the basis~\(\{\1_{\cX_m},\,1\leq m\leq M\}\) of~\(P_0^{\rm s}\ell^\infty(\cX)\), with respect to which \(-T^{(1)}\) has the matrix representation
\begin{equation}
    \label{eq:perturbation_matrix_form}
    -T^{(1)}_{m,m'} = \mu_{m}(\cL_{\rm mac}-K)\1_{\cX_{m'}},\qquad \forall\, 1\leq m,m'\leq M.
\end{equation}
Since~\(K\1_{\cX_{m'}}\) is supported on~\(\cX_{m'}\) and \(\mu_m\) is supported on~\(\cX_m\), for~\(m\neq m'\), we have
\begin{equation}
    \label{eq:coefficient_expansion}
    -T^{(1)}_{m,m'} = \mu_m\cL_{\rm mac}\1_{\cX_{m'}} = \sum_{x\in\cX_m}\sum_{x'\in\cX_{m'}}\mu_m(x)\cL_{\rm mac}(x,x')\1_{\cX_{m'}}(x')\geq 0,
\end{equation}
since \(x\neq x'\) in the summand and \(\cL_{\rm mac}\) generates a jump process on~\(\cX\). In~\eqref{eq:coefficient_expansion},~\(\cL_{\rm mac}(x,x')\) denotes the~\((x,x')\)-th entry in the canonical matrix representation of~\(\cL_{\rm mac}\).

Since \(\cL_{\rm mac}\1_{\cX}=0\), we have for any \(1\leq m\leq M\),
\begin{equation}
    -\sum_{m'=1}^M T^{(1)}_{m,m'} = \mu_m((\cL_{\rm mac}-K)\1_{\cX})= -\mu_m(K\1_{\cX})\leq 0,
\end{equation}
since \(K\) is nonnegative. Therefore,~\(-T^{(1)}\) generates a sub-Markovian semigroup on~\(P_0^{\rm s} \ell^\infty(\cX)\).

It remains to check the conditional irreducibility of~\(-T^{(1)}\). We say \(x\neq x'\in\cX\) are neighbors if \(\cL_{\rm mac}(x,x')>0\).
First, notice that if \(x\in\cX_m\) and \(x'\in\cX_{m'}\) are neighbors for \(m\neq m'\), then the formula~\eqref{eq:coefficient_expansion} implies that~\(-T^{(1)}_{m,m'}>0\). Indeed, \(\mu_m(x)>0\) for every \(x\in\cX_m\) (by the Perron--Frobenius theorem applied to~\(A_m\)), and each summand in~\eqref{eq:coefficient_expansion} is nonnegative.

Let \(1\leq m\neq m'\leq M\) and \(x\in\cX_m\),~\(x'\in \cX_{m'}\). Since~\(\cL_{\rm mac}\) is irreducible, there exists a chain of neighbors~\(x'=x_1\neq\dots \neq x_p=x\) in \(\cX\).
From this chain, we extract the sequence of jumps between fibers,~\(\left((x_{i_k},x_{i_k+1})\right)_{1\leq k\leq q}\), where~\(1\leq i_{k}< p\),~\(\xi(x_{i_k})\neq \xi(x_{i_k +1})\),~\(\xi(x_{i_k+1}) = \xi(x_{i_{k+1}})\) and~\(\xi(x_{i_1})=m'\),~\(\xi(x_{i_q+1})=m\).
Since~\(x_{i_k}\) and~\(x_{i_k +1}\) are neighbors in distinct fibers, by the previous remark, we have identified a chain~\(\left(-T^{(1)}_{\xi(x_{i_k}),\xi(x_{i_k +1})}\right)_{1\leq k\leq q}\) connecting~\(m'\) to~\(m\) using~\(-T^{(1)}\).

One should think of~\(-T^{(1)}\) as describing a macroscopic jump process between fibers, where the \((\cL_{\rm mac}-K)\)-jump rates have been averaged under the microscopic measures~\((\mu_m)_{1\leq m\leq M}\).

By the Perron--Frobenius theorem, we have a spectral gap
\begin{equation}
    \label{eq:macro_spectral_gap}
    \gamma^{\rm s} := \mathfrak{Re}\left(\eta_2^{(1)}-\eta_1^{(1)}\right) = \mathfrak{Re}\left(\eta_2^{(1)}\right)-\eta_1^{(1)} >0,
\end{equation}
and positive principal eigenvectors
\begin{equation}
    \label{eq:coarse_grained_eigenpair}
    h_0^{\rm s} = \sum_{m=1}^M \alpha_m\1_{\cX_m}\in P_0^{\rm s} \ell^\infty(\cX),\qquad \nu_0^{\rm s} = \sum_{m=1}^M \beta_m \mu_m\in \cP(\cX),
\end{equation}
such that~\(\alpha,\beta>0\) pointwise,~\(\beta\in\cP(\cZ)\), and satisfying the eigenrelations and normalization convention
\begin{equation}
    \label{eq:macro_eigenrelations}
    T^{(1)}h_0^{\rm s} = \eta_1^{(1)} h_0^{\rm s},\qquad \nu_0^{\rm s} T^{(1)}=\eta_1^{(1)}\nu_0^{\rm s},\qquad \nu_0^{\rm s}(h_0^{\rm s})=1.
\end{equation}
Since~\(T^{(1)}\1_{\cX} = P_0^{\rm s} K\1_{\cX} \geq 0\) has a strictly positive component (because~\(K\) is nonnegative and nonzero), we have~\(\nu_0^{\rm s} T^{(1)}\1_{\cX} = \eta_1^{(1)} = \nu_0^{\rm s}(P_0^{\rm s} K\1_{\cX})>0\).

We introduce two disjoint closed contours,~\(\Gamma_1^{\rm s}\) and~\(\Gamma_2^{\rm s}\), enclosing~\(-\eta_1^{(1)}\) and~\(\sigma(-T^{(1)})\setminus\{-\eta_1^{(1)}\}\) respectively, verifying
\begin{equation}
    \Gamma_2^{\rm s}\subset \left\{z\in\bbC:\,{\mathfrak Re}\,z<-\eta_1^{(1)}-\gamma^{\rm s}/2\right\}.
\end{equation}
We also define these such that~\(\varepsilon\Gamma_1^{\rm s}\) and~\(\varepsilon\Gamma_2^{\rm s}\) are both enclosed by~\(\Gamma^{\rm s}\) for~\(0<\varepsilon<\varepsilon_0\).
A schematic representation of the clustering of~\(\sigma(-T(\varepsilon))\) by~\(\Gamma^{\rm s}_1\) and~\(\Gamma^{\rm s}_2\) is given in the right panel of Figure~\ref{fig:contours}.

\paragraph{Uniform version of Proposition~\ref{prop:spectral_properties}.}
We consider two successive perturbations, corresponding to the two levels of clustering introduced in the previous paragraph. Let~\(U^{\rm s}_\varepsilon\) denote Kato's connection~\eqref{eq:kato_connection} associated with~\(P^{\rm s}_\varepsilon\) acting on~\(\ell^\infty(\cX)\). The conjugate
\begin{equation}
    \label{eq:reduced_slow_operator}
    T^{\rm s}_\varepsilon = \varepsilon^{-1}U^{\rm s}_\varepsilon T(\varepsilon)\left(U^{\rm s}_\varepsilon\right)^{-1}
\end{equation}
has~\(P^{\rm s}_0\ell^\infty(\cX)\) as an invariant subspace, and restricted to this space has eigenvalues~\((\eta_j(\varepsilon)/\varepsilon)_{1\leq j\leq M}\). By a direct computation (since~\(P_0^{\rm s}T(0)=T(0)P_0^{\rm s}=0\)) we have
\begin{equation}
    \label{eq:projector_quadratic_bound}
    \|P^{\rm s}_\varepsilon T(0)P^{\rm s}_\varepsilon\|= \|(P^{\rm s}_\varepsilon-P^{\rm s}_0)T(0)(P^{\rm s}_\varepsilon-P^{\rm s}_0)\|\leq C\varepsilon^2,
\end{equation}
so that~\eqref{eq:projector_estimate} and~\eqref{eq:kato_connection} together with the intertwining relations~\(U_\varepsilon^{\rm s}P_\varepsilon^{\rm s}=P_0^{\rm s}U_{\varepsilon}^{\rm s}\),~\((U_{\varepsilon}^{\rm s})^{-1}P_0^{\rm s} = P_{\varepsilon}^{\rm s}(U_{\varepsilon}^{\rm s})^{-1}\) give
\begin{equation}
    \label{eq:reduced_slow_estimate}
    \begin{aligned}
        \left\|\left(T^{\rm s}_\varepsilon - T^{(1)}\right)P^{\rm s}_0\right\| & = \left\|\varepsilon^{-1}U^{\rm s}_\varepsilon P^{\rm s}_\varepsilon\left(T(0)-\varepsilon S\right)P^{\rm s}_\varepsilon\left(U^{\rm s}_\varepsilon\right)^{-1}+P^{\rm s}_0SP^{\rm s}_0\right\|                                                                            \\
                                                                               & \leq \varepsilon^{-1}\left\|U_\varepsilon^{\rm s}P_\varepsilon^{\rm s}T(0)P_\varepsilon^{\rm s}(U_{\varepsilon}^{\rm s})^{-1}\right\|+\left\|P_0^{\rm s}SP_0^{\rm s}-U_\varepsilon^{\rm s}P_\varepsilon^{\rm s}S P_\varepsilon^{\rm s}(U_\varepsilon^{\rm s})^{-1}\right\| \\
                                                                               & =\varepsilon^{-1}\left\|U_\varepsilon^{\rm s}P_\varepsilon^{\rm s}T(0)P_\varepsilon^{\rm s}(U_{\varepsilon}^{\rm s})^{-1}\right\| + \left\|P_0^{\rm s}(S-U_{\varepsilon}^{\rm s}S(U_\varepsilon^{\rm s})^{-1})P_0^{\rm s}\right\|                                          \\
                                                                               & \leq \varepsilon^{-1}C\varepsilon^2 + C\varepsilon\leq C\varepsilon,
    \end{aligned}
\end{equation}
where we used~\eqref{eq:projector_quadratic_bound} for the last inequality. That is,~\(T^{\rm s}_\varepsilon\) is a perturbation of~\(T^{(1)}\) on~\(P_0^{\rm s}\ell^\infty(\cX)\). Denote by~\(P_\varepsilon^{\rm s,1}\) the spectral projector associated to the~\(\Gamma^{\rm s}_1\)-part of~\(\sigma(-T_\varepsilon^{\rm s})\), and~\(U_\varepsilon^{\rm s,1}:P_\varepsilon^{\rm s,1}\ell^\infty(\cX) \to P_0^{\rm{s},1}\ell^\infty(\cX)\) the associated Kato connection, defined on~\(P_0^{\rm s}\ell^\infty(\cX)\). We stress that~\(P_{\varepsilon}^{\rm s,1}\) is a projector on~\(P_0^{\rm s}\ell^\infty(\cX)\), not~\(P_{\varepsilon}^{\rm s}\ell^\infty(\cX)\).
Upon shrinking~\(\Gamma^{\rm s}_1\), we may assume that~\(\Gamma^{\rm s}_1\subset\{z\in\bbC:\,\mathfrak{Re}\,z>-\eta^{(1)}_1-\gamma^{\rm s}/4\}\). Since~\(P^{{\rm s},1}_0 = h^{\rm s}_0\nu^{\rm s}_0\) has rank one, so does~\(P^{{\rm s},1}_\varepsilon\) for~\(\varepsilon\in(0,\varepsilon_0)\), and~\(-T^{\rm s}_\varepsilon\) has a unique, simple eigenvalue enclosed by~\(\Gamma^{\rm s}_1\), which is necessarily the principal eigenvalue~\(-\lambda^X_\varepsilon\) of the similar operator~\(\cL^X_\varepsilon=-T(\varepsilon)/\varepsilon\), since all other eigenvalues of~\(\cL_\varepsilon^X\) are enclosed by~\(\Gamma_2^{\rm s}\) or~\(\varepsilon^{-1}\Gamma^{\rm f}\).
In particular, \(\lambda^X_\varepsilon = \eta_1(\varepsilon)/\varepsilon<\eta^{(1)}_1+\gamma^{\rm s}/4\).

We now write the principal spectral projector of~\(\cL^X_\varepsilon\), namely~\((\Id-P_\varepsilon^X) = h^X_\varepsilon\nu^X_\varepsilon\), via transport of~\(P_0^{\rm s,1}\) through the Kato connections, which gives
\begin{equation}
    \label{eq:principal_projector}
    h^X_\varepsilon\nu^X_\varepsilon = (U_\varepsilon^{\rm s})^{-1}(U_{\varepsilon}^{\rm s,1})^{-1}P_0^{\rm s,1}U_{\varepsilon}^{\rm s,1}U_\varepsilon^{\rm s}P^{\rm s}_\varepsilon,\end{equation}
so that~\eqref{eq:kato_connection} and~\(P_0^{\rm s,1}P_0^{\rm s}=h_0^{\rm s}\nu_0^{\rm s}\) give~\(\|h_\varepsilon^X\nu_\varepsilon^X - h_0^{\rm s}\nu_0^{\rm s}\|\leq C\varepsilon\).
Since~\(\nu^X_\varepsilon \1_\cX = \nu_0^{\rm s}\1_{\cX}=1\), we have~\(\|h_\varepsilon^X-h_0^{\rm s}\|\leq \|h_\varepsilon^X\nu_\varepsilon^X - h_0^{\rm s}\nu_0^{\rm s}\|\|\1_{\cX}\|\leq C\varepsilon\).
Therefore,~\(|\nu_0^{\rm s}h_\varepsilon^{X}-\nu_0^{\rm s}h_0^{\rm s}|\leq C\varepsilon\), and writing~\(\nu_\varepsilon^X = \frac{\nu_0^{\rm s}h_\varepsilon^X \nu_\varepsilon^X}{\nu_0^{\rm s}h_\varepsilon^X}\), we get
\begin{equation}
    \label{eq:qsd_ratios}
    \|\nu_\varepsilon^X-\nu_0^{\rm s}\|_{*} \leq \|\nu_0^{\rm s}\|_{*}\left\|\frac{h_\varepsilon^X \nu_\varepsilon^X}{\nu_0^{\rm s}h_{\varepsilon}^X}-h_0^{\rm s}\nu_0^{\rm s}\right\|\leq C\varepsilon,
\end{equation}
which together give
\begin{equation}
    \label{eq:principal_convergence}
    \|h^X_\varepsilon-h^{\rm s}_0\|+\|\nu^X_\varepsilon-\nu^{\rm s}_0\|_*\leq C\varepsilon,\qquad \min_{\cX}h^X_\varepsilon\geq \frac12\min_{1\leq m\leq M}\alpha_m.
\end{equation}
The lower bound comes from the explicit expression~\eqref{eq:coarse_grained_eigenpair} for \(h_0^{\rm s}\).

Since~$\cL_{\rm mic},\cL_{\rm mac}$ are mass-conserving,~$\cL_\varepsilon^X \1_{\cX} = -K\1_{\cX}$, so that~$\lambda_\varepsilon^X = \nu_\varepsilon^X(K\1_{\cX})$. Similarly, since~$P_0^{\rm s}\1_{\cX}=\1_{\cX}$,~$\lambda_0^{\rm s} = \eta_1^{(1)} = \nu_0^{\rm s}(P_0^{\rm s}K\1_{\cX}) =\nu_0^{\rm s}(K\1_{\cX})$ (recall~$K$ is a diagonal operator), so that
\begin{equation}
    \label{eq:principal_eigenvalue_convergence}
    |\lambda_\varepsilon^X-\lambda_0^{\rm s}| \leq \|\nu_\varepsilon^X-\nu_0^{\rm s}\|_*\|K\1_{\cX}\|\leq C\varepsilon,
\end{equation}
from~\eqref{eq:principal_convergence}. This shows the second inequality in~\eqref{eq:adiabatic_limit}.

Finally, using~\eqref{eq:sum_of_projectors}, we may write \(P^X_\varepsilon = \Id-h^X_\varepsilon\nu^X_\varepsilon = P^{\rm f}_\varepsilon + (U^{\rm s}_\varepsilon)^{-1}P^{{\rm s},2}_\varepsilon U^{\rm s}_\varepsilon P^{\rm s}_\varepsilon\).
Applying the semigroup estimate~\eqref{eq:semigroup_estimate} to~\(-T(\varepsilon)\) with~\(\Gamma^{\rm f}\) at time~\(t/\varepsilon\), and to~\(-T^{\rm s}_\varepsilon\) with~\(\Gamma^{\rm s}_2\) at time~\(t\), we get
\begin{equation}
    \label{eq:semigroup_estimate_X_gamma}
    \forall\,t\geq 0,\qquad \|Q^X_{t,\varepsilon}P^X_\varepsilon\|\leq C\left(\e^{-\gamma^{\rm f}t/(4\varepsilon)}+\e^{-(\eta^{(1)}_1+\gamma^{\rm s}/2)t}\right)\leq C\e^{-(\lambda^X_\varepsilon+\gamma^{\rm s}/4)t}
\end{equation}
for~\(\varepsilon\) small enough. This verifies the estimate~\eqref{eq:spectral_gap_estimate} from Proposition~\ref{prop:spectral_properties} with a uniform spectral gap~\(\gamma^X_\varepsilon = \gamma^{\rm s}/4 \geq 1/C\) and prefactor~\(C^X_{\gamma,\varepsilon}\leq C\) uniformly in~\(\varepsilon\in(0,\varepsilon_0)\).
Decomposing the semigroup~\(Q^X_{t,\varepsilon} = \e^{-\lambda^X_\varepsilon t}h^X_\varepsilon\nu^X_\varepsilon+Q^X_{t,\varepsilon}P^X_\varepsilon\), we then estimate using the convergence~\(h_\varepsilon^X\nu_\varepsilon^X \to h_0^{\rm s}\nu_0^{\rm s}\)
\begin{equation}
    \label{eq:semigroup_estimate_X_lambda}
    \forall\,t\geq 0,\qquad \|Q^X_{t,\varepsilon}\| \leq C\e^{-\lambda^X_\varepsilon t},
\end{equation}
which proves the estimate~\eqref{eq:principal_decay} from Proposition~\ref{prop:spectral_properties} with a uniform prefactor~\(C^X_{\lambda,\varepsilon}\leq C\).
Note this argument could be optimized to obtain a sharper spectral gap.

\paragraph{Convergence of~\(\Pi_\varepsilon^X\).}
Define~\(\Pi^Z_0 : \ell^\infty(\cX)\to \ell^\infty(\cZ)\) by~\(\Pi^Z_0 f(m) = \mu_m(f)\) for~\(m\in\cZ\), so that~\(P^{\rm s}_0 = \xi^\star\Pi^Z_0\) (see~\eqref{eq:principal_eigenprojector}). Since~\(\nu^{\rm s}_0 = \sum_m\beta_m\mu_m\), the operator~\(\Pi^Z_0\) is defined by the disintegration of~\(\nu^{\rm s}_0\) with respect to~\(\xi\), just as~\(\Pi^Z_\varepsilon\) was defined by the disintegration of~\(\nu^X_\varepsilon\) with respect to~\(\xi\) (see~\eqref{eq:nu_disintegration}).
By the convergence~\eqref{eq:principal_convergence},~\(\nu_\varepsilon^X \1_{\cX_m} \to \nu^{\rm s}_0\1_{\cX_m} = \beta_m>0\) (see~\eqref{eq:coarse_grained_eigenpair}), and~\(\|\nu_\varepsilon^X f-\nu_0^{\rm s}f\|\leq C\varepsilon\|f\|\), we obtain
\begin{equation}
    \label{eq:fiber_operator_convergence}
    \|\Pi^Z_\varepsilon-\Pi^Z_0\|+\|\Pi^X_\varepsilon-P^{\rm s}_0\|\leq (1+\|\xi^\star\|)\|\Pi^Z_\varepsilon-\Pi^Z_0\|\leq C\varepsilon,
\end{equation}
which proves the first estimate in~\eqref{eq:adiabatic_limit}.

\paragraph{Separation of timescales and uniform closure defect.}
Let~\(\Pi^\perp_\varepsilon = \Id-\Pi^X_\varepsilon\), and consider the operators~\(H_\varepsilon = \varepsilon\Pi^\perp_\varepsilon\cL^X_\varepsilon\Pi^\perp_\varepsilon\). Since~\(\Pi^X_\varepsilon \ell^\infty(\cX) \subset P_0^{\rm s}\ell^\infty(\cX)\) (\(\Pi_\varepsilon^X\) is a fiber-averaging operator), we have~\(R\Pi^X_\varepsilon = 0\). Using~\(P^{\rm s}_0R = 0\) and~\eqref{eq:fiber_operator_convergence}, we estimate
\begin{equation}
    \label{eq:compressed_operator_perturbation_bound}
    \begin{aligned}
        \|H_\varepsilon-R\| & \leq \left\|(\Id-\Pi_\varepsilon^X)R(\Id-\Pi_\varepsilon^X)-R\right\| +\varepsilon\|(\Id-\Pi_\varepsilon^X)S(\Id-\Pi_\varepsilon^X)\| \\
                            & \leq \left\|(\Id-\Pi_\varepsilon^X)R -R\right\| + C\varepsilon                                                                        \\
                            & = \left\|(P_0^{\rm s}-\Pi_\varepsilon^X)R\right\|+C\varepsilon                                                                        \\
                            & \leq C\varepsilon.
    \end{aligned}
\end{equation}
That is, \(H_\varepsilon\) is a perturbation of~\(R\) vanishing on the~\(M\)-dimensional space~\(\Pi^X_\varepsilon \ell^\infty(\cX)\) and leaving~\(\Pi^\perp_\varepsilon \ell^\infty(\cX)\) invariant. By perturbation theory,~\(H_\varepsilon\) has exactly~\(M\) eigenvalues enclosed by~\(\Gamma^{\rm s}\), which are therefore its zero eigenvalues on~\(\Pi^X_\varepsilon \ell^\infty(\cX)\), and the spectral projector of~\(H_\varepsilon\) associated with~\(\Gamma^{\rm f}\) is~\(\Pi^\perp_\varepsilon\) (take~\(\Gamma\) such that~$P_\varepsilon(\Gamma)=\Id$ in~\eqref{eq:sum_of_projectors}). Applying the semigroup estimate~\eqref{eq:semigroup_estimate} at time~\(t/\varepsilon\) then gives
\begin{equation}
    \forall\,t\geq 0,\qquad \left\|\e^{t\Pi^\perp_\varepsilon\cL^X_\varepsilon\Pi^\perp_\varepsilon}\Pi^\perp_\varepsilon\right\|\leq C\e^{-\gamma^{\rm f}t/(2\varepsilon)},
\end{equation}
which verifies~\eqref{eq:separation_of_timescales} with~\(\zeta_\varepsilon = \gamma^{\rm f}/(2\varepsilon) \geq 1/(C\varepsilon)\) and~\(C_{\zeta,\varepsilon}\leq C\) uniformly in~\(\varepsilon\in(0,\varepsilon_0)\). In particular,~\(\zeta_\varepsilon>2(\lambda^X_\varepsilon+\gamma^X_\varepsilon)\) for~\(\varepsilon<\varepsilon_0\) small enough, and the assumptions of Theorem~\ref{thm:cg_existential_result} have been verified.

To control~\(A_\varepsilon^X\), we perform a similar computation, writing
\begin{equation}
    \begin{aligned}
        A_\varepsilon^X & = \cL_\varepsilon^X\Pi_\varepsilon^X - \Pi_\varepsilon^X \cL_\varepsilon^X                      \\
                        & = (\varepsilon^{-1}R+S)\Pi_\varepsilon^X - \Pi_\varepsilon^X (\varepsilon^{-1}R+S)              \\
                        & =(S\Pi_\varepsilon^X - \Pi_\varepsilon^X S) + \varepsilon^{-1}(P_0^{\rm s}-\Pi_\varepsilon^X)R.
    \end{aligned}
\end{equation}
A triangle inequality together with the convergence~\eqref{eq:fiber_operator_convergence} then yields~\(\|A_\varepsilon^X\| \leq \|S\Pi_\varepsilon^X - \Pi_\varepsilon^X S\| + \varepsilon^{-1}C\varepsilon \leq C\) uniformly in~\(\varepsilon\in(0,\varepsilon_0)\).
In particular, the approximation parameter in Theorem~\ref{thm:cg_existential_result} satisfies~\(\|A^X_\varepsilon\|/\zeta_\varepsilon\leq C\varepsilon\).

\paragraph{Uniform estimates for~\(Z_\varepsilon\).}
Since~\(R\xi^\star = 0\) on~\(\ell^\infty(\cZ)\), the generator~\eqref{eq:cg_generator} of the canonical model is given by~\(\cL^Z_\varepsilon = \Pi^Z_\varepsilon S\xi^\star\), and by~\eqref{eq:fiber_operator_convergence},~\(\|\cL^Z_\varepsilon-\cL_0^Z\|\leq C\varepsilon\), where~\(\cL_0^Z = \Pi^Z_0 S\xi^\star\) is conjugated with~\(-T^{(1)}\) via the isomorphism~\(\xi^\star: \ell^\infty(\cZ)\to P_0^{\rm s}\ell^\infty(\cX)\). One can simply check that~\(\cL_0^Z\) has principal eigenvectors~\(\alpha \in \ell^\infty(\cZ)\) and~\(\beta \in \cP(\cZ)\), where we recall~\eqref{eq:coarse_grained_eigenpair}.

By perturbation theory and the clustering of~\(\sigma(-T^{(1)})\) from above,~\(\cL_\varepsilon^Z\) has exactly one simple eigenvalue enclosed in~\(\Gamma_1^{\rm s}\), with the rest of the spectrum enclosed by~\(\Gamma_2^{\rm s}\) for~\(\varepsilon\in(0,\varepsilon_0)\).
Denote~\(P_\varepsilon^{Z,1} = h_\varepsilon^Z \nu_\varepsilon^Z\) and~\(P_\varepsilon^{Z,2}\) the associated spectral projectors, satisfying~\(\Id = P_\varepsilon^{Z,1}+P_\varepsilon^{Z,2}\) on~\(\ell^\infty(\cZ)\). In particular~$P_\varepsilon^Z = P_\varepsilon^{Z,2}$.

Applying~\eqref{eq:projector_estimate}, we get~\(\|h_\varepsilon^Z\nu_\varepsilon^Z-\alpha\beta\|\leq C\varepsilon\), and reasoning as for~\eqref{eq:principal_convergence}, we get
\begin{equation}
    \label{eq:principal_convergence_Z}
    \|h^Z_\varepsilon-\alpha\|+\|\nu^Z_\varepsilon-\beta\|_*\leq C\varepsilon,\qquad \min_{\cZ}h^Z_\varepsilon\geq \frac12\min_{1\leq m\leq M}\alpha_m.
\end{equation}

Applying the semigroup estimate~\eqref{eq:semigroup_estimate} (as we did for~\eqref{eq:semigroup_estimate_X_gamma}, but without the contribution of~$\Gamma^{\rm f}$) to~\(\cL_\varepsilon^Z\) with~\(\Gamma^{\rm s}_2\), we obtain
\begin{equation}
    \label{eq:semigroup_estimate_Z_gamma}
    \forall\,t\geq 0,\qquad \|Q^Z_{t,\varepsilon}P^Z_\varepsilon\|\leq C\e^{-(\lambda^X_\varepsilon+\gamma^{\rm s}/4)t} = C\e^{-(\lambda^Z_\varepsilon+\gamma^{\rm s}/4)t},
\end{equation}
and decomposing~\(Q^Z_{t,\varepsilon} = \e^{-\lambda^Z_\varepsilon t} h_\varepsilon^Z \nu_\varepsilon^Z + Q^Z_{t,\varepsilon}P^Z_\varepsilon\) as for~\eqref{eq:semigroup_estimate_X_lambda} yields
\begin{equation}
    \label{eq:semigroup_estimate_Z_lambda}
    \forall\,t\geq 0,\qquad \|Q^Z_{t,\varepsilon}\|\leq C\e^{-\lambda_\varepsilon^Z t},
\end{equation}
which verifies Proposition~\ref{prop:spectral_properties} for~\(Z_\varepsilon\), with a uniform spectral gap~\(\gamma_\varepsilon^Z \geq 1/C\), and uniform prefactors~\(C_{\lambda,\varepsilon}^Z,C_{\gamma,\varepsilon}^Z\leq C\).
This finishes the proof of~\eqref{eq:adiabatic_constants_control}.

\paragraph{$h^Y_\varepsilon$-norm estimates and dependence on~\(\eta_0^X\).}
We finally prove the last two estimates in~\eqref{eq:adiabatic_limit} and the estimates~\eqref{eq:initial_condition_control}.

We start with~\eqref{eq:adiabatic_limit}. For~$Y=X,Z$, notice that
\begin{equation}
    \begin{aligned}
        &\left|\underset{\cY}{\min}\,\frac{\Pi_\varepsilon^Y h_\varepsilon^X}{h_\varepsilon^Y} -1 \right|=\left|\underset{\cY}{\min}\,\frac{\Pi_\varepsilon^Y h_\varepsilon^X - h_\varepsilon^Y}{h_\varepsilon^Y}\right|\leq \left(\underset{\cY}{\min}\,h_\varepsilon^Y\right)^{-1}\|\Pi_\varepsilon^Y h_\varepsilon^X-h_\varepsilon^Y\|\\
        &\qquad\leq C \|\Pi_\varepsilon^Y h_\varepsilon^X-h_\varepsilon^Y\|,
    \end{aligned}
\end{equation}
where we use the last estimate in either~\eqref{eq:principal_convergence} or~\eqref{eq:principal_convergence_Z} in the last line. We then use the quasimode estimates~\eqref{eq:X_quasimode_estimate} and~\eqref{eq:Z_quasimode_estimate}, together with~\eqref{eq:adiabatic_constants_control} to control the constants, and conclude that
\begin{equation}
    \left|\underset{\cY}{\min}\,\frac{\Pi_\varepsilon^Y h_\varepsilon^X}{h_\varepsilon^Y} -1 \right|\leq C\varepsilon,
\end{equation}
which finishes the proof of~\eqref{eq:adiabatic_limit}.

Finally, we verify~\eqref{eq:initial_condition_control}.
We write
\begin{equation}
    \left|\eta_0^X(h_\varepsilon^X)-\eta_0^X(h_0^{\rm s})\right|\leq \|\eta_0^X\|_*\|h_\varepsilon^X-h_0^{\rm s}\| = \|h_\varepsilon^X-h_0^{\rm s}\|\leq C\varepsilon
\end{equation}
using~\eqref{eq:principal_convergence}, and using the compatibility condition~$\eta_0^Z=\xi_\star\eta_0^X$,
\begin{equation}
    \left|\eta_0^Z(h_\varepsilon^Z)-\eta_0^X(h_0^{\rm s})\right|=\left|\eta_0^Z(h_\varepsilon^Z)-\eta_0^X(\xi^\star\alpha)\right|\leq \|\eta_0^Z\|_*\|h_\varepsilon^Z-\alpha\|=\|h_\varepsilon^Z-\alpha\|\leq C\varepsilon,
\end{equation}
using~\eqref{eq:principal_convergence_Z}. Finally, the lower bound~$\eta_0^X(h_0^{\rm s})\geq 1/C$ follows from the strict positivity of~$\alpha$.

\section{Implementation details}
\label{sec:details}
\subsection{Synthetic prior construction}
\label{sec:synth_prior_specification}
We describe the hand-crafted synthetic prior used in the experiments of Section~\ref{sec:num}. All random draws (indicated by~\(\sim\)) appearing in this paragraph are independent.

A sample from~\(\rho_{\rm dyn}\) is specified by random physical parameters~\(U\), \(D\) and a random inverse temperature~\(\beta\sim\cU([\beta_{\min},\beta_{\max}])\).
The energy function~\(U\) is given by first constructing the random function
\begin{equation}
    \label{eq:prior_energy}
    \widetilde U(z) = 1-\e^{-\frac{(z-c)^2}{2\kappa^2}} + \sum_{i=1}^{m_U} h_i\,\e^{-\frac{(z-c_{i,U})^2}{2\kappa_{i,U}^2}} + bz,
\end{equation}
composed of a Gaussian well centered at \({c\sim\cU([0.15,0.85])}\) and of width \({\kappa\sim\cU([0.2,0.6])}\), perturbed by \({m_U\sim\cU\{1,\dots,6\}}\) internal energy barriers, each given, for \({1\leq i\leq m_U}\), by a Gaussian profile centered at \({c_{i,U}\sim\cU([0,1])}\),
of height \({h_i\sim\cU([0,\delta_{\max}])}\) and of width \({\kappa_{i,U}}=\kappa_{i,U}'/m_U\), where \({\kappa_{i,U}'\sim\cU([0.2,0.6])}\).
A linear correction term \(b z\) is added to enforce \({|\widetilde U(1)-\widetilde U(0)|=\delta_E\sim\cU([0,\delta_{\max}])}\).

Then, \(U\) is constructed using an affine transformation of \(\widetilde U\) to enforce \({U([0,1])=[0,1]}\). Setting a fixed energy scale is motivated by the fact that the parameter \(\beta\) modulates the height of energy barriers relative to the thermal energy of the system.

The diffusion coefficient~\(D\) is constructed by first drawing a constant diffusion \({\sigma=\e^{s}}\) where \({s\sim\cU([s_{\min},s_{\max}])}\), which is either kept constant with probability~\(1/2\) (in which case~\(D(z)=\sigma\)), or else perturbed by~\(m_D\sim\cU\{1,\dots,6\}\) Gaussian profiles
\begin{equation}
    \label{eq:prior_diffusion}
    D(z) = \sigma + q(z) - \min q, \qquad q(z)=\sum_{i=1}^{m_D} \sigma g_i\,\e^{-\frac{(z-c_{i,D})^2}{2(\kappa_{i,D})^2}},
\end{equation}
with~\(g_i\sim\mathcal{N}(0,1)\), \(c_{i,D}\sim\cU([0,1])\) and~\(\kappa_{i,D}=\kappa_{i,D}'/m_D\), \(\kappa_{i,D}'\sim\cU([0.2,0.6])\), so that~\(\min_{[0,1]} D=\sigma\).

The prior~\(\rho_{0}\) is specified by sampling~\(Z_0\sim \cU(\cZ)\) uniformly on the grid~$\cZ\subset(0,1)$.

The various (training and misspecified) priors of Section~\ref{sec:num_synth} differ in the ranges \({[\beta_{\min},\beta_{\max}]}\), \({[s_{\min},s_{\max}]}\) and~\(\delta_{\max}\).
All the other parameter distributions entering~\eqref{eq:prior_energy} and~\eqref{eq:prior_diffusion} are fixed.
The parameter ranges used in the experiments of Section~\ref{sec:num_synth} are given in Table~\ref{tab:prior_shifts}.

\begin{table}[H]
    \centering\small
    \begin{tabular}{lccc}
        \toprule
        prior             & \([\beta_{\min},\beta_{\max}]\) & \([s_{\min},s_{\max}]\) & \(\delta_{\max}\) \\
        \midrule
        training          & \([1,3]\)                       & \([-4,-1]\)             & \(0.2\)           \\
        high \(D\)          & \([1,3]\)                       & \([-1,1]\)              & \(0.2\)           \\
        low \(T\)           & \([4,6]\)                       & \([-4,-1]\)             & \(0.2\)           \\
        low \(T\), high \(D\) & \([4,6]\)                       & \([-1,1]\)              & \(0.2\)           \\
        low \(T\), one well & \([4,6]\)                       & \([-4,-1]\)             & \(0.05\)          \\
        \bottomrule
    \end{tabular}
    \caption{Parameter ranges for the training prior and the misspecified priors considered in Section~\ref{sec:num_synth}.}
    \label{tab:prior_shifts}
\end{table}
\subsection{Network architecture}
\label{sec:nn_architecture}
We describe the architecture of the RNN model used for the experiments of Section~\ref{sec:num}. The details are provided in Table~\ref{tab:architecture}.
\begin{table}[H]
    \centering\small
    \begin{tabular}{llr}
        \toprule
        layer type(/non-linearity)                                                          & output dimension     & number of parameters \\
        \midrule
        \multicolumn{3}{l}{\emph{input}: \(H_k\in\R^{d_{\rm hist}}\), \(d_{\rm hist}=64\), \(m(N,\tau)=\sqrt{N\tau}\in\R\)}                                        \\
        \midrule
        \multicolumn{3}{l}{\emph{featurizer} \(f_\theta:\R^{d_{\rm hist}}\to\R^{d_{\rm cnn}}\)}                                             \\
        \quad \texttt{Conv1D}\((5,\,1\to 16)\)/\texttt{MaxPool}\(\,_2\circ\)\texttt{leakyReLU}  & \(32\times 16\)        & \(96\)                 \\
        \quad \texttt{Conv1D}\((5,\,16\to 32)\)/\texttt{MaxPool}\(\,_2\circ\)\texttt{leakyReLU} & \(16\times 32\)        & \(2\,592\)             \\
        \quad \texttt{Conv1D}\((5,\,32\to 64)\)/\texttt{MaxPool}\(\,_2\circ\)\texttt{leakyReLU} & \(8\times 64\)         & \(10\,304\)            \\
        \midrule
        \multicolumn{3}{l}{\emph{reshape layer} \(f_\theta(H_k)\star m(N,\tau)\)}                                                           \\
        \quad \texttt{concatenate}\(\circ\)\texttt{flatten}                                   & \(d_{\rm cnn} +1=513\) & 0                    \\
        \midrule
        \multicolumn{3}{l}{\emph{recurrent layer} \({\rm LSTM}_\theta\)}                                                                    \\
        \quad \texttt{LSTM}\((513\to 64)\)                                                    & \(d_{\rm rnn}=64\)     & \(147\,968\)           \\
        \quad \texttt{LSTM}\((64\to 64)\)                                                     & \(d_{\rm rnn}\)        & \(33\,024\)            \\
        \midrule
        \multicolumn{3}{l}{\emph{classifier} \(c_\theta:\R^{d_{\rm rnn}}\to[0,1]\)}                                                         \\
        \quad \texttt{Dense}\((64\to 16)\)/\texttt{leakyReLU}                                 & \(16\)                 & \(1\,040\)             \\
        \quad \texttt{Dense}\((16\to 16)\)/\texttt{leakyReLU}                                 & \(16\)                 & \(272\)                \\
        \quad \texttt{Dense}\((16\to 1)\)/\texttt{sigmoid}                                    & \(1\)                  & \(17\)                 \\
        \midrule
        \multicolumn{2}{l}{Total parameters:}                                               & \(195\,313\)                                   \\
        \bottomrule
    \end{tabular}
    \caption{Architecture of the deployed RNN diagnostic. Convolutional layers use a kernel width of~\(5\) and a zero-padding size of~\(2\).}
    \label{tab:architecture}
\end{table}

\bibliographystyle{abbrv}
\bibliography{bibliography.bib}

@book{AT17,
  title     = {Computer Simulation of Liquids},
  author    = {Allen, M.P. and Tildesley, D.J.},
  year      = {2017},
  publisher = {Oxford University Press}
}

@article{FV79,
  title={Some measure-valued {M}arkov processes in population genetics theory},
  author={Fleming, W.H. and Viot, M.},
  journal={Indiana University Mathematics Journal},
  volume={28},
  number={5},
  pages={817--843},
  year={1979}
}

@article{HTB90,
  title={Reaction-rate theory: fifty years after {K}ramers},
  author={H{\"a}nggi, P. and Talkner, P. and Borkovec, M.},
  journal={Reviews of Modern Physics},
  volume={62},
  number={2},
  pages={251},
  year={1990}
}

@article{BCOV25,
  title   = {Degenerate processes killed at the boundary of a domain},
  author  = {Bena{\"\i}m, M. and Champagnat, N. and O{\c{c}}afrain, W. and Villemonais, D.},
  journal = {The Annals of Probability},
  volume  = {53},
  number  = {2},
  pages   = {720--752},
  year    = {2025}
}

@article{DLPSS18,
  title={Quantification of coarse-graining error in {L}angevin and overdamped {L}angevin dynamics},
  author={Duong, M.H. and Lamacz, A. and Peletier, M.A. and Schlichting, A. and Sharma, U.},
  journal={Nonlinearity},
  volume={31},
  number={10},
  pages={4517--4566},
  year={2018}
}

@article{DF90,
  title={Strong stationary times via a new form of duality},
  author={Diaconis, P. and Fill, J.A.},
  journal={The Annals of Probability},
  volume={18},
  number={4},
  pages={1483--1522},
  year={1990}
}

@article{RP81,
  title={Markov functions},
  author={Rogers, L.C.G. and Pitman, J.W.},
  journal={The Annals of Probability},
  volume={9},
  number={4},
  pages={573--582},
  year={1981}
}

@article{LLO17,
  title={Pathwise estimates for an effective dynamics},
  author={Legoll, F. and Leli{\`e}vre, T. and Olla, S.},
  journal={Stochastic Processes and their Applications},
  volume={127},
  number={9},
  pages={2841--2863},
  year={2017}
}

@article{TS14,
  title={Exponential-{K}rylov methods for ordinary differential equations},
  author={Tranquilli, P. and Sandu, A.},
  journal={Journal of Computational Physics},
  volume={278},
  pages={31--46},
  year={2014}
}

@book{BDH16,
  title     = {Metastability: {A} {P}otential-{T}heoretic {A}pproach},
  author    = {Bovier, A. and Den Hollander, F.},
  volume    = {351},
  series    = {A Series of Comprehensive Studies in Mathematics},
  year      = {2016},
  publisher = {Springer}
}

@article{BLS15,
  title     = {A generalized parallel replica dynamics},
  author    = {Binder, A. and Leli{\`e}vre, T. and Simpson, G.},
  journal   = {Journal of Computational Physics},
  volume    = {284},
  pages     = {595--616},
  year      = {2015},
  publisher = {Elsevier}
}

@article{BLS25a,
  title   = {Quantitative low-temperature asymptotics for reversible diffusions in temperature-dependent domains},
  author  = {Blassel, N. and Leli\`evre, T. and Stoltz, G.},
  year    = {2026},
  journal = {Journal de Math{\'e}matiques Pures et Appliqu{\'e}es},
  volume={216},
  pages={103996}
}

@article{BLS26,
  title   = {Shape optimization of metastable states},
  author  = {Blassel, N. and Leli\`evre, T. and Stoltz, G.},
  journal = {Mathematical Models and Methods in Applied Sciences},
  volume={36},
  number={11},
  pages={2397--2475},
  year    = {2026}
}

@book{CMSM13,
  title     = {Quasi-{S}tationary {D}istributions},
  author    = {Collet, P. and Mart{\'\i}nez, S. and San Mart{\'\i}n, J.},
  series    = {Probability and its Applications},
  year      = {2013},
  publisher = {Springer}
}

@article{CV23,
  title     = {General criteria for the study of quasi-stationarity},
  author    = {Champagnat, N. and Villemonais, D.},
  journal   = {Electronic Journal of Probability},
  volume    = {28},
  pages     = {1--84},
  year      = {2023},
  publisher = {The Institute of Mathematical Statistics and the Bernoulli Society}
}

@book{FW12,
  title     = {Random {Perturbations} of {Dynamical} {Systems}},
  publisher = {Springer},
  author    = {Freidlin, M.I. and Wentzell, A.D.},
  year      = {2012},
  series    = {A Series of Comprehensive Studies in Mathematics},
  volume    = {260}
}

@article{HKN04,
  title   = {Quantitative analysis of metastability in reversible diffusion processes via a {W}itten complex approach},
  author  = {Helffer, B. and Klein, M. and Nier, F.},
  journal = {Matem{\'a}tica Contempor{\^a}nea},
  volume  = {26},
  pages   = {41--85},
  year    = {2004}
}

@article{HL19,
  title     = {{G}en. par{R}ep: A first implementation of the generalized parallel replica dynamics for the long time simulation of metastable biochemical systems},
  author    = {H{\'e}din, F. and Leli{\`e}vre, T.},
  journal   = {Computer Physics Communications},
  volume    = {239},
  pages     = {311--324},
  year      = {2019},
  publisher = {Elsevier}
}

@article{V14,
  title={General approximation method for the distribution of {M}arkov processes conditioned not to be killed},
  author={Villemonais, D.},
  journal={ESAIM: Probability and Statistics},
  volume={18},
  pages={441--467},
  year={2014}
}

@article{CCJS26,
      title={Uniform in time weak convergence for a {F}leming--{V}iot particle system with hard killing}, 
      author={Cardaliaguet, P. and Cirant, M. and Jackson, J. and Souganidis, P.E.},
      year={2026},
      journal={preprint arXiv: 2609.22561}
}

@inproceedings{CJ07,
  title={Efficient {AUC} optimization for classification},
  author={Calders, T. and Jaroszewicz, S.},
  booktitle={European Conference on Principles of Data Mining and Knowledge Discovery},
  pages={42--53},
  year={2007},
  organization={Springer}
}

@article{KB14,
  title={Adam: {A} method for stochastic optimization},
  author={Kingma, D.P. and Ba, J.},
  journal={preprint arXiv:1412.6980},
  year={2014}
}

@misc{neps,
author = {Alipour, N. and Mallik, N. and Bergman, E. and Geburek, A.M. and Schrodi, S. and Basu, S. and Birinxhiku, L. and Abou Chakra, T. and Garibov, S. and Gaur, G. and Janowski, M. and Mekic, E. and Ambattu, P.A. and Rogalla, D. and Hvarfner, C. and Ru, B. and Carstensen, T. and Hog, J. and Hutter, F. and Stoll, D.},
title = {{Neural Pipeline Search (NePS)}},
howpublished = {\url{https://github.com/automl/neps}},
note = {Version 0.12.2},
year = {2024}
}

@misc{repo,
author={Blassel, N.},
title={Paper repository},
howpublished={\url{https://github.com/noeblassel/FlemingViotDiagnostics}},
year={2026}}

@article{flux,
  author    = {Innes, M. and Saba, E. and Fischer, K. and Gandhi, D. and Rudilosso, M.C. and Joy, N.M. and Karmali, T. and Pal, A. and Shah, V.},
  title     = {Fashionable Modelling with {Flux}},
  year      = {2018},
  journal = {preprint arXiv:1811.01457}
}

@article{JM25,
  title   = {Uniform convergence of the {F}leming--{V}iot process in a hard killing metastable case},
  author  = {Journel, L. and Monmarch{\'e}, P.},
  journal = {The Annals of Applied Probability},
  volume  = {35},
  number  = {2},
  pages   = {1019--1048},
  year    = {2025}
}

@article{JR24,
  title   = {The particle approximation of quasi-stationary distributions. {P}art I: concentration bounds in the uniform case},
  author  = {Journel, L. and Rousset, M.},
  journal = {preprint arXiv: 2412.15820},
  year    = {2024}
}

@article{GDKC18,
  title={Likelihood-free inference via classification},
  author={Gutmann, M.U. and Dutta, R. and Kaski, S. and Corander, J.},
  journal={Statistics and Computing},
  volume={28},
  number={2},
  pages={411--425},
  year={2018}
}

@book{K95,
  title     = {{P}erturbation {T}heory for {L}inear {O}perators},
  author    = {Kato, T.},
  volume    = {132},
  series    = {Classics in Mathematics},
  year      = {1995},
  publisher = {Springer}
}

@article{HM82,
  title={The meaning and use of the area under a receiver operating characteristic ({ROC}) curve},
  author={Hanley, J.A. and McNeil, B.J.},
  journal={Radiology},
  volume={143},
  number={1},
  pages={29--36},
  year={1982}
}

@article{CT16,
  title={Quantitative results for the {F}leming--{V}iot particle system and quasi-stationary distributions in discrete space},
  author={Cloez, B. and Thai, M.-N.},
  journal={Stochastic Processes and their Applications},
  volume={126},
  number={3},
  pages={680--702},
  year={2016},
}

@article{JM22,
  title={Convergence of a particle approximation for the quasi-stationary distribution of a diffusion process: {U}niform estimates in a compact soft case},
  author={Journel, L. and Monmarch{\'e}, P.},
  journal={ESAIM: Probability and Statistics},
  volume={26},
  pages={1--25},
  year={2022},
  publisher={EDP Sciences}
}

@article{LBLLP12,
  title   = {A mathematical formalization of the parallel replica dynamics},
  author  = {Le Bris, C. and Leli\`evre, T. and Luskin, M. and Perez, D.},
  journal = {Monte Carlo Methods and Applications},
  volume  = {18},
  number   = {2},
  year    = {2012},
  pages   = {119--146}
}

@article{LL10,
  title     = {Effective dynamics using conditional expectations},
  author    = {Legoll, F. and Leli\`evre, T.},
  journal   = {Nonlinearity},
  volume    = {23},
  number    = {9},
  pages     = {2131--2163},
  year      = {2010},
  publisher = {IOP Publishing}
}

@book{LRS10,
  title     = {Free Energy Computations -- A Mathematical Perspective},
  author    = {Leli\`evre, T. and Rousset, M. and Stoltz, G.},
  year      = {2010},
  publisher = {Imperial College Press}
}

@article{molly,
  author  = {Greener, J.G.},
  title   = {Differentiable simulation to develop molecular dynamics force fields for disordered proteins},
  journal = {Chemical Science},
  year    = {2024},
  volume  = {15},
  pages   = {4897--4909}
}

@article{MV12,
  title   = {Quasi-stationary distributions and population processes},
  author  = {M{\'e}l{\'e}ard, S. and Villemonais, D.},
  year    = {2012},
  journal = {Probability Surveys},
  volume  = {9},
  pages   = {340--410}
}

@book{M12,
  title={Machine learning: a probabilistic perspective},
  author={Murphy, K.P.},
  year={2012},
  publisher={MIT Press}
}

@article{PUV15,
  title   = {The parallel replica dynamics method--{Coming} of age},
  volume  = {100},
  number  = {B},
  journal = {Computational Materials Science},
  author  = {Perez, D. and Uberuaga, B.P. and Voter, A.F.},
  year    = {2015},
  pages   = {90--103}
}

@article{R20,
  title   = {Convergence diagnostics for {M}arkov chain {M}onte {C}arlo},
  author  = {Roy, V.},
  journal = {Annual Review of Statistics and Its Application},
  volume  = {7},
  number  = {1},
  pages   = {387--412},
  year    = {2020}
}

@article{SL13,
  title   = {Numerical analysis of parallel replica dynamics},
  author  = {Simpson, G. and Luskin, M.},
  journal = {ESAIM: Mathematical Modelling and Numerical Analysis},
  volume  = {47},
  number  = {5},
  pages   = {1287--1314},
  year    = {2013}
}

@article{SV00,
  title     = {Temperature-accelerated dynamics for simulation of infrequent events},
  author    = {S{\o}rensen, M.R. and Voter, A.F.},
  journal   = {The Journal of Chemical Physics},
  volume    = {112},
  number    = {21},
  pages     = {9599--9606},
  year      = {2000},
  publisher = {American Institute of Physics}
}

@article{V97,
  title     = {Hyperdynamics: accelerated molecular dynamics of infrequent events},
  author    = {Voter, A.F.},
  journal   = {Physical Review Letters},
  volume    = {78},
  number    = {20},
  pages     = {3908},
  year      = {1997},
  publisher = {APS}
}

@article{V97b,
  title   = {A method for accelerating the molecular dynamics simulation of infrequent events},
  author  = {Voter, A.F.},
  journal = {The Journal of Chemical Physics},
  volume  = {106},
  number  = {11},
  pages   = {4665--4677},
  year    = {1997}
}

@article{V98,
  title   = {Parallel replica method for dynamics of infrequent events},
  author  = {Voter, A.F.},
  journal = {Physical Review B},
  volume  = {57},
  number  = {22},
  pages   = {R13985},
  year    = {1998}
}

@article{V98b,
  title   = {Accelerating the dynamics of infrequent events: combining hyperdynamics and parallel replica dynamics to treat epitaxial layer growth},
  author  = {Voter, A.F. and Germann, T.C.},
  journal = {MRS Online Proceedings Library},
  volume  = {528},
  number  = {1},
  pages   = {221--236},
  year    = {1998}
}

@article{VMG02,
  title   = {Extending the time scale in atomistic simulation of materials},
  author  = {Voter, A.F. and Montalenti, F. and Germann, T.C.},
  journal = {Annual Review of Materials Research},
  volume  = {32},
  number  = {1},
  pages   = {321--346},
  year    = {2002}
}

@article{Y47,
  title   = {Some limit theorems in the theory of branching random processes},
  author  = {Yaglom, A.M.},
  journal = {Doklady Akademii Nauk SSSR (in Russian)},
  volume  = {56},
  number  = {8},
  pages   = {795--798},
  year    = {1947}
}

@article{ZHS16,
  title     = {Effective dynamics along given reaction coordinates, and reaction rate theory},
  author    = {Zhang, W. and Hartmann, C. and Sch{\"u}tte, C.},
  journal   = {Faraday Discussions},
  volume    = {195},
  pages     = {365--394},
  year      = {2016},
  publisher = {Royal Society of Chemistry}
}

@article{LNHP26,
  title   = {Data-driven characterization and acceleration of metastable dynamics using {K}oopman operators},
  author  = {Luzzatto, J. and N\"uske, F. and Hadjiconstantinou, N. G. and Perez, D.},
  journal = {Journal of Chemical Theory and Computation},
  volume  = {22},
  number  = {7},
  pages   = {3180--3196},
  year    = {2026},
}

@article{GR92,
  title   = {Inference from iterative simulation using multiple sequences},
  author  = {Gelman, A. and Rubin, D. B.},
  journal = {Statistical Science},
  volume  = {7},
  number  = {4},
  pages   = {457--472},
  year    = {1992}
}

@article{CBL20,
  title   = {The frontier of simulation-based inference},
  author  = {Cranmer, K. and Brehmer, J. and Louppe, G.},
  journal = {Proceedings of the National Academy of Sciences},
  volume  = {117},
  number  = {48},
  pages   = {30055--30062},
  year    = {2020}
}

@book{GCSR95,
  title={{B}ayesian data analysis},
  author={Gelman, A. and Carlin, J.B. and Stern, H.S. and Rubin, D.B.},
  year={1995},
  publisher={Chapman and Hall/CRC}
}

@article{MS94,
  title={Separation of a mixture of independent signals using time delayed correlations},
  author={Molgedey, L. and Schuster, H.G.},
  journal={Physical Review Letters},
  volume={72},
  number={23},
  pages={3634},
  year={1994}
}

@article{DBSMMGRLLMal25,
  title={Simulation-based inference: A practical guide},
  author={Deistler, M. and Boelts, J. and Steinbach, P. and Moss, G. and Moreau, T. and Gloeckler, M. and Rodrigues, P.L.C. and Linhart, J. and Lappalainen, J.K. and Miller, B.K. and others},
  journal={preprint arXiv:2508.12939},
  year={2025}
}

@article{LSBP13,
  title={To milliseconds and beyond: challenges in the simulation of protein folding},
  author={Lane, T.J. and Shukla, D. and Beauchamp, K.A. and Pande, V.S.},
  journal={Current Opinion in Structural Biology},
  volume={23},
  number={1},
  pages={58--65},
  year={2013},
}

@article{GL20,
  title   = {The physics of climate variability and climate change},
  author  = {Ghil, M. and Lucarini, V.},
  journal = {Reviews of Modern Physics},
  volume  = {92},
  number  = {3},
  pages   = {035002},
  year    = {2020}
}

@article{MACFGHLPSZ20,
  title   = {Long transients in ecology: {T}heory and applications},
  author  = {Morozov, A. and Abbott, K. and Cuddington, K. and Francis, T. and Gellner, G. and Hastings, A. and Lai, Y.-C. and Petrovskii, S. and Scranton, K. and Zeeman, M.L.},
  journal = {Physics of Life Reviews},
  volume  = {32},
  pages   = {1--40},
  year    = {2020}
}

@article{RGARMWB19,
  title={Metastable brain waves},
  author={Roberts, J.A. and Gollo, L.L. and Abeysuriya, R.G. and Roberts, G. and Mitchell, P.B. and Woolrich, M.W. and Breakspear, M.},
  journal={Nature Communications},
  volume={10},
  number={1},
  pages={1056},
  year={2019}
}

@article{BV12,
  title={Statistical mechanics of two-dimensional and geophysical flows},
  author={Bouchet, F. and Venaille, A.},
  journal={Physics Reports},
  volume={515},
  number={5},
  pages={227--295},
  year={2012}}

\end{document}